\UseRawInputEncoding
\documentclass{article}

\usepackage[preprint]{neurips_2026}

\usepackage[utf8]{inputenc}
\usepackage[T1]{fontenc}
\usepackage{microtype}
\usepackage{graphicx}
\usepackage{subcaption}
\usepackage{caption}
\usepackage{booktabs} 
\usepackage{amsmath}
\usepackage{amssymb}
\usepackage{mathtools}
\usepackage{amsthm}
\usepackage{siunitx}
\usepackage{bm}
\usepackage{xcolor}
\usepackage{amsfonts}
\usepackage{nicefrac}
\usepackage{multirow}
\usepackage{array}
\usepackage{colortbl}
\usepackage{wrapfig}
\usepackage{float}
\usepackage{url}
\usepackage{hyperref}
\usepackage[capitalize,noabbrev]{cleveref}
\usepackage{titletoc}

\usepackage{amsmath,amsfonts,bm}

\def\eqref#1{equation~\ref{#1}}

\def\1{\bm{1}}

\DeclareMathAlphabet{\mathsfit}{\encodingdefault}{\sfdefault}{m}{sl}
\SetMathAlphabet{\mathsfit}{bold}{\encodingdefault}{\sfdefault}{bx}{n}

\newcommand\ie{\textit{i.e.,}}
\newcommand\eg{\textit{e.g.,}}

\newcommand\etc{\textit{etc.}}

\newcommand{\beq}{\begin{equation}}
\newcommand{\eeq}{\end{equation}}
\newcommand{\beqnn}{\begin{equation*}}
\newcommand{\eeqnn}{\end{equation*}}
\newcommand{\beqy}{\begin{eqnarray}}
\newcommand{\eeqy}{\end{eqnarray}}
\newcommand{\beqynn}{\begin{eqnarray*}}
\newcommand{\eeqynn}{\end{eqnarray*}}
\newcommand{\bit}{\begin{itemize}}
\newcommand{\eit}{\end{itemize}}
\newcommand{\ben}{\begin{enumerate}}
\newcommand{\een}{\end{enumerate}}
\newcommand{\bex}{\begin{example}}
\newcommand{\eex}{\end{example}}

\newcommand{\balg}[1]{\begin{algorithm} \caption{#1}}
\newcommand{\ealg}{\end{algorithm}}

\newcommand{\balgc}{\begin{algorithmic}[1]}
\newcommand{\ealgc}{\end{algorithmic}}

\newcommand{\bary}{\begin{array}}
\newcommand{\eary}{\end{array}}
\newcommand{\bmx}{\begin{bmatrix}}
\newcommand{\emx}{\end{bmatrix}}
\newcommand{\bsmx}{\left[\begin{smallmatrix}}
\newcommand{\esmx}{\end{smallmatrix}\right]}
\newcommand{\bmxc}[1]{\left[\begin{array}{@{}#1@{}}}
\newcommand{\emxc}{\end{array}\right]}
\newcommand{\bcn}{\begin{center}}
\newcommand{\ecn}{\end{center}}

\theoremstyle{plain}

\usepackage[textsize=tiny]{todonotes}

\title{RL Fine-Tuning Heals OOD Forgetting in SFT}

\author{%
\begin{tabular}{c}
Hangzhan Jin$^{1,2,\dagger,*}$
\quad
Sitao Luan$^{1,3,\dagger}$
\quad
Tianwei Ni$^{1,3}$
\quad
Sicheng Lyu$^{1,4}$\\
Guillaume Rabusseau$^{1,3,5}$
Reihaneh Rabbany$^{1,4,5}$
Doina Precup$^{1,4,5,6}$
Mohammad Hamdaqa$^{2}$\\[0.35em]
\normalfont\footnotesize
$^{1}$Mila - Quebec AI Institute
\quad
$^{2}$Polytechnique Montr\'eal
\quad
$^{3}$Universit\'e de Montr\'eal\\[-0.1em]
\normalfont\footnotesize
$^{4}$McGill University
\quad
$^{5}$CIFAR AI Chair
\quad
$^{6}$Google DeepMind
\end{tabular}
}

\begin{document}

\maketitle

\begingroup
\renewcommand{\thefootnote}{}
\footnotetext{
$^{\dagger}$Equal contribution.
\quad
$^{*}$Corresponding author: \texttt{hangzhan.jin@mila.quebec}.
}
\endgroup
\vspace{-0.5cm}

\looseness=-1
\begin{abstract}

Supervised Fine-Tuning (SFT) followed by Reinforcement Learning (RL) is a standard post-training recipe for improving Large Language Models (LLM) reasoning, but why it works remains unclear. We revisit the common claim that ``SFT memorizes, RL generalizes'' through checkpoint-wise analyses of in-distribution (ID) and out-of-distribution (OOD) reasoning. We find that OOD performance often peaks early during SFT and then declines despite continued improvement in ID reasoning. RL typically does not surpass this early SFT peak; rather, it restores OOD capability lost during later SFT, and only from a bounded range of SFT checkpoints. Further spectral analysis shows that this forgetting-and-recovery pattern correlates with rotations of singular vectors, while singular values remain largely stable. These findings suggest a more precise view of post-training dynamics: SFT can forget, RL can recover, and controlling singular-vector rotation may improve OOD robustness. Code is available at \href{https://github.com/jinhangzhan/RL\_Heals\_SFT.git}{https://github.com/jinhangzhan/RL\_Heals\_SFT}.

\end{abstract}
\section{Introduction}
\label{sec:introduction}
Supervised Fine-Tuning (SFT) is the most widely used method for the post-training of Large Language Models (LLMs)~\citep{howard2018universal, radford2018improving}. Recent work demonstrates that Reinforcement Learning Fine-Tuning (RLFT), especially when applied after SFT~\citep{deepseekai2025deepseekr1incentivizingreasoningcapability}, can achieve much better performance on complex reasoning tasks, such as symbolic math reasoning~\citep{deepseekai2025deepseekr1incentivizingreasoningcapability, xaigrok2025}, code generation~\citep{mirzadeh2024gsm,jiang2024survey, anthropicclaude2025},  embodied tasks~\citep{lin2025evolvenav, li2025perception,zhao2021consciousness}, video prediction~\citep{shi2025enhancing}, \etc{} 
Such a two-stage fine-tuning paradigm has rapidly become popular because of its advantages over the one-stage SFT~\citep{openr1, wang2025reinforcement}. 

\looseness=-1
Prior studies have explored how RL helps SFT in post-training: a growing body of work argues that SFT tends to memorize or overfit the training distribution, whereas RL yields better out-of-distribution (OOD) generalization~\citep{kirk2023understanding, chu2025sftmemorizesrlgeneralizes}; others emphasize that KL-regularized RL counteracts SFT's drift from the base model~\citep{fu2025srft}, and that rule-based or structure-aware RL can significantly strengthen reasoning~\citep{xie2025logicrlunleashingllmreasoning}. The authors in~\citep{xie2025logicrlunleashingllmreasoning} note that SFT pulls the policy of a model away from its base initialization, and specific RL recipes can boost reasoning. These empirical findings help to partially explore the high-level picture of two-stage fine-tuning, however, the understanding on the synergy of SFT and RL is still inconclusive. In particular, the evolution of OOD performance during the two-stage fine-tuning lacks a deep investigation. 

To fill the gaps in the above issues, we perform full-parameter SFT and RLFT and study the Out-Of-Distribution (OOD) and In-Distribution (ID) reasoning behaviors of two popular open-sourced models: LLaMA-3.2-11B-Vision~\citep{grattafiori2024LLaMA} and Qwen-2.5-7B~\citep{qwen2.5}. Specifically, we track their ID and OOD performance at different checkpoints on various reasoning tasks, including the \textit{GeneralPoints, Navigation and Rank-Determinant Computation} tasks. 
These controlled environments allow us to monitor and disentangle the evolution of model performance and investigate the roles of SFT and RL in the whole process.

During fine-tuning, we observe that: (1) OOD reasoning performance \textbf{peaks rapidly in the very early stage of SFT} and then degrades slowly as SFT continues. Such \textbf{OOD forgetting} is hard to capture by the traditional overfitting detection methods, as the learning curves for ID training/test loss will continue to decline. (2) \textbf{RL does not uniformly improve OOD reasoning from arbitrary SFT checkpoints}. It can recover OOD forgetting in SFT but barely surpasses the peak OOD performance already reached by earlier SFT. The recovery is only effective within a certain range of SFT checkpoints and we identify the shape of advantage distribution as the main cause of it.

To uncover the underlying factors that have high impacts on the fine-tuned models, we analyze the Singular-Value Decomposition (SVD) of parameter matrices and conduct ablation studies on their influence to model performance. Unlike some recent studies~\citep{bartlett2017spectrallynormalizedmarginboundsneural, yoshida2017spectralnormregularizationimproving, li2024setaroutofdistributiondetectionselective}, in our experiments, we notice that the singular values remain essentially constant throughout both SFT and RL stages. Instead, OOD forgetting and recovery highly correlate with the rotations of the singular vectors. In addition, we provide fine-grained layer-wise and top-$k$ analysis on the singular values/vectors and motivate a protected SFT variant that orthogonalizes updates against top-rank OOD-related singular-vector directions.

\section{Preliminaries}
\label{sec:preliminaries}
\paragraph{Supervised Fine-Tuning (SFT).}
SFT adapts a pre-trained model $\pi_\theta$ to a specific task using a labeled dataset $\mathcal{D} = \{(x, y)\}$~\citep{howard2018universal, radford2018improving}. The standard objective is to minimize the negative log-likelihood (NLL) of the target outputs given the inputs:
\begin{align}
    \mathcal{L}_{\text{SFT}}(\theta) = -\sum_{(x, y) \in \mathcal{D}} \log \pi_\theta(y \mid x)
\end{align}

\paragraph{Reinforcement Learning (RL) Fine-Tuning}
In contrast to SFT, RL essentially fine-tunes the model by optimizing the policy $\pi_\theta$ based on a reward signal $R(\cdot)$. The general objective is to maximize the expected reward of the response generated by the model,
$
    \max_\theta \mathbb{E}_{y \sim \pi_\theta(\cdot \mid x)}[R(x, y)].
$
The reward function $R(x,y)$ evaluates the quality of the generated response $y$ for input $x$ based on desired attributes, like correctness~\citep{ouyang2022training}, clarity~\citep{wang2023pandalm}, or adherence to rules~\citep{bai2022constitutional}. In this paper, we employ {Proximal Policy Optimization (PPO)}~\citep{schulman2017ppo}, a popular RL algorithm that stabilizes training by optimizing a clipped surrogate objective. The PPO objective is:
\begin{equation}
\label{eq:ppo}
     \mathcal{L}_{\text{PPO}}(\theta) = \mathbb{E}_{t}\left[\min\left(r_t(\theta) A_t, \text{clip}(r_t(\theta), 1 - \epsilon, 1 + \epsilon) A_t\right)\right],
\end{equation}
where $r_t(\theta) = \frac{\pi_\theta(y_t \mid x, y_{<t})}{\pi_{\theta_{\text{old}}}(y_t \mid x, y_{<t})}$ is the probability ratio for state $(x,y_{<t})$ and action $y_t$ at step $t$, $A_t$ is the advantage estimate, and $\epsilon$ is a hyperparameter that constrains the policy update step to avoid excessive shift of policy. 
The advantage $A_t$ measures how much better (or worse) taking action $y_t$ in state $(x,y_{<t})$ is compared to the average action at that state, as estimated by a value function $V_\phi(x,y_{<t})$. Intuitively, $A_t$ is positive when an action yields higher return than expected and negative otherwise. \looseness=-1

\section{Evaluation and Analysis}
\label{sec:evaluation_analysis}
In this section, we investigate the evolution of OOD reasoning ability of LLMs by analyzing the model performance at different checkpoints in SFT and RL stages. 
More specifically, in Section~\ref{sec:evaluation_settings}, we introduce the experimental settings, including the tasks, models and evaluation methods. In Section~\ref{sec:results_analysis}, we present the results and conduct detailed analysis on ID and OOD reasoning performance.

\subsection{Evaluation Settings}
\label{sec:evaluation_settings}
\paragraph{Task Description}
We use \textit{GeneralPoints, Navigation}~\citep{chu2025sftmemorizesrlgeneralizes} and  \textit{Rank-Determinant Computation}~\citep{sun2025omega}~\footnotemark[1] to evaluate the arithmetic, spatial and cross-concept math reasoning abilities of models. The detailed task descriptions and prompts are shown in Appendix~\ref{appendix:task_description} and we only use \textit{GeneralPoints} \textbf{as an example} in main paper. The \textit{GeneralPoints} environment~\citep{chu2025sftmemorizesrlgeneralizes} is instantiated on top of the \textit{Points24} environment~\citep{zhai2024fine}. Each state $s$ contains four poker cards, described in text directly. The goal is to produce an equation that equals a target number ($24$ by default), with four numbers from the cards used only once. Particularly, the cards $'J, Q, K'$ are all interpreted as the same number $10$ in the original setting (for training). For example, provided with four cards $[5,4,10,7]$, we aim to output the equation \textit{(7-5)*10+4} as the desired output. 

\paragraph{Evaluation of OOD Generalization} To disentangle the evaluation of superficial format learning and real arithmetic reasoning ability, we modify the rule of \textit{GeneralPoints} as~\citep{chu2025sftmemorizesrlgeneralizes} and test both ID and OOD reasoning performance of models. Specifically, instead of interpreting $'J, Q, K'$ all as the same number $10$, the new rule interprets them as $11, 12,$ and $13$, respectively. If the model can really obtain arithmetic reasoning ability, they should perform well on such OOD settings.  We record the model performance at different checkpoints to show how the ID and OOD generalization abilities evolve. See the ID and OOD evaluation setups of other tasks in Appendix~\ref{appendix:task_description}. 

\paragraph{Why Controlled and Public Evaluations Are Complementary.}
Controlled tasks serve primarily as diagnostic environments, not as the only evidence for broad reasoning generalization. Their ID/OOD shifts are explicit by construction, rules are known, and dense checkpoint-wise SFT/RL sweeps are feasible, allowing us to separate superficial format alignment from reasoning-rule transfer and reduce ambiguity from pre-training contamination. Since controlled tasks do not fully represent open-ended reasoning, we additionally evaluate a broader math/reasoning pipeline using Open-R1 SFT~\citep{openr1} followed by DAPO-math RL with DAPO~\citep{yu2025dapoopensourcellmreinforcement}; these public benchmarks test whether the same qualitative SFT-forgetting/RL-recovery pattern persists beyond controlled environments, as shown in Section~\ref{sec:broader}.

\paragraph{Models and Setup} 
We use two widely used open-source base models, LLaMA-3.2-11B-Vision~\citep{grattafiori2024LLaMA} and Qwen-2.5-7B~\citep{qwen2.5}, as the base models. Following the commonly used two-stage pipeline for post-training~\citep{deepseekai2025deepseekr1incentivizingreasoningcapability}, we first warm-up the model with SFT, and then run RL on top of SFT checkpoint. The format of the prompt is the same as~\citep{chu2025sftmemorizesrlgeneralizes}. For \textit{GeneralPoints}, we follow the setup in~\citep{chu2025sftmemorizesrlgeneralizes} as our standard setting. We denote the checkpoints when SFT and RL end as $\text{SFT}_\text{End}$ and $\text{RL}_\text{End}$. Besides the standard setting, to track the impact of RL on the SFT model more carefully, we apply RL at different SFT checkpoints \{0, 90, 140, 200, 300, 400, \dots, 1600\}, and evaluate the ID and OOD reasoning performance before and after RL. See the computational resources and setups in Appendix~\ref{appendix:computational_resources_setups}. We use PPO in the main paper, and see the results of GRPO in Appendix~\ref{appendix:results_grpo}.  


\begin{figure*}[h]
  \centering
    \centering
    \includegraphics[width=\linewidth]
    {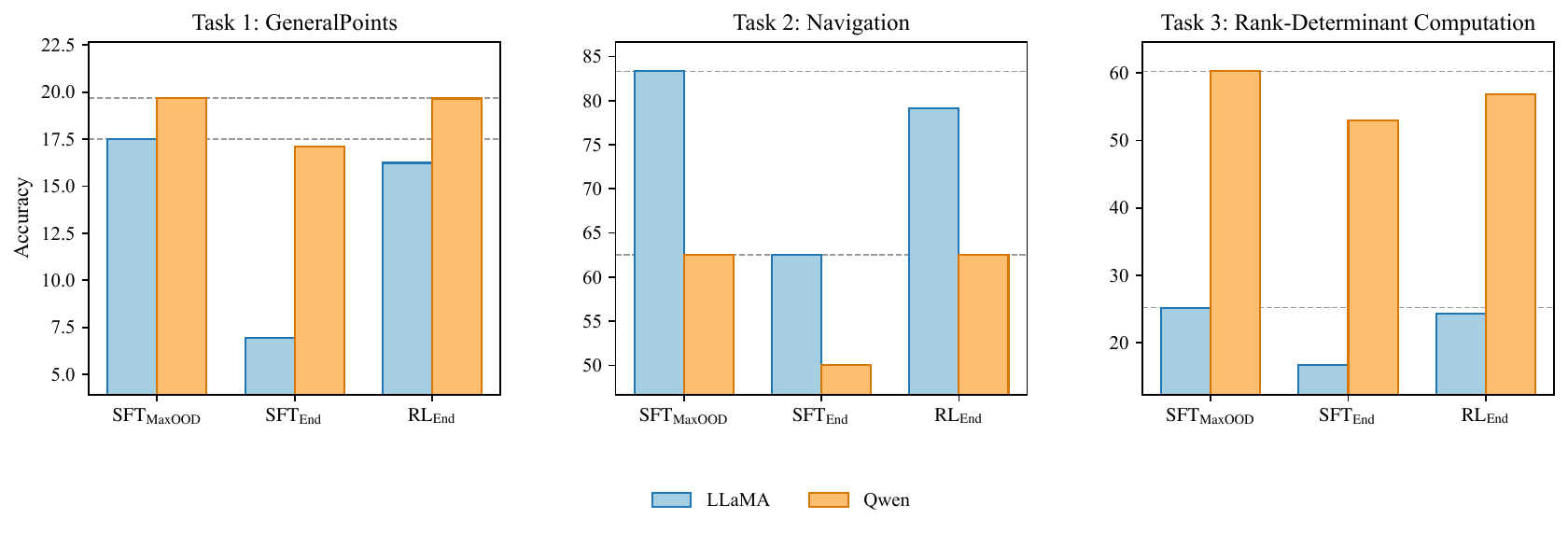}
  \caption{Comparison of OOD performance in three tasks for LLaMA and Qwen at different checkpoints ($\text{SFT}_\text{MaxOOD}, \text{SFT}_\text{End}$ and $\text{RL}_\text{End}$).}
  \label{fig:gp_ood_acc}
\end{figure*}
\subsection{Results and Analysis}
\label{sec:results_analysis}
\begin{figure*}[!t]
  \begin{subfigure}[t]{0.49\linewidth}
    \centering
    \includegraphics[width=\linewidth,height=0.2\textheight,keepaspectratio]{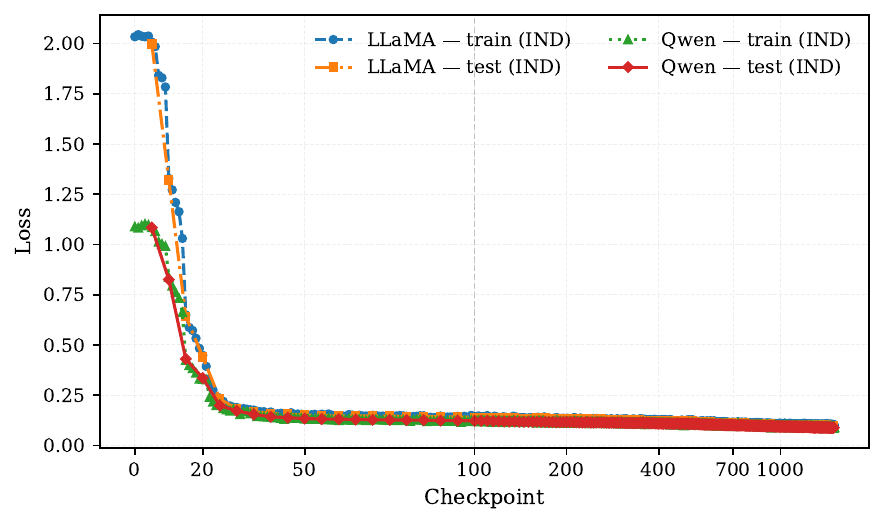}
    
     \caption{In-distribution training and test loss}
    \label{fig:sft_training_test_loss}
  \end{subfigure}
  \begin{subfigure}[t]{0.49\linewidth}
    \centering
    \includegraphics[width=\linewidth,height=0.2\textheight,keepaspectratio]{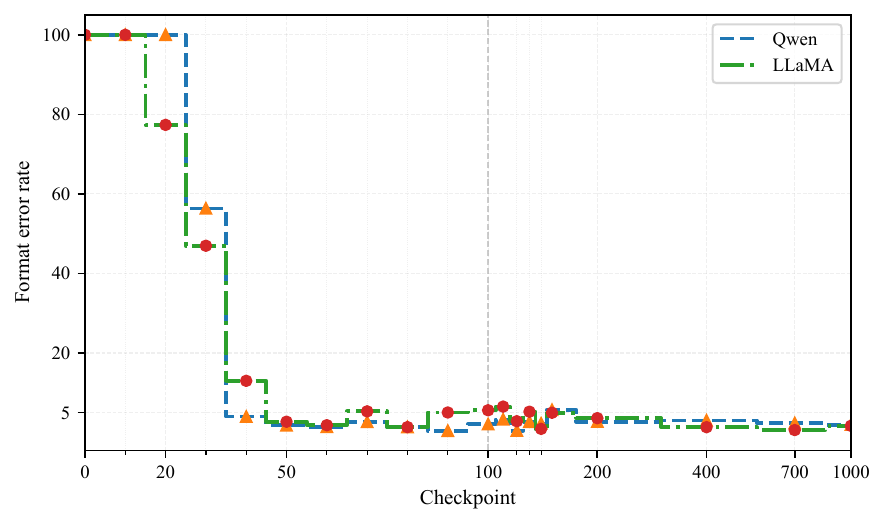}
     \caption{Format error}
    \label{fig:sft_format_loss}
  \end{subfigure}
    \hfill
  \begin{subfigure}[t]{0.49\linewidth}
    \centering
    \includegraphics[width=\linewidth,height=0.2\textheight,keepaspectratio]{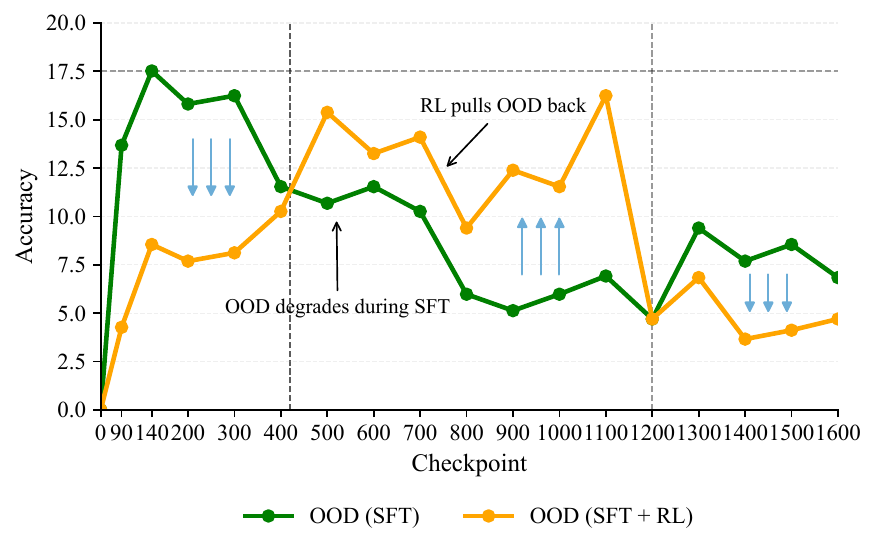}
     \caption{Evolution of OOD test accuracy.}
    \label{fig:rl_sft_acc_ood_evolution}
  \end{subfigure}
  \begin{subfigure}[t]{0.49\linewidth}
    \centering
    \includegraphics[width=\linewidth,height=0.2\textheight,keepaspectratio]{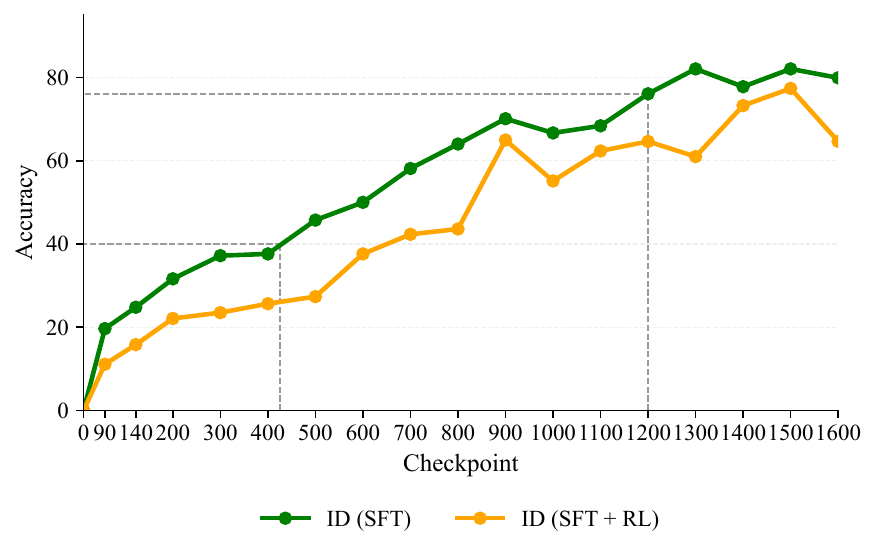}
     \caption{Evolution of ID test accuracy.}
    \label{fig:rl_sft_acc_id_evolution}
  \end{subfigure}
    \hfill                           
  \caption{(\ref{fig:sft_training_test_loss}) Training and test loss, and (\ref{fig:sft_format_loss}) format error curves during SFT. Evolution of (\ref{fig:rl_sft_acc_ood_evolution}) OOD and (\ref{fig:rl_sft_acc_id_evolution}) ID test accuracy of SFT and RL at different checkpoints (take LLaMA as the main example).
  } 
  \label{fig:rl_sft_acc_evolution}
\end{figure*}

\paragraph{What Is Missing in "SFT memorizes, RL generalizes"?} It has recently been found that, in the two-stage fine-tuning pipeline, SFT can stabilize the model output before RL, and RL can enhance the OOD generalization capability of the SFT model~\citep{chu2025sftmemorizesrlgeneralizes}. It highlights the complementary roles of SFT and RL, and the claim "SFT memorizes, RL generalizes" has become popular. As shown in Figure~\ref{fig:gp_ood_acc}, we reproduce the results in~\citep{chu2025sftmemorizesrlgeneralizes}, where the RL fine-tuned models at $\text{RL}_\text{End}$ significantly outperform models at the checkpoint $\text{SFT}_\text{End}$ on three different tasks. However, when tracking the evolution of OOD performance in the whole SFT process, we can always find a checkpoint where the SFT models outperform the RLFT models. This indicates that the conclusion that RL can enhance the OOD reasoning capacity of SFT model is over-simplified and the best overall OOD performance has already been achieved at an earlier SFT checkpoint. We denote this checkpoint as $\text{SFT}_\text{MaxOOD}$. However, $\text{SFT}_\text{MaxOOD}$ is hard to capture only based on ID training/test losses as shown in Figure~\ref{fig:sft_training_test_loss}. People still tend to manually set up a terminal checkpoint $\text{SFT}_\text{End}$ and then do RL.

LLM keeps losing OOD capability from $\text{SFT}_\text{MaxOOD}$ to $\text{SFT}_\text{End}$. The claim "SFT memorizes, RL generalizes" made in~\citep{chu2025sftmemorizesrlgeneralizes} is only based on the observations that, starting from $\text{SFT}_\text{End}$, the continued RLFT model is better than SFT model. However, $\text{SFT}_\text{End}$  already suffers from severe OOD forgetting. Therefore, the evidence at a single fixed checkpoint is insufficient to provide a comprehensive and strict comparison between SFT and RL. Their claim reflects only one aspect of a broader picture, where RL recovers the degradation in $\text{SFT}_\text{End}$, but barely surpasses the best of SFT. To depict the whole story and verify our new claim, we track the OOD performance at various SFT checkpoints and apply RLFT. We use LLaMA on \textit{GeneralPoints} as the dense-sweep example, and use the Qwen/GRPO and Open-R1/DAPO experiments in Appendix Figure~\ref{fig:qwen_ood_recovery_boundary_appendix}, Tables~\ref{tab:qwen_grpo_boundary_full} and~\ref{tab:dapo_complex_boundary} to test the same boundary pattern beyond this main sweep. Our observations of the whole fine-tuning process are as follows.

\paragraph{SFT forgets.} 

The training loss and ID test loss during SFT are shown in Figure~\ref{fig:sft_training_test_loss}, the format loss is shown in Figure~\ref{fig:sft_format_loss}, and the OOD and ID test accuracy curve (take LLaMA as an example) is shown in Figure~\ref{fig:rl_sft_acc_ood_evolution} and ~\ref{fig:rl_sft_acc_id_evolution}. As shown in Figure~\ref{fig:sft_format_loss}, the format loss converges at checkpoint $50$ and stays almost unchanged afterwards, which means the model completes format alignment at $\text{SFT}_{50}$. During $50$ to 140 checkpoints, the performance gain in OOD reasoning is mainly from the improved arithmetic reasoning ability. As shown in Figure~\ref{fig:rl_sft_acc_ood_evolution}, the OOD test accuracy declines after $\text{SFT}_\text{MaxOOD}$, although the training loss and ID test loss continue to decrease. This performance divergence indicates that the model starts to focus too much on adapting to the rules of the target game, instead of really learning the arithmetic reasoning ability. Such over-specialization causes the model to forget the acquired OOD reasoning ability. Note that we \textbf{do not have an overfitting problem} here, because ID test loss keeps decreasing and ID test accuracy continues to increase. However, in this situation, we still keep losing the OOD reasoning ability, and we call such a phenomenon \textbf{OOD forgetting}.

\begin{figure*}[!t]
  \begin{subfigure}[t]{0.245\linewidth}
    \centering
    \includegraphics[width=\linewidth]{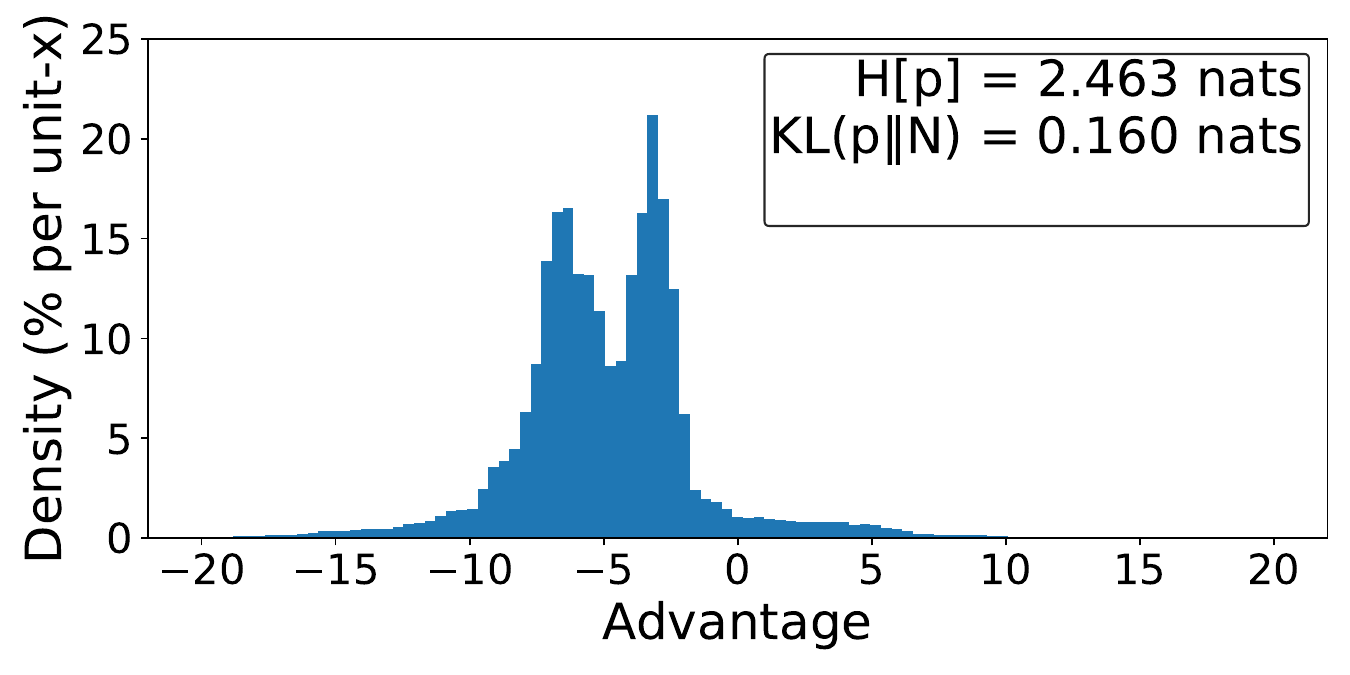}
    \caption{Checkpoint 140}
    \label{fig:rl_adv_density_140}
  \end{subfigure}                        
  \begin{subfigure}[t]{0.245\linewidth}
    \centering
    \includegraphics[width=\linewidth]{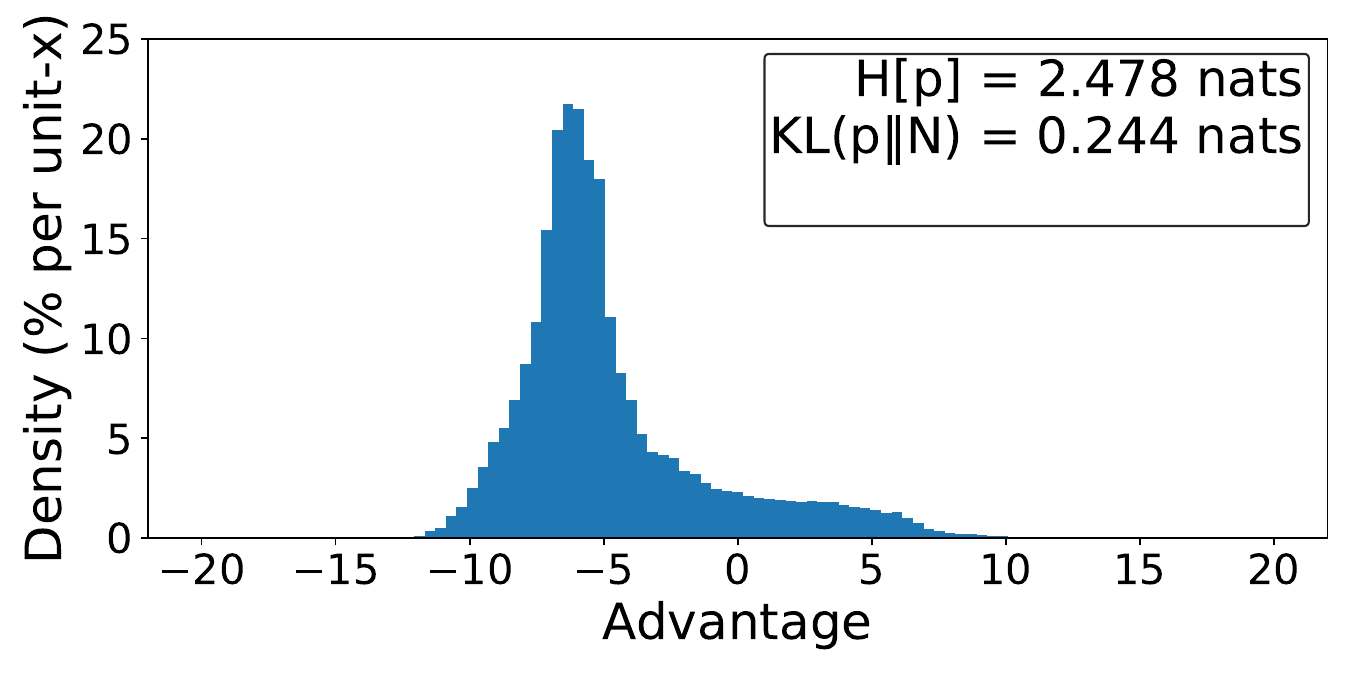}
    \caption{Checkpoint 400}
    \label{fig:rl_adv_density_400}
  \end{subfigure}
   \begin{subfigure}[t]{0.245\linewidth}
    \centering
    \includegraphics[width=\linewidth]{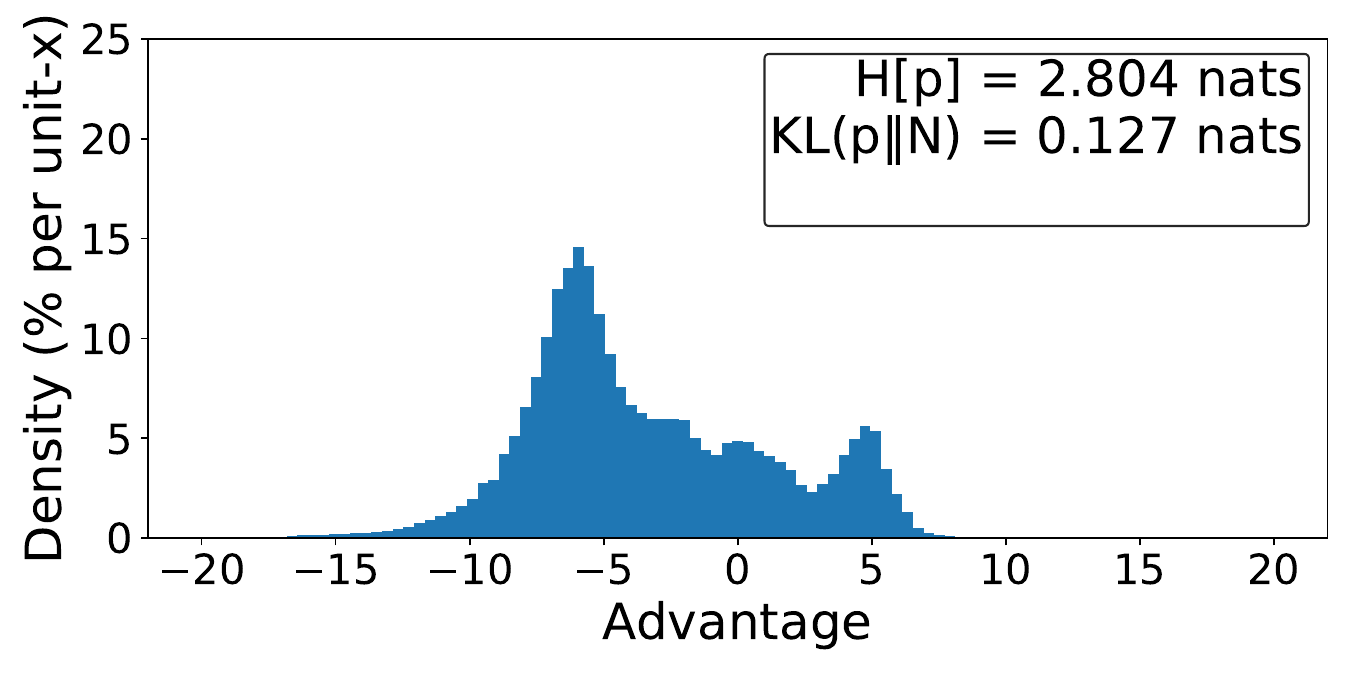}
    \caption{Checkpoint 600}
    \label{fig:rl_adv_density_600}
  \end{subfigure}                      
  \begin{subfigure}[t]{0.245\linewidth}
    \centering
    \includegraphics[width=\linewidth]{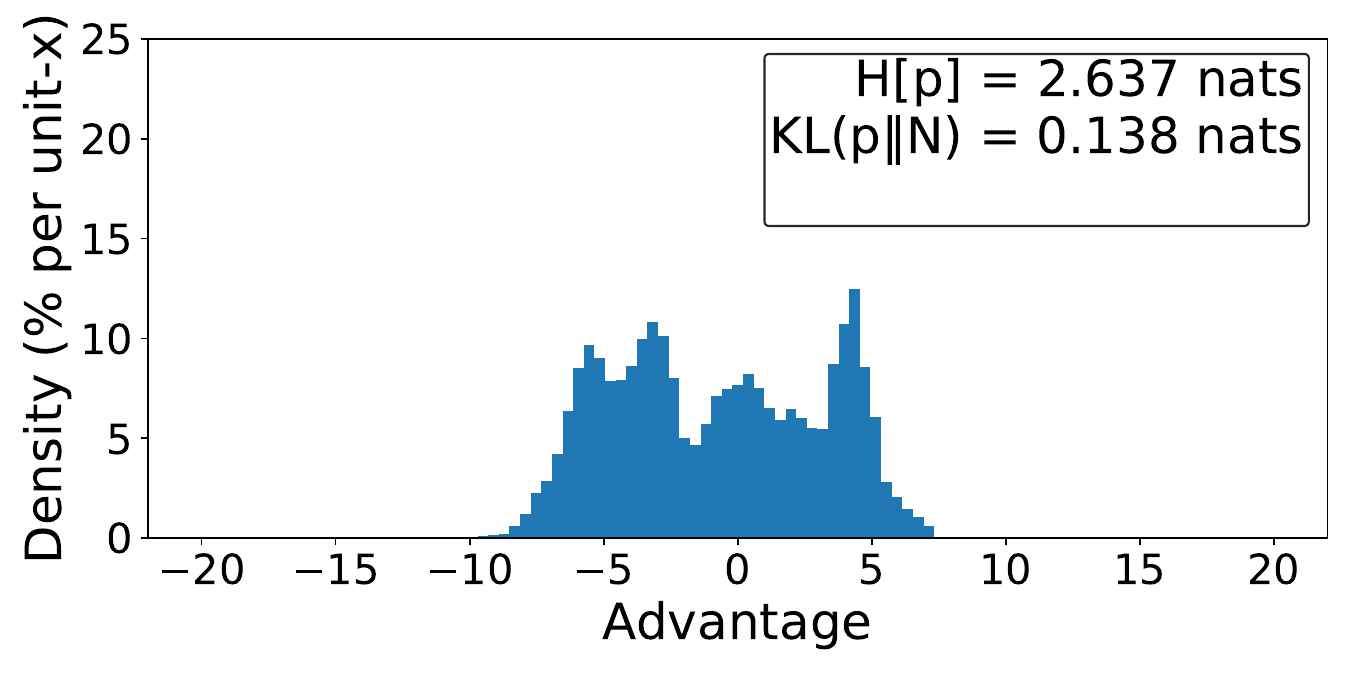}
    \caption{Checkpoint 800}
    \label{fig:rl_adv_density_800}
  \end{subfigure}                        
  \hfill
   \begin{subfigure}[t]{0.245\linewidth}
    \centering
    \includegraphics[width=\linewidth]{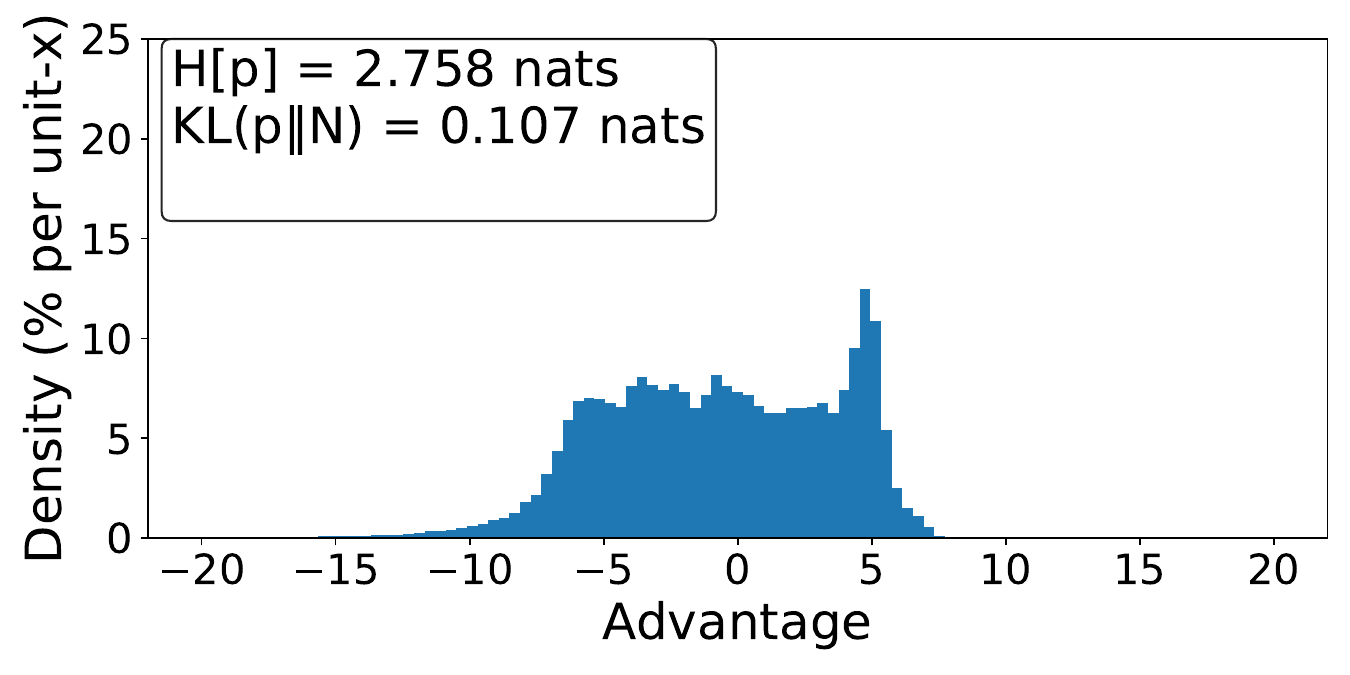}
    \caption{Checkpoint 1000}
    \label{fig:rl_adv_density_1000}
  \end{subfigure}                        
   \begin{subfigure}[t]{0.245\linewidth}
    \centering
    \includegraphics[width=\linewidth]{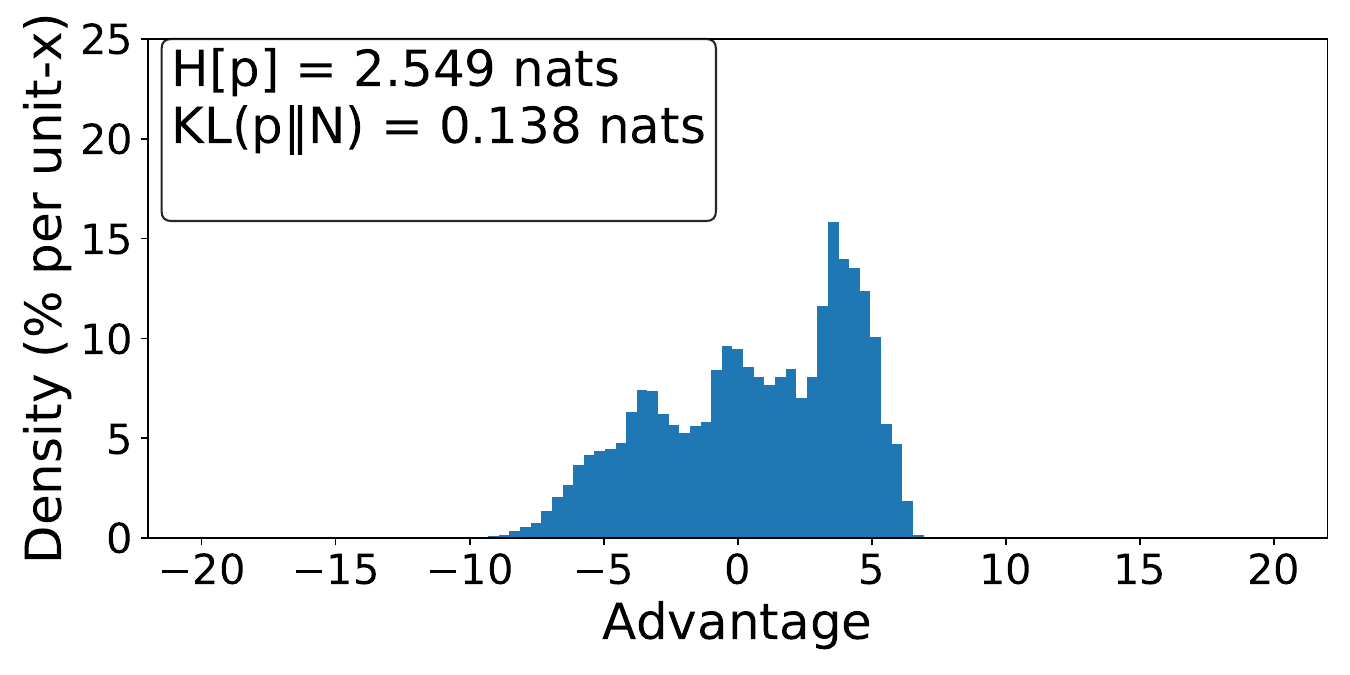}
    \caption{Checkpoint 1200}
    \label{fig:rl_adv_density_1200}
  \end{subfigure}                        
  \begin{subfigure}[t]{0.245\linewidth}
    \centering
    \includegraphics[width=\linewidth]{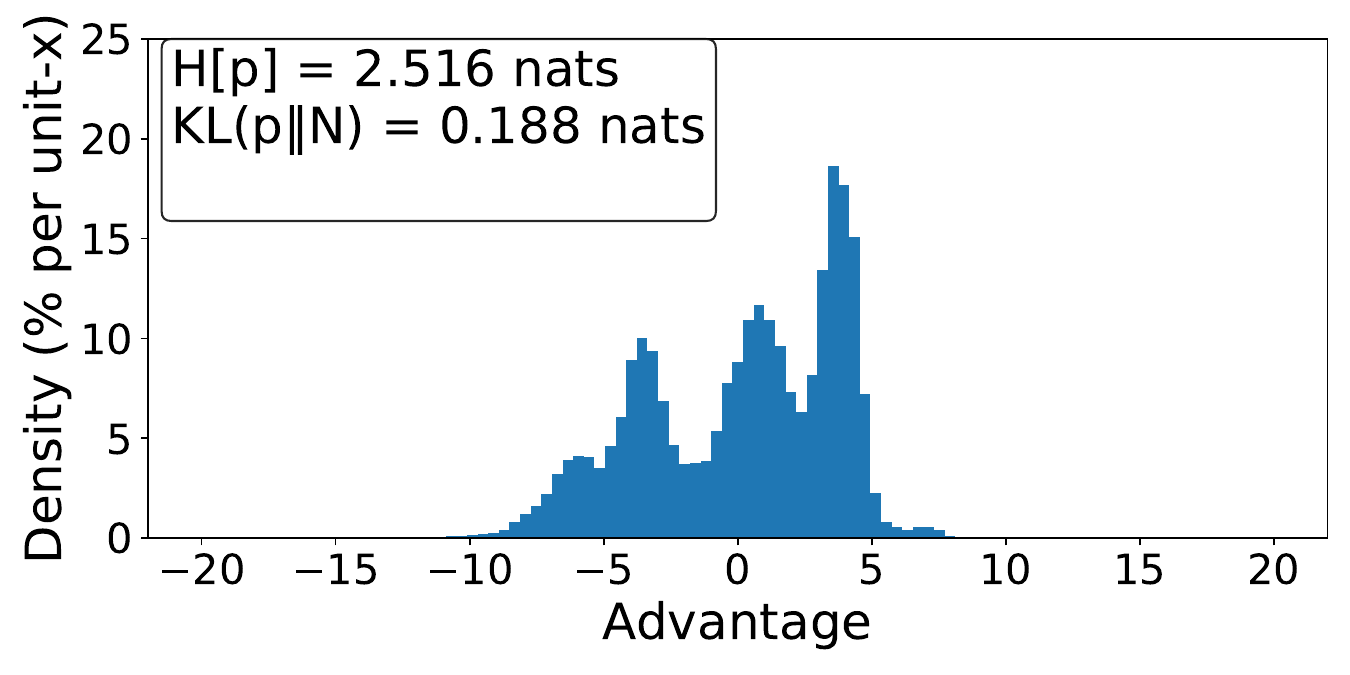}
    \caption{Checkpoint 1600}
    \label{fig:rl_adv_density_1600}
  \end{subfigure}
    \begin{subfigure}[t]{0.245\linewidth}
    \centering
    \includegraphics[width=\linewidth]{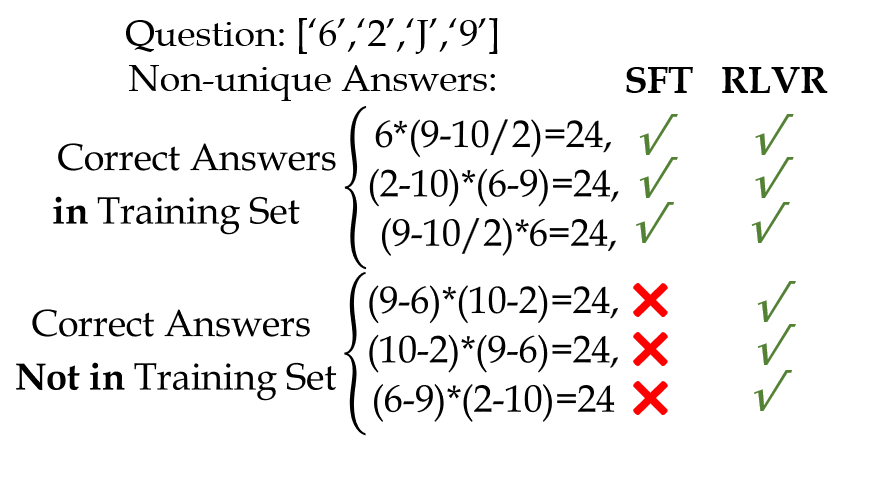}
    \caption{Non-unique Solution}
    \label{fig:demo_nonunique_solution}
  \end{subfigure}
  \caption{Advantage estimation distribution (a-g) and demonstration of questions with non-unique solutions (h).}
  \label{fig:rl_fig:rl_adv_density}
\end{figure*}

\paragraph{RL recovers.} As shown in Figure~\ref{fig:rl_sft_acc_ood_evolution}, there exists an interval of SFT checkpoints where the RL curve is higher than the corresponding SFT curve, indicating that RL can recover part of the OOD ability lost during SFT, with some trade-off against ID specialization as shown in Figure~\ref{fig:rl_sft_acc_id_evolution}. However, in the settings we study, RL generally does not surpass the best earlier SFT/OOD checkpoint $\text{SFT}_{\text{MaxOOD}}$. Thus, rather than treating RL as a general source of new OOD capability, our evidence suggests that RL primarily acts as an OOD restorer when initialized from a bounded SFT checkpoint.

Interestingly, there exists \textbf{a clear boundary for the recovery effect of RL}, \ie{} RL can only restore the lost OOD capability in SFT within checkpoint $[420, 1200]$. The reason is that PPO needs a balanced ratio of positive vs. negative reward signals to be trained stably and effectively. Highly skewed reward distribution can lead to high variance in advantage estimates, poor exploration, and unstable policy updates. Empirically, the proper ratio of positive reward in our experiment can be roughly estimated by the ID accuracy of SFT model as shown in Figure~\ref{fig:rl_sft_acc_id_evolution}, \ie around $[40\%, 80\%]$. Additional sweeps in Appendix Tables~\ref{tab:qwen_grpo_boundary_full} and~\ref{tab:dapo_complex_boundary} show the same intermediate-checkpoint recovery pattern for Qwen/GRPO and broader Open-R1 SFT followed by DAPO-math RL. The cold-start evidence in Appendix~\ref{appendix:cold_start_rl_evidence}, including Table~\ref{tab:qwen_cold_start_rl}, further shows that RL alone can hurt OOD metrics when the policy lacks enough task competence, supporting our view that RL is not a general source of new OOD capability.


\paragraph{Advantage Distribution} To understand the RL recovery boundary, we analyze advantage distributions across SFT initialization checkpoints. We summarize each distribution by its center, standard deviation, skewness, entropy, and KL divergence to a matched normal distribution; detailed definitions are moved to Appendix~\ref{appendix:advantage_metric_definitions}.

The results are demonstrated in Figure~\ref{fig:rl_fig:rl_adv_density}(a-g) (See full results in Appendix~\ref{Appendix:advantage_comparison}). Through the comparison of the statistics, we observe that, for the checkpoints within the effective boundaries of RL \textbf{(1)} the centers do not significantly deviate from 0; \textbf{(2)} empirically, the effective checkpoints have entropy larger than $2.55$ and KL divergence against the matched normal distribution smaller than $0.16$; these statistics indicate that RL recovery requires advantage signals that are sufficiently balanced, less spiky, and structured; \textbf{(3)} moderate skewness and multi-modality are acceptable.

Note that our observations of the boundary also echo some empirical observations in recent studies~\citep{liu2025uft, wang2025reinforcement} that we need the base model to be strong enough \citep{foster2025learningreasonfrontierlearnability} for RL to be effective, this is because a base policy with only weak prior knowledge cannot gain positive rewards as a learning signal during reinforcement fine-tuning; on the other hand, too much SFT will lead to policy entropy collapse and hurt exploration and learning capability~\citep{lanchantin2025bridging, kang2024learningdynamicsrevealgeneralization}. 

\paragraph{Verifiable Reward Shines in Problems With Non-Unique Solutions} RL recovers the OOD ability lost in SFT by replacing token-level imitation signals with outcome-level verifiable rewards, especially for questions with non-unique solutions. We demonstrate it in Figure~\ref{fig:demo_nonunique_solution} and the example in Appendix~\ref{appendix:task_description}. As shown in the "number" and "formula" steps, multiple correct formulas can be derived based on the same set of question numbers. However, the token-level cross-entropy loss in SFT will only give "positive reward" to the correct answers that exist in the training data. For other newly explored correct solutions, it will give "negative rewards", which provides incorrect gradient directions and leads to high perplexity on them. This is pronounced on reasoning tasks with multiple answers. Therefore, as long as RL can stably work within the boundary, it heals the OOD forgetting. 

\subsection{Evidence Beyond Controlled Tasks}
\label{sec:broader}

\begin{figure*}[h]
  \centering
  \includegraphics[width=0.72\textwidth]{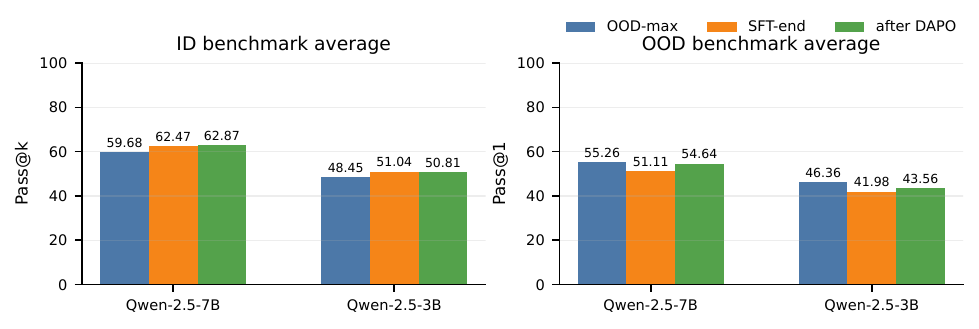}
  \caption{SFT on Open-R1, followed by RL on DAPO-math. Unless noted otherwise, accuracy and rate values in the paper and appendix are reported on a 0--100 scale. ID uses pass@$k$ averages, while OOD uses pass@1 averages.}
  \label{fig:broader_dapo_summary_main}
\end{figure*}

\begin{table*}[h]
\centering
\footnotesize
\setlength{\tabcolsep}{2.5pt}
\resizebox{\textwidth}{!}{%
\begin{tabular}{@{}llrrrrrrr|rrrrrrr@{}}
\toprule
& & \multicolumn{7}{c}{\textbf{ID math/reasoning pass@$k$}} & \multicolumn{7}{c}{\textbf{OOD benchmark pass@1}} \\
\cmidrule(lr){3-9}\cmidrule(l){10-16}
\textbf{Model} & \textbf{Setting} & \textbf{AIME24} & \textbf{AIME25} & \textbf{AMC} & \textbf{MATH} & \textbf{Minerva} & \textbf{Olympiad} & \textbf{Avg} & \textbf{GPQA} & \textbf{IFEval} & \textbf{MMLU-Pro} & \textbf{SuperGPQA} & \textbf{Safety} & \textbf{TruthfulQA} & \textbf{Avg} \\
\midrule
Qwen-2.5-7B & OOD-max    & 36.67 & 36.67 & 90.00 & 84.60 & 55.17 & 54.96 & 59.68 & 33.93 & 61.37 & 67.14 & 27.65 & 74.50 & 66.96 & 55.26 \\
Qwen-2.5-7B & SFT-end    & 50.00 & 36.67 & 87.50 & 87.60 & 58.09 & 54.96 & 62.47 & 32.80 & 52.13 & 65.71 & 26.42 & 70.10 & 59.50 & 51.11 \\
Qwen-2.5-7B & after DAPO & 46.67 & 33.33 & 90.00 & 89.60 & 58.09 & 59.55 & 62.87 & 33.92 & 52.31 & 71.43 & 32.10 & 72.90 & 65.20 & 54.64 \\
\midrule
Qwen-2.5-3B & OOD-max    & 20.00 & 20.00 & 80.00 & 79.20 & 50.00 & 41.48 & 48.45 & 27.01 & 54.16 & 52.86 & 18.27 & 65.90 & 59.94 & 46.36 \\
Qwen-2.5-3B & SFT-end    & 30.00 & 23.33 & 82.50 & 80.60 & 47.42 & 42.37 & 51.04 & 30.58 & 47.32 & 45.71 & 19.26 & 59.60 & 49.42 & 41.98 \\
Qwen-2.5-3B & after DAPO & 23.33 & 33.33 & 75.00 & 81.00 & 47.79 & 44.44 & 50.82 & 27.68 & 48.43 & 47.14 & 22.72 & 62.50 & 52.92 & 43.57 \\
\bottomrule
\end{tabular}%
}
\caption{SFT on Open-R1, followed by RL on DAPO-math. ID columns report pass@$k$: AIME24/AIME25/AMC use pass@16, MATH-500/Olympiad use pass@4, and Minerva uses pass@8. OOD columns report pass@1.}
\label{tab:broader_dapo_passk_main}
\end{table*}

We further test whether the same qualitative behavior appears beyond controlled diagnostic tasks. In addition to the checkpoint-wise sweeps above, we evaluate a broader math/reasoning pipeline using Open-R1 SFT~\citep{openr1} followed by DAPO-math RL with DAPO~\citep{yu2025dapoopensourcellmreinforcement}. As shown in Figure~\ref{fig:broader_dapo_summary_main} and Table~\ref{tab:broader_dapo_passk_main}, SFT improves or maintains ID performance but reduces the OOD benchmark average, while DAPO partially restores the lost OOD performance. For Qwen-2.5-7B, the OOD average drops from $55.26$ at $\text{SFT}_{\text{MaxOOD}}$ to $51.11$ at $\text{SFT}_{\text{End}}$, and DAPO recovers it to $54.64$. For Qwen-2.5-3B, the OOD average drops from $46.36$ to $41.98$, and DAPO recovers it to $43.57$. These results support our central claim that RL is a strong post-SFT restorer, but does not generally exceed the earlier OOD-max checkpoint. Full per-task ID and OOD results are provided in Appendix~\ref{appendix:broader_reasoning_dapo}.

\vspace{-0.2cm}
\section{Rotation Matters: A SVD Analysis on Parameter Matrices}
\label{sec:param_matrices_analysis}

Based on our "SFT forgets, RL recovers", found in Section~\ref{sec:results_analysis}, we would like to understand what is the underlying mechanism that causes the different behaviors of SFT vs. RL. Recent work has shown that the spectrum of parameter matrices can offer an interpretable window on how its internal representations evolve and how they relate to downstream performance~\citep{staats2025smallsingularvaluesmatter, yunis2024approaching}. With this lens, we can track the changes in parameter space during SFT and RL stages with Singular Value Decomposition (SVD)~\citep{aghajanyan2020intrinsic, yunis2024approaching}  and conduct ablation studies to explore the impacts of singular values/vectors of weight matrices on model performance. We introduce the experimental setup in Section~\ref{sec:svd_setup}, present the results and analysis in Section~\ref{sec:ablation_singular_values} and~\ref{sec:ablation_singular_vectors}.
\subsection{Setup}
\label{sec:svd_setup}
Based on some recent findings~\citep{staats2025smallsingularvaluesmatter, ijcai2023p493, yuan2024asvdactivationawaresingularvalue}, which highlight the significance of self-attention parameter matrices in weight adaptation, our analysis focuses on two sets of parameter matrices:
\begin{itemize}
    \item  \textbf{$\bm{W}_Q, \bm{W}_K, \bm{W}_V$ in self-attention matrices} are the core components of the self-attention mechanism~\citep{vaswani2017attention}. They function by projecting the input embeddings into distinct subspaces to compute attention scores and construct context-aware representations. 
    \item \textbf{$\bm{W}_\text{MLP}$ in MLP layer} in both LLM models, every MLP block uses an up-projection to widen the hidden state, a gate-projection to apply the SwiGLU gate~\citep{shazeer2020gluvariantsimprovetransformer}, and a down-projection to shrink it back. We did not include the bias term $\bm{b}_\text{MLP}$ in SVD analysis because this term is found to only have minor impact on model performance.
\end{itemize}

To investigate how SFT- and RL-reshaped parameter matrices impact the model performance, we conduct ablation studies on the singular values/vectors of the above parameter matrices (we use the result on \textit{GeneralPoints} as an example). Specifically,
\begin{itemize}
    \item for singular values, we restore the singular values of the fine-tuned parameter matrices, while keeping the corresponding singular vectors unmodified, and see if the model performance (OOD forgetting and recovery) will be reverted accordingly. In other words, we roll back $\bm{\Sigma}_{\text{SFT}_\text{End}} \rightarrow \bm{\Sigma}_{\text{SFT}_\text{MaxOOD}}, \bm{\Sigma}_{\text{RL}_\text{End}} \rightarrow \bm{\Sigma}_{\text{SFT}_\text{End}}$, and evaluate the models with parameter matrices $\bm{U}_{\text{SFT}_\text{End}} \bm{\Sigma}_{\text{SFT}_\text{MaxOOD}} \bm{V}_{\text{SFT}_\text{End}}^\top$ and $\bm{U}_{\text{RL}_\text{End}} \bm{\Sigma}_{\text{SFT}_\text{End}} \bm{V}_{\text{RL}_\text{End}}^\top$ and check the performance shifts.
    \item Similar to the restoration of singular vectors, we evaluate the model performance with parameter matrices $\bm{U}_{\text{SFT}_\text{MaxOOD}} \bm{\Sigma}_{\text{SFT}_\text{End}} \bm{V}_{\text{SFT}_\text{MaxOOD}}^\top$ and $\bm{U}_{\text{SFT}_\text{End}} \bm{\Sigma}_{\text{RL}_\text{End}} \bm{V}_{\text{SFT}_\text{End}}^\top$.
\end{itemize} 
For LLaMA $\text{SFT}_\text{MaxOOD} = 140, \text{SFT}_\text{End} =1100$ and for Qwen$\text{SFT}_\text{MaxOOD} = 120, \text{SFT}_\text{End} =800$. 

To identify which layers and which set of singular values/vectors play a more important role in OOD forgetting and recovery, we perform the restoration process step by step across layers and top-$k$ singular values/vectors. More specifically,
\begin{itemize}
    \item for layer-wise study, we restore the singular values/vectors for every top-$k$ layer, where $k=5,10,15,20,\dots, L$ and $L$ is the total number of layers;
    \item for singular values and vectors, we restore the top-$k$ singular values/vectors for all layers, where $k=64, 256, 512, 768, 1024, 1536, 2048, 2560,$ $ 3072, 3584, (4096 \text{ for LLaMA})$;
\end{itemize}

The results are shown in Section~\ref{sec:ablation_singular_values} and ~\ref{sec:ablation_singular_vectors}.

\begin{figure*}[h]
  \begin{subfigure}[t]{0.24\linewidth}
    \centering
    \includegraphics[width=\linewidth]{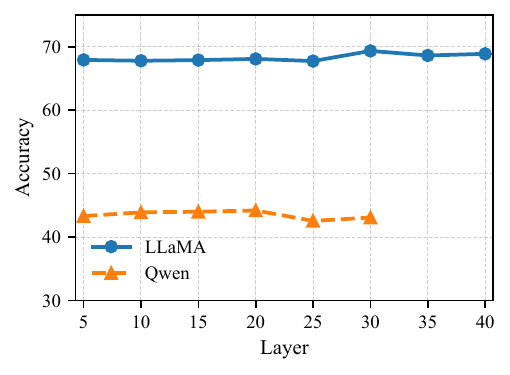}
    \caption{Layer-wise (ID) }
    \label{fig:ablation_sft_singular_value_layerwise_id}
  \end{subfigure}
  \begin{subfigure}[t]{0.24\linewidth}
    \centering
    \includegraphics[width=\linewidth]{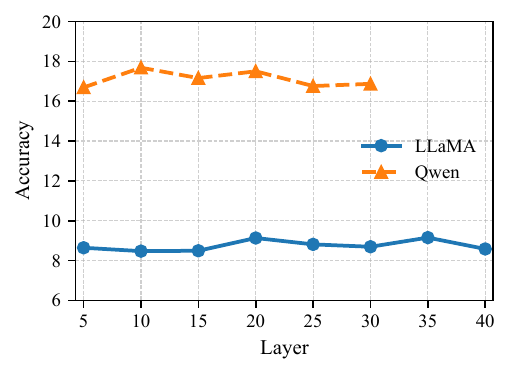}
    \caption{Layer-wise (OOD)}
    \label{fig:ablation_sft_singular_value_layerwise_ood}
  \end{subfigure}
   \begin{subfigure}[t]{0.24\linewidth}
    \centering
    \includegraphics[width=\linewidth]{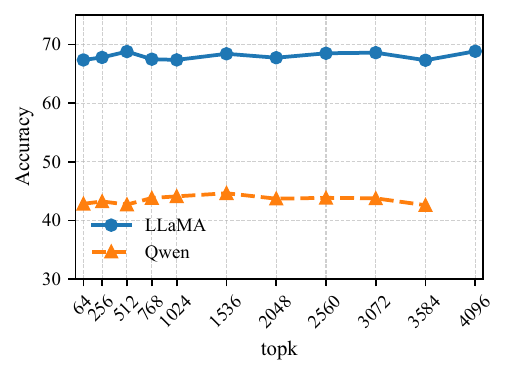}
    \caption{Top-$k$ (ID)}
    \label{fig:ablation_sft_singular_value_topk_id}
  \end{subfigure}
  \begin{subfigure}[t]{0.24\linewidth}
    \centering
    \includegraphics[width=\linewidth]{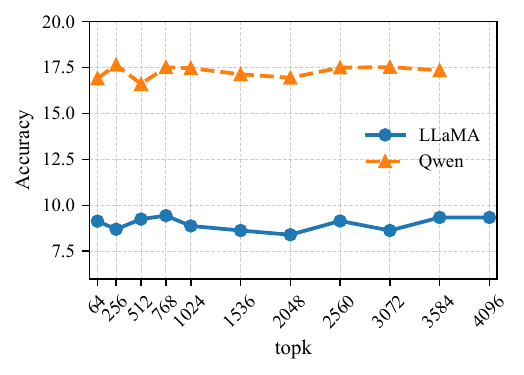}
    \caption{Top-$k$ (OOD)}
    \label{fig:ablation_sft_singular_value_topk_ood}
  \end{subfigure}
  \caption{Singular \textbf{value} restoration for \textbf{SFT} stage. 
  }
  \label{fig:sft_singular_value_recover_accuracy}
\end{figure*}
\begin{figure*}[h]
  \begin{subfigure}[t]{0.24\linewidth}
    \centering
    \includegraphics[width=\linewidth]{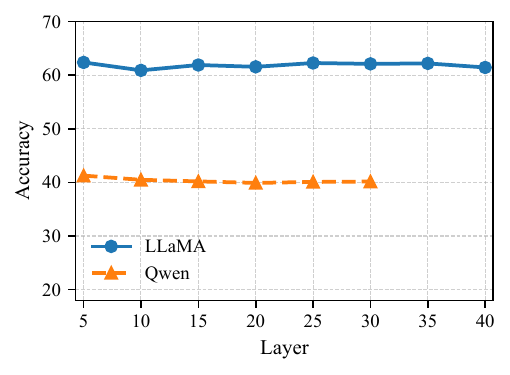}
    \caption{Layer-wise (ID)}
    \label{fig:ablation_rl_singular_value_layerwise_id}
  \end{subfigure}                        
  \begin{subfigure}[t]{0.24\linewidth}
    \centering
    \includegraphics[width=\linewidth]{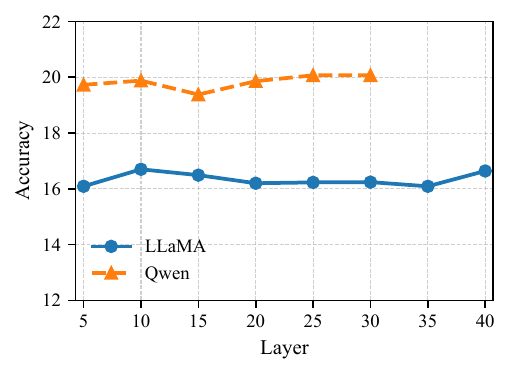}
    \caption{Layer-wise (OOD)}
    \label{fig:ablation_rl_singular_value_layerwise_ood}
  \end{subfigure}
   \begin{subfigure}[t]{0.24\linewidth}
    \centering
    \includegraphics[width=\linewidth]{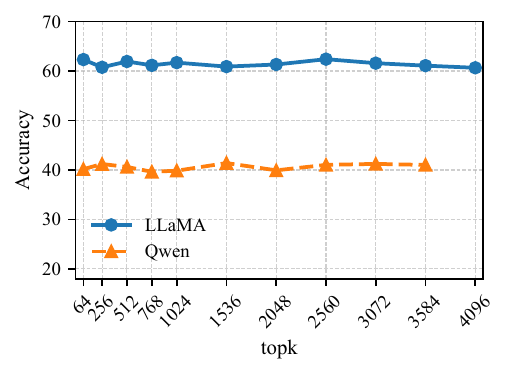}
    \caption{Top-$k$ (ID)}
    \label{fig:ablation_rl_singular_value_topk_id}
  \end{subfigure}                        
  \begin{subfigure}[t]{0.24\linewidth}
    \centering
    \includegraphics[width=\linewidth]{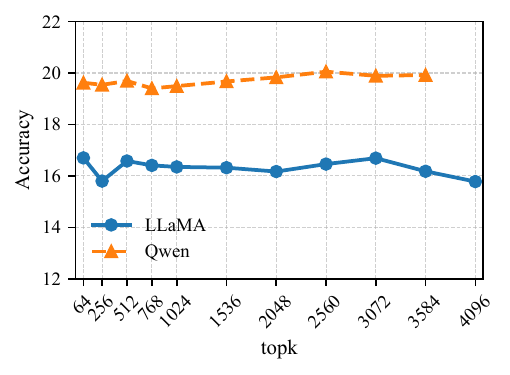}
    \caption{Top-$k$ (OOD)}
    \label{fig:ablation_rl_singular_value_topk_ood}
  \end{subfigure}
  \caption{Singular \textbf{value} restoration for \textbf{RL} stage.}
  \label{fig:rl_singular_value_recover_accuracy}
\end{figure*}

\subsection{Ablation Studies on Singular Values}
\label{sec:ablation_singular_values}
It is found in existing literature that the intrinsic capacity of the model is mainly reflected by the singular values~\citep{bartlett2017spectrallynormalizedmarginboundsneural, yoshida2017spectralnormregularizationimproving, li2024setaroutofdistributiondetectionselective}.
However, from our results of singular value restoration in SFT stage shown in Figure~\ref{fig:sft_singular_value_recover_accuracy}, and the results in RL stage shown in Figure~\ref{fig:rl_singular_value_recover_accuracy}~\footnote{See a more detailed study in Appendix~\ref{app:singular_value_sft_rl}}, we observe that: \textbf{the restoration of the singular values of parameter matrices has negligible impact on ID and OOD performance for both SFT and RL fine-tuned models}. 

Besides, as the additional evidence shown in Appendix~\ref{app:singular_value_sft_rl}, compared to the original values, the differences of singular values caused by fine-tuning only fluctuate from $0$ to $0.005$, which act almost as zero-centered noisy signals. This indicates that the fine-tuning process does not significantly amplify or diminish specific singular values. And we do not observe significant shifts concentrated in any particular region, such as the head (largest values) or tail (smallest values), which is found in previous studies~\citep{staats2025smallsingularvaluesmatter, Thamm_2022, saada2025mindgapspectralanalysis, cancedda2024spectralfiltersdarksignals, hsu2022languagemodelcompressionweighted}. 

\subsection{Ablation Studies on Singular Vector Directions}
\label{sec:ablation_singular_vectors}
\begin{figure*}[t]        
  \begin{subfigure}[t]{0.24\linewidth}
    \centering
    \includegraphics[width=\linewidth]{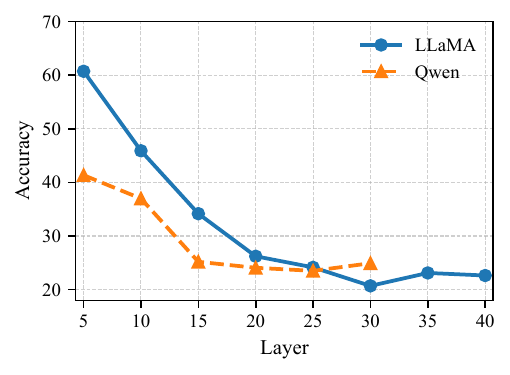} 
    \caption{Layer-wise (ID) }
    \label{fig:ablation_sft_singular_vector_layerwise_id}
  \end{subfigure}                      
  \begin{subfigure}[t]{0.24\linewidth}
    \centering
    \includegraphics[width=\linewidth]{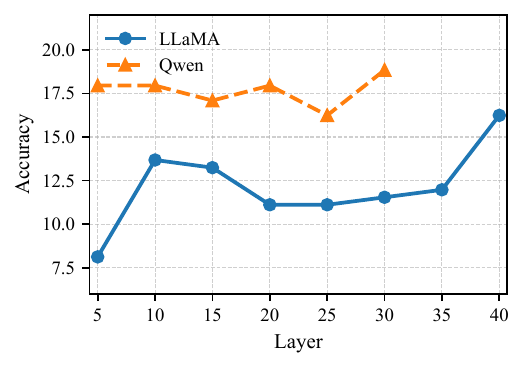}
    \caption{Layer-wise (OOD) }
        \label{fig:ablation_sft_singular_vector_layerwise_ood}
  \end{subfigure}
  \begin{subfigure}[t]{0.24\linewidth}
    \centering
    \includegraphics[width=\linewidth]{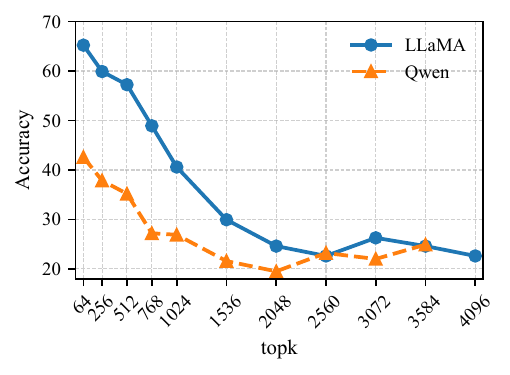} 
    \caption{Top-$k$ (ID)}
    \label{fig:ablation_sft_singular_vector_topk_id}
  \end{subfigure}
  \begin{subfigure}[t]{0.24\linewidth}
    \centering
    \includegraphics[width=\linewidth]{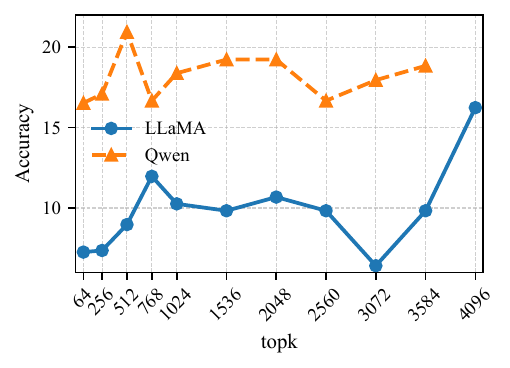}
    \caption{Top-$k$ (OOD)}
    \label{fig:ablation_sft_singular_vector_topk_ood}
  \end{subfigure}
  \caption{Singular \textbf{vector} restoration for \textbf{SFT} stage.}
  \label{fig:sft_singular_vector_recover}
\end{figure*}

\begin{figure*}[htbp]        
  \begin{subfigure}[t]{0.24\linewidth}
    \centering
    \includegraphics[width=\linewidth]{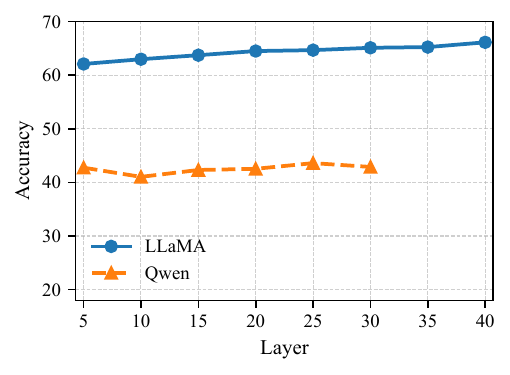} 
    \caption{Layer-wise (ID) }
    \label{fig:ablation_rl_singular_vector_layerwise_id}
  \end{subfigure}
  \begin{subfigure}[t]{0.24\linewidth}
    \centering
    \includegraphics[width=\linewidth]{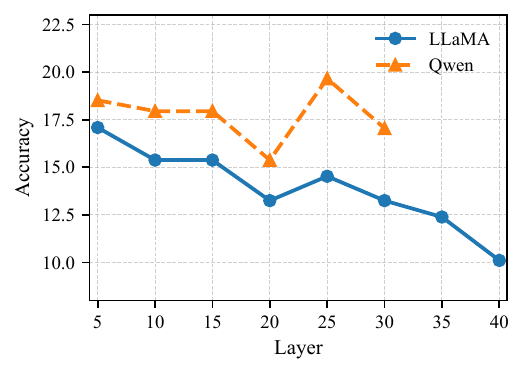}
    \caption{Layer-wise (OOD) }
    \label{fig:ablation_rl_singular_vector_layerwise_ood}
  \end{subfigure}
  \begin{subfigure}[t]{0.24\linewidth}
    \centering
    \includegraphics[width=\linewidth]{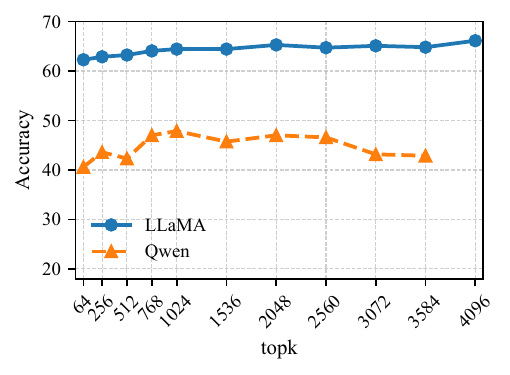} %
    \caption{Top-$k$ (ID)}
    \label{fig:ablation_rl_singular_vector_topk_id}
  \end{subfigure}
  \begin{subfigure}[t]{0.24\linewidth}
    \centering
    \includegraphics[width=\linewidth]{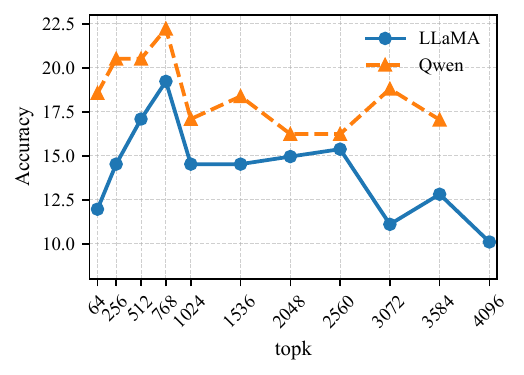}
    \caption{Top-$k$ (OOD)}
    \label{fig:ablation_rl_singular_vector_topk_ood}
  \end{subfigure}
  \caption{Singular \textbf{vector} restoration for \textbf{RL} stage.}
  \label{fig:rl_singular_vector_recover}
\end{figure*}

The results of singular vector restoration in SFT and RL stages are shown in Figure~\ref{fig:sft_singular_vector_recover} and Figure~\ref{fig:rl_singular_vector_recover}. It is quite clear that \textbf{the rotation of the singular vectors plays a more important role than singular values in fine-tuning}, as the ID and OOD performance shift much more significantly.
We analyze their fine-grained correlations in SFT stage as follows, 
\begin{itemize}
    \item \textbf{Layer-wise Analysis} As shown in Figure~\ref{fig:ablation_sft_singular_vector_layerwise_id} and~\ref{fig:ablation_sft_singular_vector_layerwise_ood}, restoring the singular vectors of first $30$ layers of LLaMA and first $15$ layers of Qwen causes significant degradation of ID performance. And the restoration of first $10$ and last $5$ layers leads to the recovery of OOD performance in LLaMA, however, Qwen stays relatively robust. This suggests that, in SFT stage, the task-specific knowledge  does not depend too much on the last several layers and OOD capabilities are highly impacted by the top and bottom blocks of the models.
	\item \textbf{Top-$k$ Analysis} As shown in Figure~\ref{fig:ablation_sft_singular_vector_topk_id} and~\ref{fig:ablation_sft_singular_vector_topk_ood}, restoring the top $2560$ singular vectors of LLaMA and top $2048$ singular vectors of Qwen causes significant degradation of ID performance. And the restoration of top $768$ singular vectors and last $1024$ singular vectors leads to the recovery of OOD performance in LLaMA, however, Qwen stays relatively robust again. This indicates that, in SFT stage, the task-specific knowledge is mainly stored in the first several singular vectors and OOD capabilities in the top and bottom blocks.
\end{itemize}

\paragraph{Algorithmic implication.} 

The top-$k$ singular-vector analysis motivates a protected-SFT update that suppresses rotation in OOD-related subspaces. For $W=U\Sigma V^\top$ and SFT gradient $G$, we project
\(\widetilde{G}_{\perp}=(I-U_kU_k^\top)G(I-V_kV_k^\top),\; W\leftarrow W-\eta\widetilde{G}_{\perp}\),
where $U_k,V_k$ are protected top-$k$ singular-vector subspaces. We apply this projection to selected attention and MLP matrices. Appendix Figure~\ref{fig:svd_protected_sft_appendix} and Tables~\ref{tab:svd_protected_sft_full},~\ref{tab:rotation_protected_sft_full} show that this preliminary protected-SFT variant preserves more late-SFT OOD accuracy while maintaining comparable ID accuracy.

In RL stage, we observe that
\begin{itemize}
    \item \textbf{Layer-wise Analysis} As shown in Figure~\ref{fig:ablation_rl_singular_vector_layerwise_id} and~\ref{fig:ablation_rl_singular_vector_layerwise_ood}, the restoration of singular vectors consistently causes performance degradation of ID and OOD performance for LLaMA, with some perturbations in intermediate ($15-25$) layers for OOD performance. ID and OOD performance of Qwen is relatively robust, and also has some perturbations in intermediate ($15-25$) layers for OOD performance.
    This indicates that RL uniformly impacts each layer in LLaMA for both task-specific knowledge and OOD ability.
    \item \textbf{Top-$k$ Analysis} As shown in Figure~\ref{fig:ablation_rl_singular_vector_topk_id} and~\ref{fig:ablation_rl_singular_vector_topk_ood}, the restoration of singular vectors uniformly causes a performance degradation of ID performance for LLaMA, Qwen is relatively robust. For OOD performance, the top ($1024$) and bottom ($2560-4096$ for LLaMA, $2560-3584$ for Qwen) singular vectors are highly relevant.
\end{itemize}

\section{Related Work on RL Reasoning}
\label{sec:related_work}

\paragraph{RL Improves Reasoning and OOD Generalization.}
Recent large-scale RL fine-tuning has substantially improved LLM reasoning performance, especially on math and coding tasks~\citep{deepseekai2025deepseekr1incentivizingreasoningcapability, wang2025reinforcement}. This has motivated a growing line of work studying how RL differs from SFT in reasoning and generalization. In particular, prior work argues that SFT tends to memorize training distributions, whereas RL can induce more transferable reasoning behavior~\citep{chu2025sftmemorizesrlgeneralizes}. Rule-based RL has also been shown to improve reasoning on synthetic logic tasks and transfer to harder benchmarks such as AIME and AMC~\citep{xie2025logicrlunleashingllmreasoning}. Other work combines SFT and RL more directly, for example by injecting supervised signals into RL-style training objectives~\citep{liu2025uft, huang2025blending}. These studies support the view that RL is an important component of modern reasoning post-training.

\paragraph{Does RL Create New Reasoning Capability?}
At the same time, several recent studies question whether RL truly expands the model's underlying reasoning capability. \citet{yue2025doesreinforcementlearningreally} argue that RL with verifiable rewards mainly improves sampling efficiency rather than the capability boundary, and that base models can outperform RL-trained models at high pass@$k$. \citet{kim2025reinforcementlearningvsdistillation} similarly suggest that RL improves easier problems while potentially hurting harder ones. \citet{zheng2025fanchuan} further show that RL-enhanced reasoning models can underperform their base models on parody detection, suggesting that RL gains may not transfer uniformly across reasoning tasks.

\paragraph{Our Perspective.}
Our work provides a checkpoint-wise explanation for these mixed observations. Rather than comparing only a terminal SFT checkpoint with a subsequent RL checkpoint, we track ID and OOD behavior throughout SFT and then apply RL from multiple SFT checkpoints. This reveals that SFT can first improve and then forget OOD reasoning ability, while RL mainly restores the OOD ability lost during later SFT. Therefore, the claim ``SFT memorizes, RL generalizes'' captures only part of the story: in our experiments, RL mainly recovers lost OOD capability within a bounded range of SFT initializations, while the best OOD behavior often appears earlier in the SFT trajectory.

\section{Conclusions} 
\label{sec:conclusion_future_work}
In this paper, we re-examine the roles of SFT and RL in two-stage fine-tuning and generalize the common belief ``SFT memorizes, RL generalizes'' to ``SFT forgets, RL recovers''. We identify OOD forgetting during SFT, OOD recovery during RL, and a boundary on when RL remains effective. RLFT restores lost OOD ability from late-SFT checkpoints and rarely surpasses the best OOD checkpoint already reached during SFT. Advantage-distribution analysis shows that flat, less spiky, structured advantages correlate with effective RL; robustness checks across tasks, seeds, and RL algorithms support the claim. Finally, SVD analysis shows that OOD behavior tracks singular-vector rotation rather than singular-value changes, suggesting a route to mitigate OOD forgetting during SFT. We discuss limitations and future directions in Appendix~\ref{appendix:limitations}.




\clearpage
\bibliographystyle{plainnat}
\bibliography{main}

\clearpage
\onecolumn
\appendix

\startcontents[appendix]
\section*{Appendix Contents}
\printcontents[appendix]{}{1}{\setcounter{tocdepth}{2}}
\clearpage

\section{Limitations}
\label{appendix:limitations}

Our study has several limitations. First, the densest checkpoint-wise SFT/RL sweeps are conducted on controlled reasoning tasks, where the ID/OOD split is explicitly defined by rule changes. This design is useful for diagnosing fine-tuning dynamics and reducing ambiguity from unknown pre-training contamination, but it does not cover the full diversity of open-ended reasoning distributions. To partially address this, we include broader Open-R1 SFT followed by DAPO-math RL results on public math and reasoning benchmarks, but denser checkpoint sweeps on such benchmarks remain computationally expensive.

Second, our main mechanistic analysis focuses on singular-value and singular-vector changes in selected attention and MLP parameter matrices. The restoration and protected-SFT experiments suggest that singular-vector rotation is strongly related to OOD forgetting and recovery, but they do not constitute a complete causal theory of the learning dynamics. Other factors, such as representation drift, optimization noise, reward design, data mixture, and decoding behavior, may also interact with the observed phenomenon.

Third, the proposed singular-vector-protected SFT variant is preliminary. It is intended to validate the algorithmic implication of our SVD analysis rather than to provide a fully optimized training algorithm. Future work should study how to choose the protected subspace automatically, how to balance ID specialization and OOD preservation, and how this regularization interacts with different RL algorithms and model scales.

Finally, our conclusion that RL restores rather than creates fundamentally new OOD capability is based on the post-training settings studied in this paper. We evaluate multiple models, tasks, seeds, and RL algorithms, but larger-scale training regimes, different reward models, or substantially different data distributions may exhibit additional behaviors. These limitations do not affect the central empirical observation that terminal-SFT-vs.-RL comparisons can miss earlier SFT checkpoints with stronger OOD performance, but they motivate broader future studies of SFT/RL dynamics.

\section{Additional Evidence and Full Data Tables}
\label{appendix:additional_evidence_full_tables}

This section collects auxiliary evidence for the main claims: broader reasoning benchmarks and DAPO results, boundary sweeps, cold-start RL evidence, and mitigation/mechanistic evidence.

\subsection{Broader Benchmark Details: Open-R1 SFT Followed by DAPO}
\label{appendix:broader_reasoning_dapo}

The main paper reports the broader benchmark summary in Table~\ref{tab:broader_dapo_passk_main}, including all ID pass@$k$ scores, all OOD pass@1 scores, and their averages. To avoid repeating the same results, this appendix reports only the additional ID avg@$k$ datapoints for the Open-R1 SFT~\citep{openr1} followed by DAPO-math RL with DAPO~\citep{yu2025dapoopensourcellmreinforcement} experiment. These avg@$k$ results complement the pass@$k$ scores in the main paper and show the same qualitative pattern: DAPO improves the average ID reasoning score after SFT, while the main-paper OOD pass@1 table shows only partial OOD restoration.

\begin{table}[htbp]
\centering
\small
\resizebox{\textwidth}{!}{%
\begin{tabular}{@{}llrrrrrrr@{}}
\toprule
\textbf{Model} & \textbf{Setting}
& \textbf{AIME24 avg@16}
& \textbf{AIME25 avg@16}
& \textbf{AMC avg@16}
& \textbf{MATH-500 avg@4}
& \textbf{Minerva avg@8}
& \textbf{Olympiad avg@4}
& \textbf{Avg(avg@$k$)} \\
\midrule
Qwen-2.5-7B & OOD-max    
& 12.08 & 7.70 & 52.03 & 72.85 & 37.09 & 37.41 & 36.53 \\
Qwen-2.5-7B & SFT-end    
& 15.21 & 19.38 & 50.47 & 72.85 & 31.52 & 37.41 & 37.81 \\
Qwen-2.5-7B & after DAPO 
& 18.75 & 20.83 & 58.13 & 79.35 & 38.97 & 43.44 & 43.24 \\
\midrule
Qwen-2.5-3B & OOD-max    
& 4.79 & 2.29 & 37.03 & 63.35 & 26.24 & 26.22 & 26.65 \\
Qwen-2.5-3B & SFT-end    
& 3.96 & 5.00 & 35.00 & 60.95 & 22.75 & 26.25 & 25.65 \\
Qwen-2.5-3B & after DAPO 
& 4.79 & 5.42 & 37.50 & 65.90 & 25.32 & 27.44 & 27.73 \\
\bottomrule
\end{tabular}%
}
\caption{
Additional ID avg@$k$ results for Open-R1 SFT followed by DAPO-math RL with DAPO. The corresponding ID pass@$k$ results and all OOD pass@1 results are reported in the main paper in Table~\ref{tab:broader_dapo_passk_main}.
}
\label{tab:broader_id_avg_full}
\end{table}

\subsection{Boundary Evidence Beyond the Main LLaMA Sweep}
\label{appendix:additional_boundary_evidence}

\begin{figure}[htbp]
  \centering
  \includegraphics[width=0.75\linewidth]{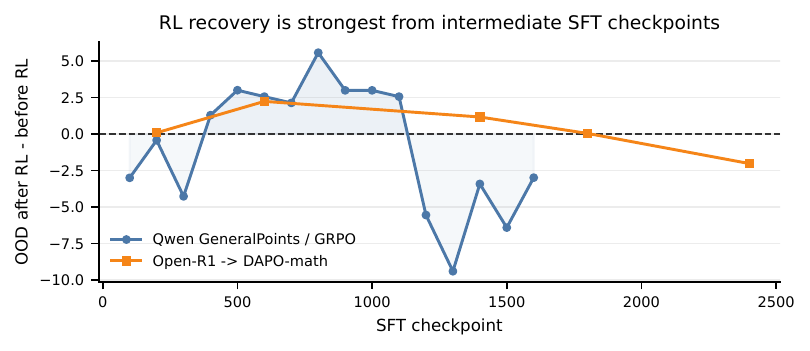}
  \caption{OOD recovery deltas for Qwen with GRPO and Open-R1 SFT followed by DAPO-math RL. Positive values mean RL improves OOD over the SFT initialization.}
  \label{fig:boundary_recovery_delta_appendix}
\end{figure}

\begin{figure}[htbp]
  \centering
  \includegraphics[width=\linewidth]{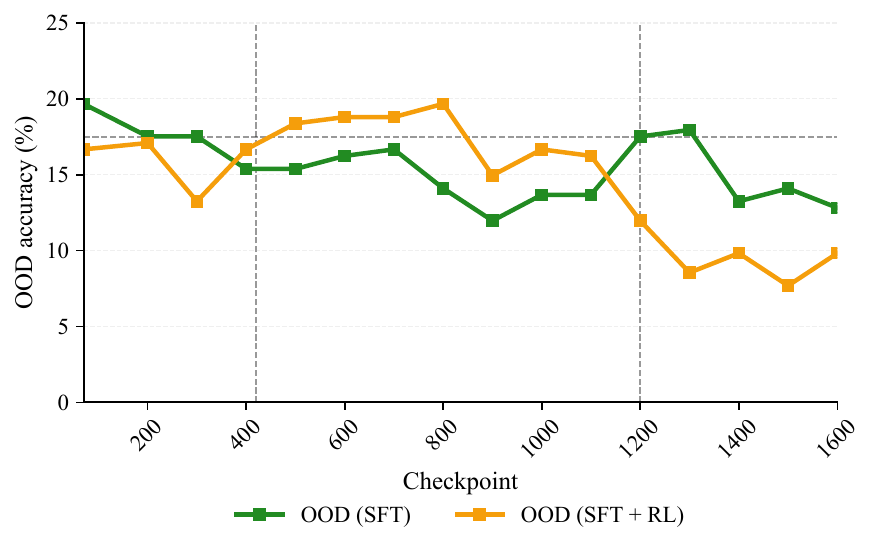}
  \caption{Qwen checkpoint-wise OOD recovery boundary with GRPO on GeneralPoints.}
  \label{fig:qwen_ood_recovery_boundary_appendix}
\end{figure}

We further test the recovery boundary with two complementary datasets. Figure~\ref{fig:boundary_recovery_delta_appendix} summarizes the recovery deltas, Figure~\ref{fig:qwen_ood_recovery_boundary_appendix} and Table~\ref{tab:qwen_grpo_boundary_full} give the full Qwen checkpoint sweep with GRPO on GeneralPoints, and Table~\ref{tab:dapo_complex_boundary} gives the sparse boundary evaluation for a broader reasoning setup, using Open-R1 SFT followed by DAPO-math RL with DAPO.

\begin{table}[htbp]
\centering
\small
\begin{tabular}{@{}rrrrr@{}}
\toprule
\textbf{SFT checkpoint} & \textbf{ID before RL} & \textbf{OOD before RL} & \textbf{ID after RL} & \textbf{OOD after RL} \\
\midrule
100  & 20.09 & 19.67 & 20.59 & 16.67 \\
200  & 25.21 & 17.52 & 25.84 & 17.09 \\
300  & 32.05 & 17.52 & 32.85 & 13.25 \\
400  & 32.48 & 15.38 & 33.29 & 16.67 \\
500  & 32.05 & 15.38 & 32.85 & 18.38 \\
600  & 35.04 & 16.24 & 35.90 & 18.80 \\
700  & 32.91 & 16.67 & 34.19 & 18.80 \\
800  & 46.58 & 14.10 & 47.74 & 19.67 \\
900  & 47.86 & 11.97 & 49.06 & 14.96 \\
1000 & 49.57 & 13.68 & 50.81 & 16.67 \\
1100 & 57.26 & 13.68 & 63.25 & 16.24 \\
1200 & 61.11 & 17.52 & 60.26 & 11.97 \\
1300 & 60.70 & 17.95 & 57.26 & 8.55 \\
1400 & 64.10 & 13.25 & 62.82 & 9.83 \\
1500 & 61.11 & 14.10 & 67.52 & 7.69 \\
1600 & 67.52 & 12.82 & 73.08 & 9.83 \\
\bottomrule
\end{tabular}
\caption{Full Qwen checkpoint sweep with GRPO on GeneralPoints. The OOD recovery effect is strongest for intermediate checkpoints and weakens for late checkpoints.}
\label{tab:qwen_grpo_boundary_full}
\end{table}

\begin{table}[htbp]
\centering
\small
\resizebox{\linewidth}{!}{%
\begin{tabular}{@{}rlrrrr@{}}
\toprule
\textbf{Checkpoint} & \textbf{Setting} & \textbf{IFEval-loose} & \textbf{MMLU-Pro} & \textbf{Safety} & \textbf{TruthfulQA-MC} \\
\midrule
0    & OOD-max    & 54.16 & 52.86 & 65.90 & 59.94 \\
200  & SFT-end    & 51.76 & 44.29 & 62.70 & 54.82 \\
200  & after DAPO & 49.91 & 47.14 & 62.90 & 53.95 \\
600  & SFT-end    & 47.32 & 45.71 & 59.60 & 49.42 \\
600  & after DAPO & 48.43 & 47.14 & 62.50 & 52.92 \\
1400 & SFT-end    & 44.54 & 45.71 & 59.50 & 48.83 \\
1400 & after DAPO & 44.92 & 47.14 & 60.30 & 50.88 \\
1800 & SFT-end    & 42.88 & 41.43 & 62.80 & 49.42 \\
1800 & after DAPO & 41.96 & 44.28 & 60.60 & 49.85 \\
2400 & SFT-end    & 39.37 & 42.86 & 61.80 & 50.00 \\
2400 & after DAPO & 38.08 & 42.85 & 58.00 & 47.01 \\
\bottomrule
\end{tabular}%
}
\caption{Boundary for more complex reasoning: Open-R1 for SFT, then DAPO-math for RL with the DAPO algorithm on Qwen-2.5-3B.}
\label{tab:dapo_complex_boundary}
\end{table}

\subsection{Cold-Start RL Evidence}
\label{appendix:cold_start_rl_evidence}

Table~\ref{tab:qwen_cold_start_rl} reports the Qwen-2.5-7B cold-start GRPO result. It supports the boundary intuition that a policy without enough SFT task competence can receive poor reward/advantage signals and fail to recover OOD ability.

\begin{table}[htbp]
\centering
\small
\resizebox{\textwidth}{!}{%
\begin{tabular}{@{}lrrrrrr@{}}
\toprule
\textbf{Setting} & \textbf{GPQA} & \textbf{IFEval-loose} & \textbf{Safety} & \textbf{AIME24 avg@16} & \textbf{AIME25 avg@16} & \textbf{AMC avg@16} \\
\midrule
Base         & 33.93 & 61.37 & 74.50 & 12.08 & 7.70 & 52.03 \\
RL-400 steps & 15.18 & 58.96 & 6.50 & 13.33 & 12.29 & 53.13 \\
\bottomrule
\end{tabular}%
}
\caption{Qwen-2.5-7B cold-start RL with GRPO.}
\label{tab:qwen_cold_start_rl}
\end{table}

\subsection{Mitigation and Singular-Vector Evidence}
\label{appendix:mitigation_singular_vector_evidence}

\begin{figure}[htbp]
  \centering
  \includegraphics[width=0.75\linewidth]{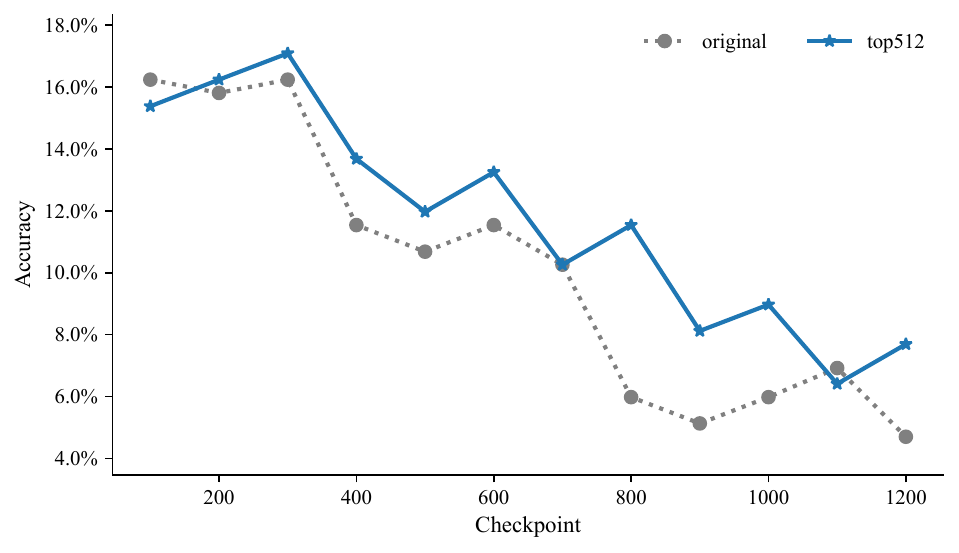}
  \caption{Preliminary SFT-side protection result. Orthogonalizing top-rank singular-vector directions during SFT is promising for preserving OOD ability.}
  \label{fig:svd_protected_sft_appendix}
\end{figure}

Figure~\ref{fig:svd_protected_sft_appendix} and Tables~\ref{tab:svd_protected_sft_full} and~\ref{tab:rotation_protected_sft_full} provide the full datapoints for protected SFT. The protected-SFT experiment is preliminary, but it motivates the mitigation direction of protecting or regularizing important singular-vector subspaces during SFT.

\begin{table}[htbp]
\centering
\small
\begin{tabular}{@{}rrrrr@{}}
\toprule
\textbf{Checkpoint} & \textbf{OOD} & \textbf{OOD protected} & \textbf{ID} & \textbf{ID protected} \\
\midrule
100  & 16.24 & 15.38 & 24.79 & 21.79 \\
200  & 15.81 & 16.24 & 31.62 & 26.07 \\
300  & 16.24 & 17.09 & 37.18 & 34.19 \\
400  & 11.54 & 13.68 & 37.61 & 36.19 \\
500  & 10.68 & 11.97 & 45.73 & 44.87 \\
600  & 11.54 & 13.25 & 50.00 & 47.86 \\
700  & 10.26 & 10.26 & 58.12 & 60.68 \\
800  & 5.98 & 11.54 & 64.00 & 66.67 \\
900  & 5.13 & 8.12 & 70.09 & 69.66 \\
1000 & 5.98 & 8.97 & 66.67 & 70.35 \\
1100 & 6.92 & 6.41 & 68.38 & 72.65 \\
1200 & 4.70 & 7.69 & 76.07 & 75.68 \\
\bottomrule
\end{tabular}
\caption{SFT with singular-vector protection on GeneralPoints. Protection often preserves more OOD accuracy in late SFT while maintaining comparable ID accuracy.}
\label{tab:svd_protected_sft_full}
\end{table}

\begin{table}[htbp]
\centering
\small
\resizebox{\textwidth}{!}{%
\begin{tabular}{@{}rrrrr@{}}
\toprule
\textbf{rank $k$} & \textbf{SFT V rotation} & \textbf{Protected SFT V rotation} & \textbf{SFT-to-RL V rotation} & \textbf{Protected-to-RL V rotation} \\
\midrule
1   & 0.0000  & 0.0000  & 0.274112 & 0.464360 \\
2   & 0.0000  & 0.0000  & 0.472712 & 0.480513 \\
4   & 0.0000  & 0.0000  & 0.596433 & 0.562322 \\
8   & 0.0000  & 0.0000  & 0.681275 & 0.707206 \\
16  & 0.0885  & 0.0000  & 0.767958 & 0.787830 \\
32  & 0.6904  & 0.0000  & 0.920899 & 0.918132 \\
64  & 0.7130  & 0.0000  & 0.938786 & 0.941908 \\
128 & 0.7538  & 0.0000  & 0.956544 & 0.958383 \\
256 & 0.8379  & 0.0000  & 0.977187 & 0.979388 \\
384 & 0.8838  & 0.7044  & 1.000342 & 1.003272 \\
512 & 13.9290 & 11.7914 & 1.025075 & 1.027744 \\
\bottomrule
\end{tabular}%
}
\caption{Principal-angle rotation of SFT, singular-vector-protected SFT, and their RL stages relative to the base-model subspace. Values are degrees.}
\label{tab:rotation_protected_sft_full}
\end{table}

\begin{table}[htbp]
\centering
\small
\begin{tabular}{@{}lrrrr@{}}
\toprule
\textbf{Metric} & \textbf{GP-L} & \textbf{V-IRL-L} & \textbf{GP-VL} & \textbf{V-IRL-VL} \\
\midrule
ID increase & +15.3 & +15.0 & +27.4 & +3.29 \\
OOD increase & +3.5 & +11.0 & +3.0 & +9.3 \\
OOD improve / ID increase & 22.9 & 73.3 & 10.9 & 282.7 \\
\bottomrule
\end{tabular}
\caption{Additional recovery-efficiency summary from the existing evidence notes.}
\label{tab:recovery_efficiency_summary}
\end{table}

\section{Clarification: Forgetting, Over-Specialization, Over-Fitting, Over-Training}

We would like to clarify the differences between the following concepts to highlight the uniqueness of our study on OOD forgetting and avoid confusion.

\begin{itemize}
    \item \textbf{Catastrophic Forgetting} means that a model loses prior knowledge or skills when trained on new data~\citep{li2024revisiting, kotha2024understanding}. More specifically, when we fine-tune an LLM on a new task, it underperforms the original LLM on previously learned tasks/domains.
    \item \textbf{Over-Specialization} refers to format specialization~\citep{wang2024two}, which means that a model becomes narrowly specialized to the format of a task during fine-tuning, even on some inappropriate places. It is a form of forgetting and will lead to failure of OOD generalization. It often happens rapidly at the early stage of forgetting, but may not degrade the deeper knowledge of the LLM.
    \item \textbf{Over-Fitting} happens when a model becomes too much tailored to the training (fine-tuning) data, instead of capturing the general patterns of the corresponding domain. We will observe a decrease in training loss and an increase in validation and test loss. Note that over-fitting is only relevant to the in-distribution generalization settings, but not OOD generalization.
    \item \textbf{Over-Training} happens in model pre-training stage. It means that when pretraining extends too long, even though the base model improves, the post-training performance will drop due to the increased sensitivity to parameter modification~\citep{springer2025overtrained}.
\end{itemize}

\section{More Details of Experimental Settings}
\label{appendix:experimental_settings}

\subsection{Task Descriptions and Prompts}
\label{appendix:task_description}

\paragraph{GeneralPoints}
Prompts and examples for the GeneralPoints game are shown as follows.

\noindent\fbox{%
  \begin{minipage}{\dimexpr\textwidth-2\fboxsep-2\fboxrule}
    \small
    [Task Description] \\
    You are an expert 24 points card game player. You will receive a set of 4 cards. Note that 'J', 'Q', and 'K' count as '10', and each card must be used once. Your goal is to output a formula that evaluates to 24 using numbers from the cards and operators such as `+', `-', `*', `/', and `='.

    [Input] \\
    Cards: [1, 3, K, 6]

    [Output] \\
    \{\,
      "cards": [x, y, z, w], where \{face\_card\_msg\},\\
      "number": [a, b, c, d], where a, b, c, and d are the numbers on the cards,\\
      "formula": 'an equation that equals 24',\\
    \}

   \begin{minipage}{\dimexpr\textwidth-2\fboxsep-2\fboxrule}
    \small
    For In-distribution Response:\\
    \{\\
    \quad "cards":[1, 3, K, 6],\\
    \quad "number":[1, 3, 10, 6],\\
    \quad "formula":"($10\times3$) $-$ ($6\div1$) = 24"\\
    \}\\[6pt]

    For Out-of-Distribution Response:\\
    \{\\
    \quad "cards":[1, 3, K, 6],\\
    \quad "number":[1, 3, 13, 6],\\
    \quad "formula":"($6\times(13-1)$)\,$\div\,3$ = 24"\\
    \}
  \end{minipage}%
  \end{minipage}%
}

\paragraph{Navigation}
For the Navigation task, which is used in ~\citep{chu2025sftmemorizesrlgeneralizes}, we train the model with an absolute direction \eg{} turn (source west), then we evaluate the model's OOD performance by relative direction \eg{} turn(left). 
Prompts and examples for Navigation are shown as follows.

\noindent\fbox{%
  \begin{minipage}{\dimexpr\textwidth-2\fboxsep-2\fboxrule}
    \small
    [Task Description] \\
    You are an expert in navigation. You will receive a sequence of instructions to follow. You
are also provided with your observation and action history in text. Your goal is to first analyze the instruction and identify the next sentence to be executed. 
Then, you need to provide the action to be taken based on the current observation and instruction.

    [Instruction] \\
1. First, turn right to face north.\\
2. Move forward until you reach next intersection.\\
3. Turn left to face west.\\
4. Move forward until you reach next intersection.\\
5. Turn left to face north.\\
6. Move forward until you reach next intersection.\\
7. Turn right to face east.\\
8. Move forward until you reach next intersection where Levi \& Korsinsky, LLP is on your right behind.\\
9. Turn left to face north.\\
10. Move forward until you reach next intersection.\\
11. Turn slightly right to face northeast.\\
12. Move forward until you reach next intersection.\\
13. Turn right to face northwest.\\
14. Move forward until you reach next intersection where Mr Goods Buy \& Sell is on your left front.\\
15. Turn left to face northeast.\\
16. Move forward until you reach next intersection where Skullfade Barbers is on your left front.\\
17. Turn right to face northwest.\\
18. Move forward until you reach destination where The destination Ann Cleaners is on your left.\\

    [Action space] \\
    forward(): indicates moving forward one step \\
    turn direction(x): indicates adjust the ego agent direction towards x direction. x could be any following 8 directions ['north', 'northeast', 'east', 'southeast', 'south', 'southwest', 'west', 'northwest']\\
    stop(): indicates the navigation is finished.\\
    vspace{6pt} \\

    [Observations and action sequence] \\
    $O_1$: No landmarks nearby; \\
    $A_1$: \\
   \begin{minipage}{\dimexpr\textwidth-2\fboxsep-2\fboxrule}
    \small
    For In-distribution Response:\\
    \{\\
    \quad "current observation": "No landmarks nearby; "\\
    \quad "current instruction": "First, turn right to face north."\\
    \quad "action": "turn direction(north)"\\
    \}\\[6pt]

    For Out-of-Distribution Response:\\
    \{\\
    \quad "current observation": "No landmarks nearby; "\\
    \quad "current instruction": "First, turn right to face north."\\
    \quad "action": "turn direction (right)"\\
    \}
  \end{minipage}
  \end{minipage}%
}

\paragraph{Rank-Determinant Computation}
For the matrix computation task, we train LLaMA and Qwen to compute the rank of a matrix with given dimension, \eg{}, 4$\times$5, then we employ the determinant compute as an OOD task to evaluate both models, which is adapted from~\citep{sun2025omega}. It evaluates not only the math computation, but also the cross-concept math reasoning ability, which is much more complex than the task in~\citep{sun2025omega}. 

For in-distribution training, the prompt is:

\noindent\fbox{%
  \begin{minipage}{\dimexpr\textwidth-2\fboxsep-2\fboxrule}
    \small
    [Task Description] \\
    You are an expert in linear algebra. You will receive a square matrix. Find the rank of the matrix and output the integer result.

    [Input] \\
    Matrix:[[-1, -2, 9, 3, -5], [0, -3, 9, 9, -6], [-2, -2, 12, 0, -6], [3, -2, -3, 15, -1]]

    [Output] \\
    \{\,
      "answer": 2,\\
    \}
  \end{minipage}%
}

For out-of-distribution evaluation, the prompt is:

\noindent\fbox{%
  \begin{minipage}{\dimexpr\textwidth-2\fboxsep-2\fboxrule}
    \small
    [Task Description] \\
    You are an expert in linear algebra. You will receive a square matrix. Compute its determinant and output the integer result.

    [Input] \\
    Matrix:[[-4, 3], [-3, -2]]

    [Output] \\
    \{\,
      "answer": 12,\\
    \}
  \end{minipage}%
}

\subsection{Computational Resources and Setups}
\label{appendix:computational_resources_setups}
All our RL fine-tuning is implemented on 8xH100 GPUs. SFT utilizes 4xH100 GPUs, the learning rate is 1e-6, a mini batch size of 64, and cosine is used as the learning rate schedule. We use PPO with rollout 256 to fine-tune the model after supervised fine-tuning. The checkpoint for different checkpoints may vary slightly due to the precision or computational resources. We adjust the number of checkpoints of SFT from 400 to 1100 as we employ 4 H100 GPUs for SFT instead of 8 H800 in the original paper. 

We only use LLaMA with GeneralPoints as an example to do the checkpoint-wise analysis, as the checkpoint sweep across all models and tasks is prohibited due to computational constraints. For example, in RL stage, we need to use 4 H100 GPUs to train for 24 hours or 8 H100 GPUs to train for 12 hours for each single setting of [model, checkpoint, task].

The decrease of ID performance is not due to an inadequate training issue in our RL setting, as we follow the same setup as~\cite{chu2025sftmemorizesrlgeneralizes}, \ i.e., we use 4 H100 GPUs to train 24 hours or 8 H100 GPUs to train 12 hours, depending on the availability of resources. Continued RLFT will lead to further decline for both ID and OOD performance. Such model deterioration of long-term RL training was also found by other researchers or practitioners. Therefore, we believe that there is no inadequate training problem.

We run 1100 SFT checkpoints in total for LLaMA-3.2-11B-Vision, 800 SFT checkpoints for Qwen-2.5-7B. Because both checkpoints are the boundaries for two models, the boundaries may change slightly with each run, and then 15 RL checkpoints for both of them. 

\begin{table}[h]
\caption{Details of implementation configurations.}
\label{training_settings}
\centering
\small
\begin{tabular}{@{}llcccccc@{}}
\toprule
\textbf{Task} & \textbf{Model} & \textbf{MaxOOD} & \textbf{SFT data} & \textbf{RL begin} & \textbf{\#RL ck} & \textbf{RL data} & \textbf{Eval data} \\
\midrule
GeneralPoint & LLaMA & 140 & 100k & 500--1100 & 15 & 60k & 234 \\
GeneralPoint & Qwen  & 120 & 80k  & 400--800  & 15 & 60k & 234 \\
Navigation   & LLaMA & 45  & 5k   & 60        & 15 & 60k & 234 \\
Navigation   & Qwen  & 95  & 10k  & 100       & 15 & 60k & 234 \\
Matrix       & LLaMA & 50  & 15k  & 140       & 24 & 10k & 234 \\
Matrix       & Qwen  & 650 & 80k  & 800       & 39 & 20k & 234 \\
\bottomrule
\end{tabular}
\label{tab:training_results}
\end{table}

For more details, please refer to the supplementary material.

\section{More Experimental Results}
\label{appendix:more_experimental_results}

\subsection{ID and OOD Loss in SFT}

After $50$ checkpoints, we find that the ID and OOD cross-entropy losses go to different directions. The ID loss approaches $0.15$, then keeps stable, and OOD loss increases after the same checkpoints. However, based on the results in Figure~\ref{fig:rl_sft_acc_ood_evolution}, the OOD accuracy still increases during checkpoint $50$ to $140$. Such \textbf{loss-accuracy discrepancy} exists for both LLaMA and Qwen. After going through the training and test data as shown in Appendix~\ref{appendix:task_description} during these checkpoints, we found that such discrepancy is caused by OOD rule forgetting and OOD reasoning enhancement. To be more specific, after the completion of format alignment at checkpoint $50$, the model starts to suffer from over-specification to the ID rule, failing to turn $'J, Q, K'$ as number $11, 12, 13$, \ie{} error in "number" step in OOD response will increase. The failure of "number" step will be very likely to cause failure in "formula" step, which will result in large OOD cross-entropy loss. However, during checkpoint $50$ to $140$, the arithmetic reasoning ability keep improving, \ie{} once the model succeed to interpret $'J, Q, K'$ as number $11, 12, 13$, the model has much higher probability to get a correct "formula". But compared with the increased loss in both "number" and "formula" steps, the improved accuracy in "formula" step will only cause a relative smaller decline of loss. So overall, in such mixture of status, we will observe and increased OOD loss together with increased OOD accuracy. From another perspective, the {loss-accuracy discrepancy} tells us that the token-level cross-entropy loss cannot fully reflect the real reasoning capacity of model.

\begin{figure}[htbp]
  \begin{subfigure}[t]{0.48\linewidth}
    \centering
    \includegraphics[width=\linewidth]{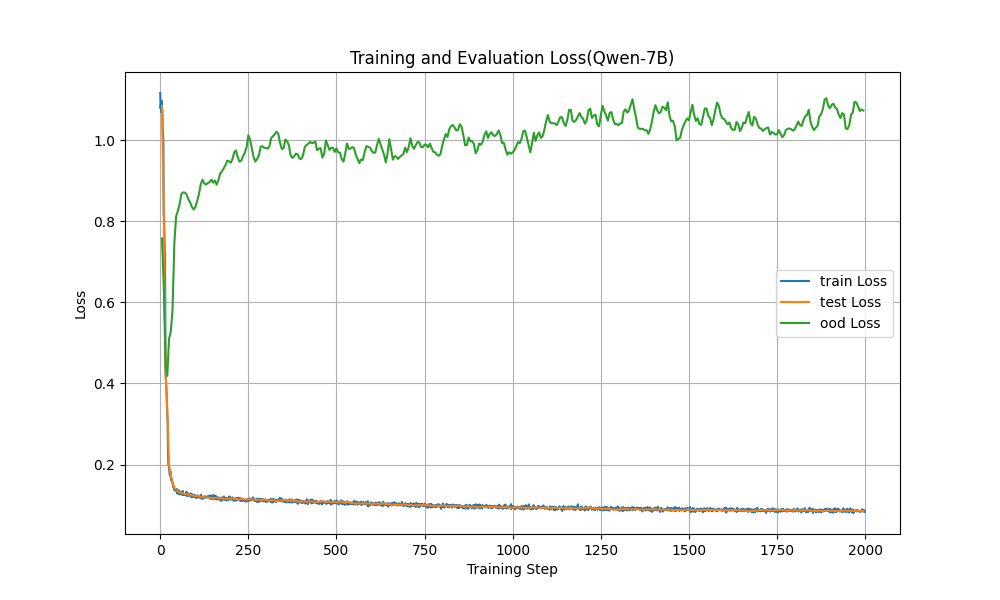}
    \label{fig:first}
  \end{subfigure}
  \hfill                           
  \begin{subfigure}[t]{0.48\linewidth}
    \centering
    \includegraphics[width=\linewidth]{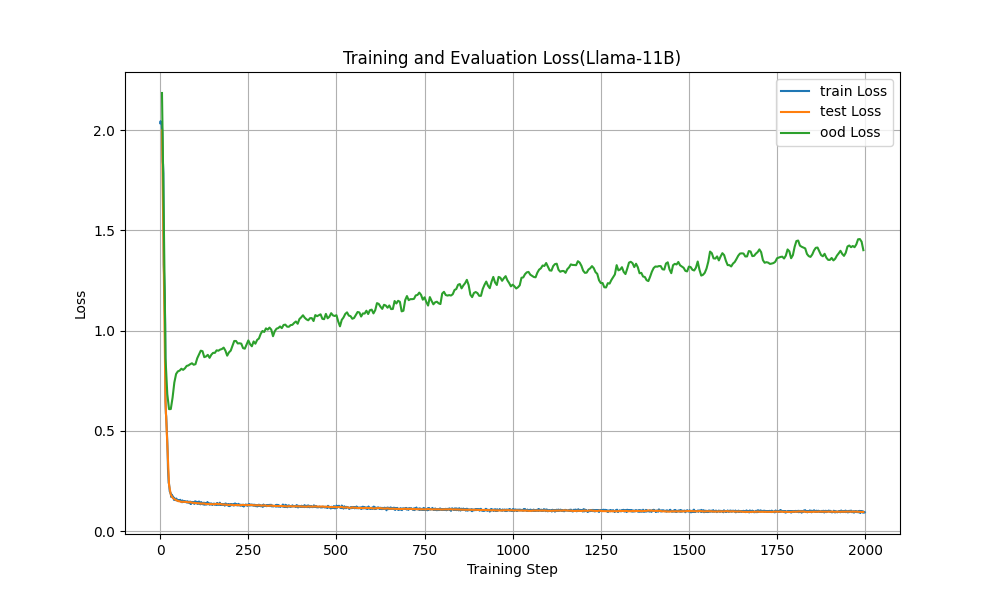}
    \label{fig:second}
  \end{subfigure}
  \caption{In-distribution training/test loss and OOD loss curves for LLaMA-3.2-11B-Vision and Qwen-2.5-7B during SFT.}
  \label{fig:sft_loss_ood}
\end{figure}

\subsection{Results with GRPO}
\label{appendix:results_grpo}
\begin{figure}[htbp]
  \centering
    \includegraphics[width=\linewidth]
    {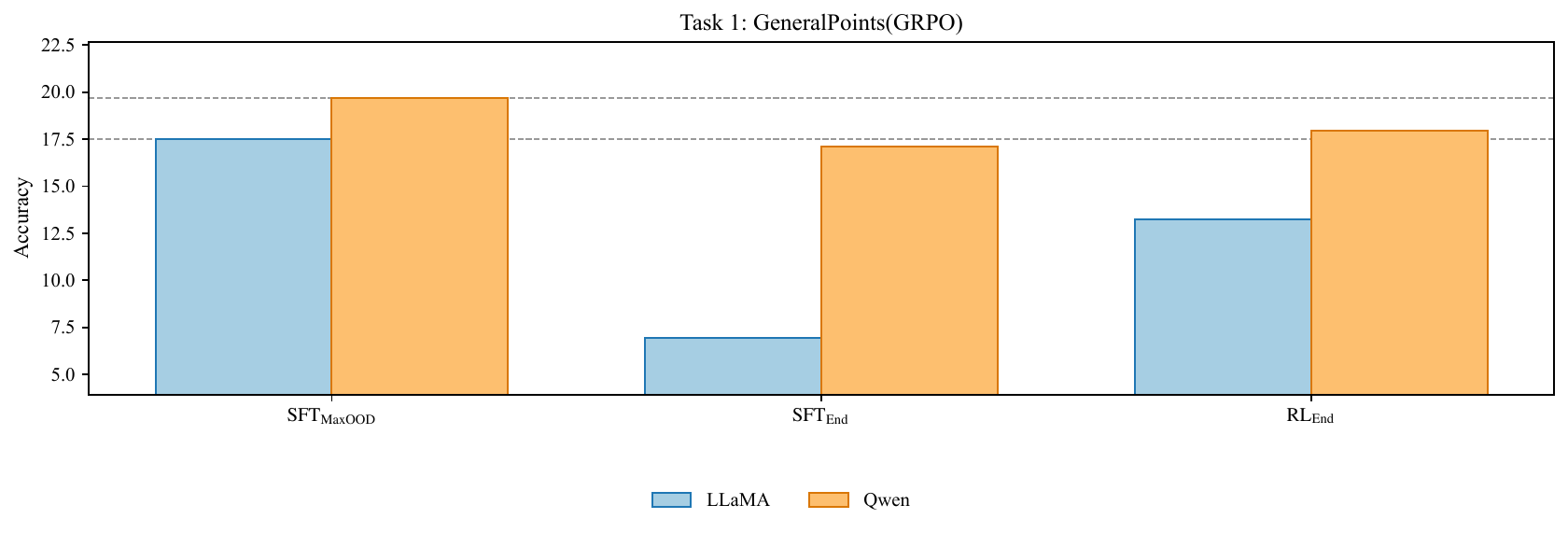}
  \caption{Results of GRPO on GeneralPoints (OOD)}
  \label{fig:gp_ood_acc_grpo}
\end{figure}

\begin{figure}[htbp]
  \centering
    \includegraphics[width=\linewidth]
    {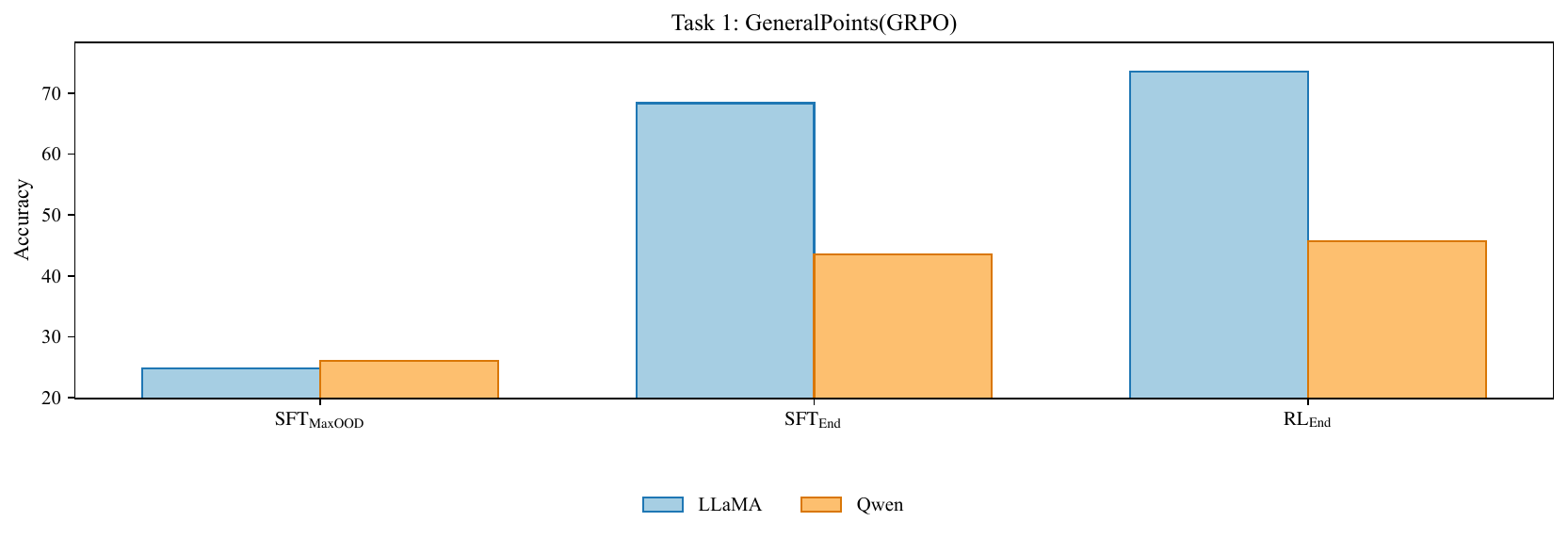}
  \caption{Results of GRPO on GeneralPoints (ID)}
  \label{fig:gp_id_acc_grpo}
\end{figure}

We have verified our claims with GRPO (group size = 4) and the OOD and ID results of LLaMA and Qwen on GeneralPoints are shown in Figure~\ref{fig:gp_ood_acc_grpo} and Figure~\ref{fig:gp_id_acc_grpo}. Interestingly, GRPO performs better than PPO in terms of ID performance and worse than PPO for OOD; and overall, both algorithms align with our claim that RL heals OOD forgetting but does not surpass the best of SFT.

\subsection{More Results of In-Distribution Generalization Performance}
\label{appendix:results_ID}
\begin{figure}[htbp]
  \centering
    \centering
    \includegraphics[width=\linewidth]
    {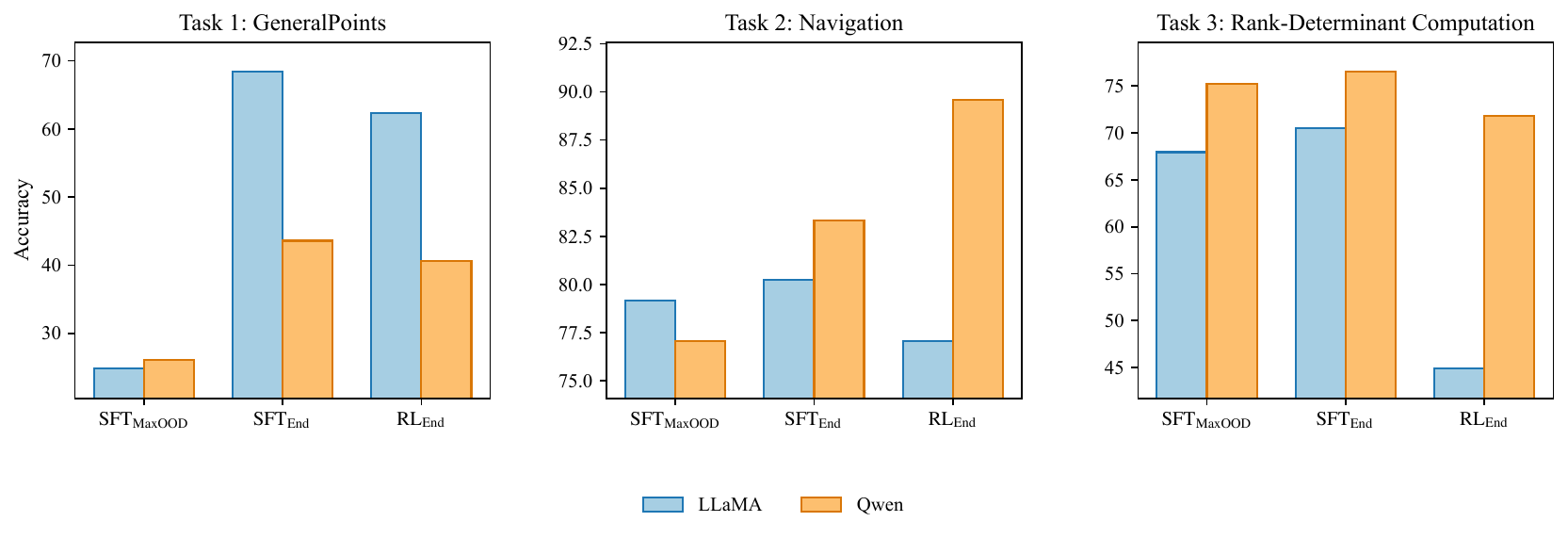}
    \caption{ID performance on three tasks}
  \label{fig:rl_sft_id_acc_comp}
\end{figure}

The in-distribution performance on \textit{GeneralPoints, Navigation} and \textit{Rank-Determinant Computation} are shown in Figure~\ref{fig:rl_sft_id_acc_comp}.
\newpage
\subsection{OOD Results on Other Benchmark Tasks}
\label{appendix:results_other_tasks}

\begin{table}[htbp]
  \centering
  \caption{OOD results at checkpoints SFT$_\text{MaxOOD}$, SFT$_\text{End}$ and RL$_\text{End}$ on five more benchmark datasets for LLaMA and Qwen}
    \begin{tabular}{c|c|ccc}
    \toprule
    \toprule
    \multicolumn{2}{c|}{\multirow{2}[2]{*}{Model/Tasks}} & \multicolumn{3}{c}{Checkpoints} \\
    \multicolumn{2}{c|}{} & SFT$_\text{MaxOOD}$ & SFT$_\text{End}$ & RL$_\text{End}$ \\
    \midrule
          & ARC-Challenge & 80.00 & 70.00 & 80.00 \\
          & CommonsenseQA & 81.00 & 62.00 & 69.00 \\
    LLaMA & GPQA  & 44.00 & 32.00 & 44.00 \\
          & IFEval-loose & 27.00 & 27.00 & 27.00 \\
          & MMLU-Pro & 46.00 & 36.00 & 37.00 \\
    \midrule
          & ARC-Challenge & 92.00 & 88.00 & 88.00 \\
          & CommonsenseQA & 79.00 & 73.00 & 75.00 \\
    Qwen  & GPQA  & 44.00 & 42.00 & 44.00 \\
          & IFEval-loose & 63.00 & 67.00 & 70.00 \\
          & MMLU-Pro & 73.00 & 66.00 & 66.00 \\
    \bottomrule
    \bottomrule
    \end{tabular}%
  \label{tab:additional_tasks}%
\end{table}%

We have verified our claims on other diverse benchmark datasets: ARC-Challenge~\cite{clark2018think}, CommonsenseQA~\cite{talmor2019commonsenseqa}, GPQA~\cite{rein2024gpqa}, IFEval-loose~\cite{zhou2023instruction}, MMLU-Pro ~\cite{wang2024mmlu}. The results are shown in Table~\ref{tab:additional_tasks} and they align with our observations that the RL heals the OOD forgetting in SFT but barely surpasses the best of SFT. This indicates that our conclusion is robust and generalizable across different tasks.

\subsection{Loss of Single-Stage RL Fine-tuning}
\label{app:loss_rl}
As summarized in Section~\ref{sec:related_work}, there are numerous studies that give completely different conclusions about the effectiveness of RL fine-tuning, especially for single-stage RL. So in this paper, we also verify RL fine-tuning without SFT as cold start.

\begin{figure*}[htbp]
    \centering
    \begin{subfigure}[b]{0.48\textwidth}
    \includegraphics[width=\textwidth]{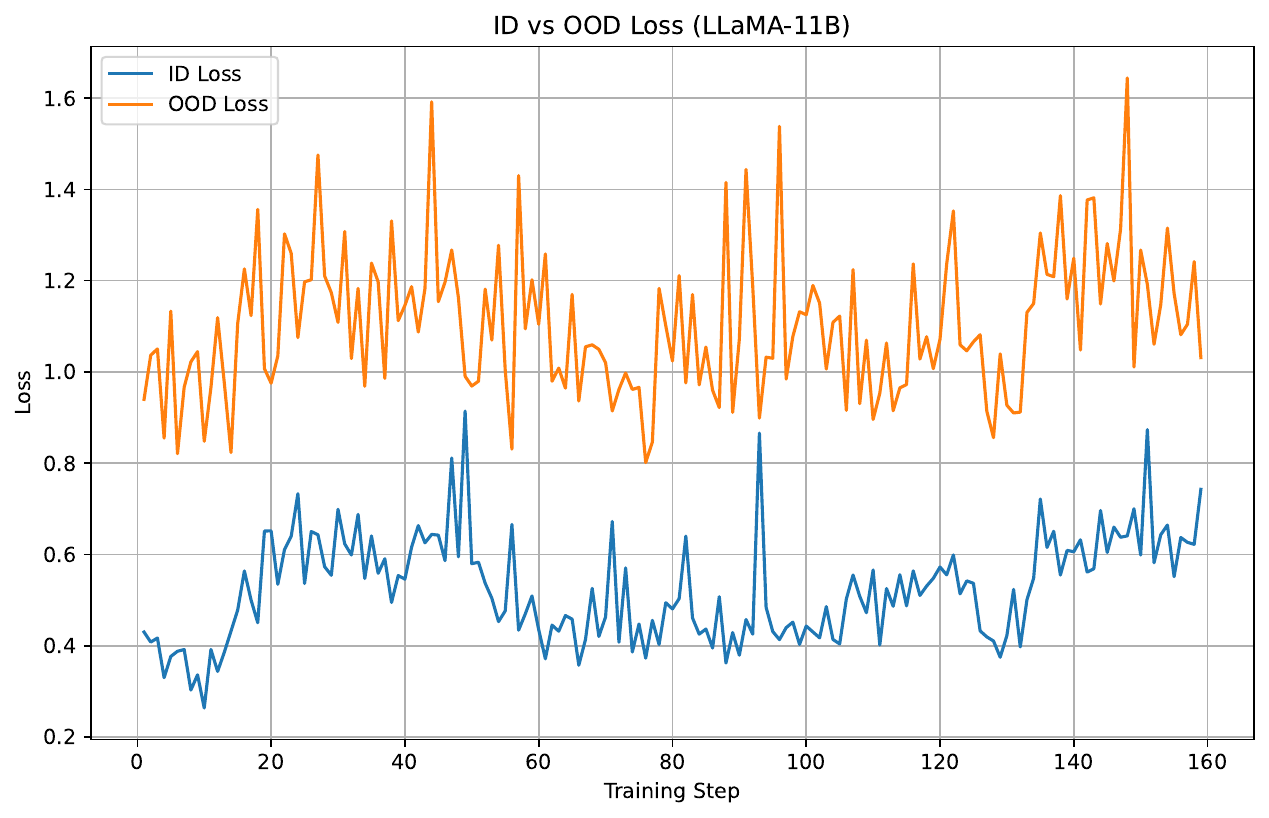}
    \captionsetup{justification=centering}
    \caption{Loss in LLaMA}
    \label{fig:rl_loss_LLaMA}
    \end{subfigure}
    \centering
    \begin{subfigure}[b]{0.48\textwidth}
    \includegraphics[width=\textwidth]{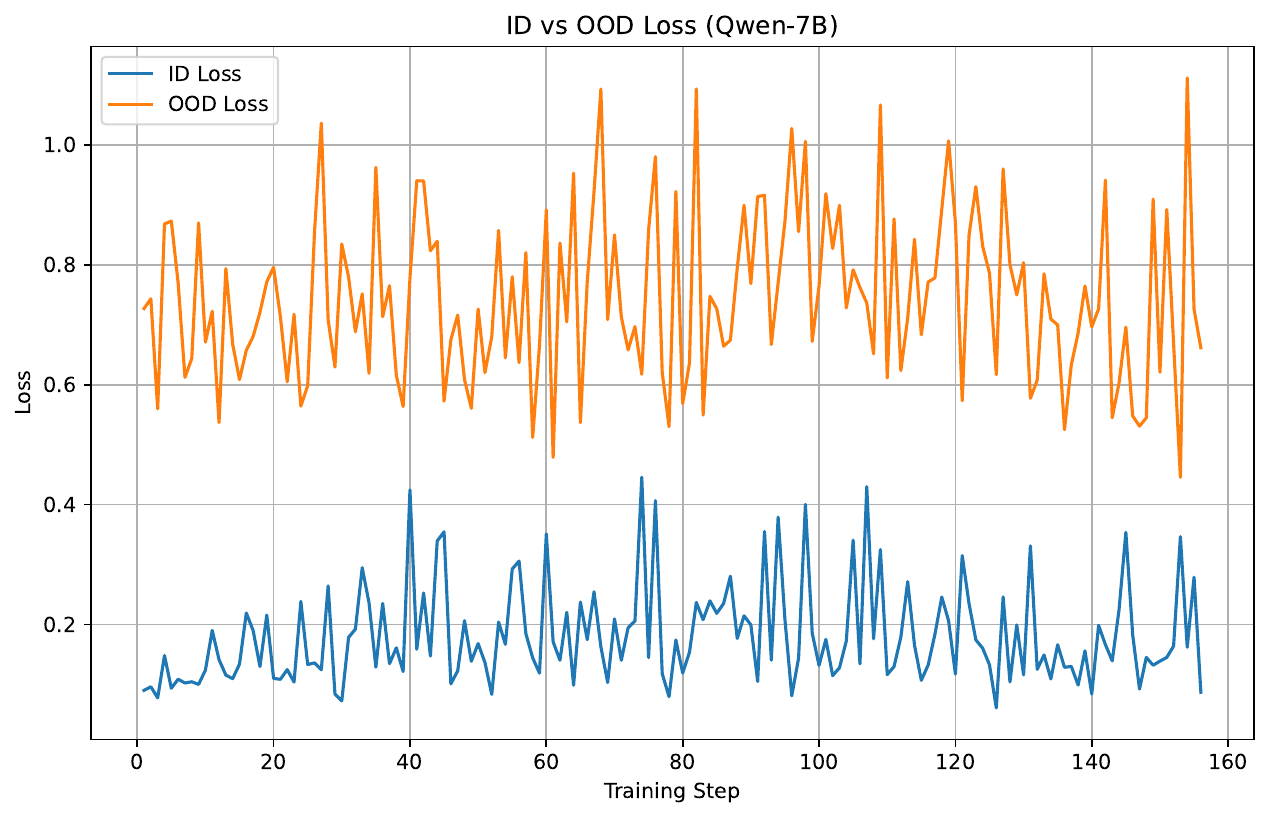}
    \captionsetup{justification=centering}
    \caption{Loss in Qwen}
    \label{fig:rl_loss_qwen}
    \end{subfigure}
    \caption{Loss of single-stage RL fine-tuning}
    \label{fig:rlft_loss}
\end{figure*}

From Figure~\ref{fig:rlft_loss}, we observe that RL can hardly converge without SFT. This is because the base model has poor task-following ability, which would give overwhelmingly low scores for RL, leading to unstable updates and collapse in training. On the other hand, SFT can provide a safe starting point and policy initialization, where the model can at least align the format and generate reasonable candidates for the reward model to evaluate.

\subsection{Examples for Reward Hacking}
\label{app:reward_hacking}
Inconsistent with previous research \citep{deepseekai2025deepseekr1incentivizingreasoningcapability}, as demonstrated below, reward hacking occurs when we fine-tune the models by pure RL from scratch or an early SFT checkpoint.
\begin{figure*}[htbp]
    \centering
    \includegraphics[width=0.85\textwidth]{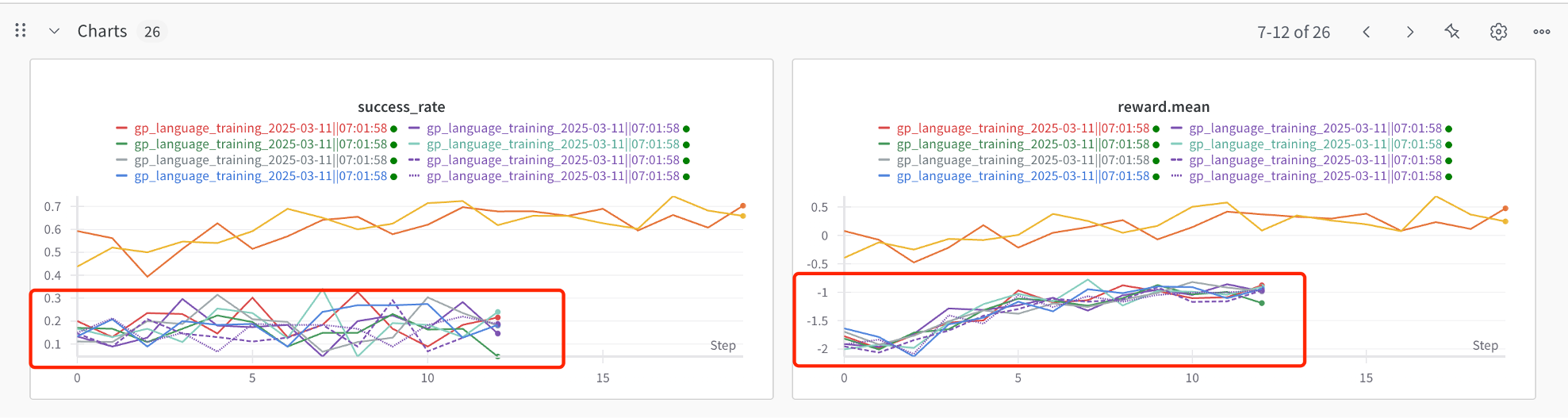}
    \captionsetup{justification=centering}
    \caption{An example of \emph{reward hacking}. The RL-only curve sees an increasing reward signal (right panel) but stagnant or low success rates (left panel).}
    \label{fig:curves}
\end{figure*}

\section{Additional SVD Analysis}
\label{appendix:additional_svd_analysis}

\subsection{SVD Notation}
\label{appendix:svd_notation}
For a parameter matrix $\bm{M}\in\mathbb{R}^{m\times n}$, we write its singular-value decomposition as $\bm{M}=\bm{U\Sigma V}^{\top}$, where $\bm{U}$ and $\bm{V}$ contain the left and right singular vectors and $\bm{\Sigma}$ contains non-negative singular values $\sigma_1\ge\sigma_2\ge\dots\ge0$. In our analysis, the singular values measure the strength of each mode and the singular vectors define its directions. We apply this decomposition to self-attention and MLP matrices in LLaMA and Qwen.

\subsection{Changes of Singular Values}
\label{app:singular_value_sft_rl}
\begin{figure}[ht]
  \centering
  \begin{subfigure}[b]{0.32\textwidth}
    \includegraphics[width=\linewidth]{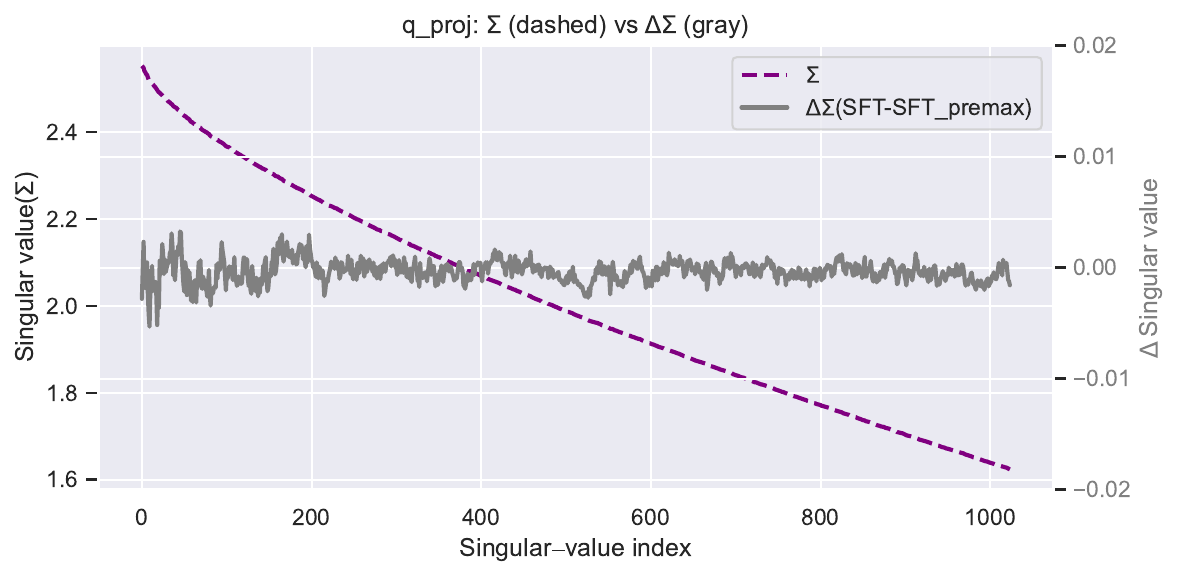}
    \caption{$W_q$ changes during SFT}
    \label{fig:1a}
  \end{subfigure}\hfill
  \begin{subfigure}[b]{0.32\textwidth}
    \includegraphics[width=\linewidth]{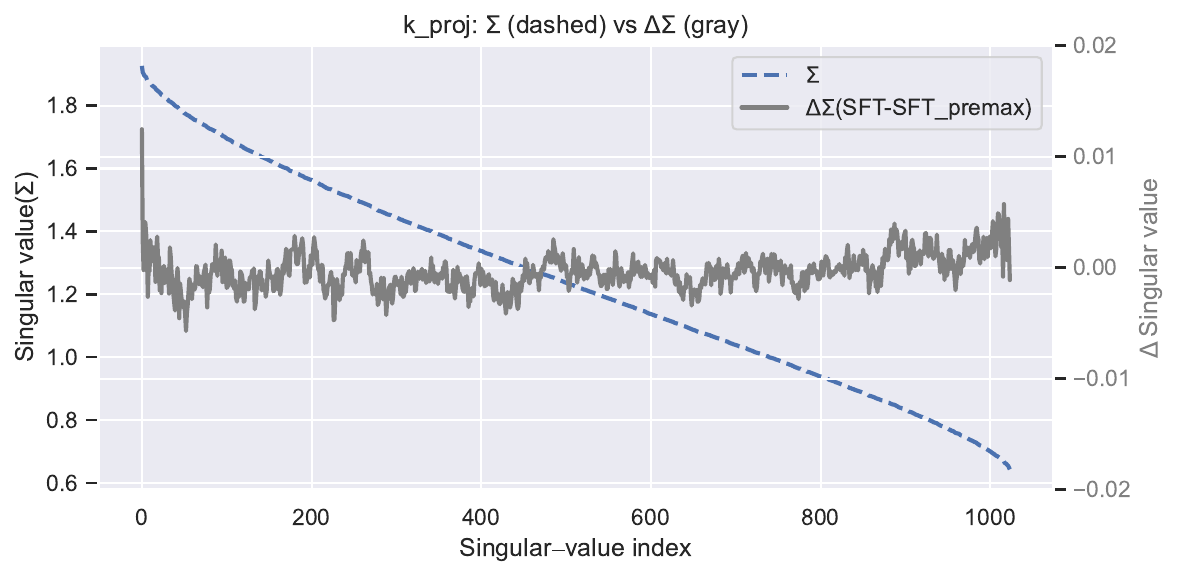}
    \caption{$W_k$ changes during SFT}
    \label{fig:1b}
  \end{subfigure}\hfill
  \begin{subfigure}[b]{0.32\textwidth}
    \includegraphics[width=\linewidth]{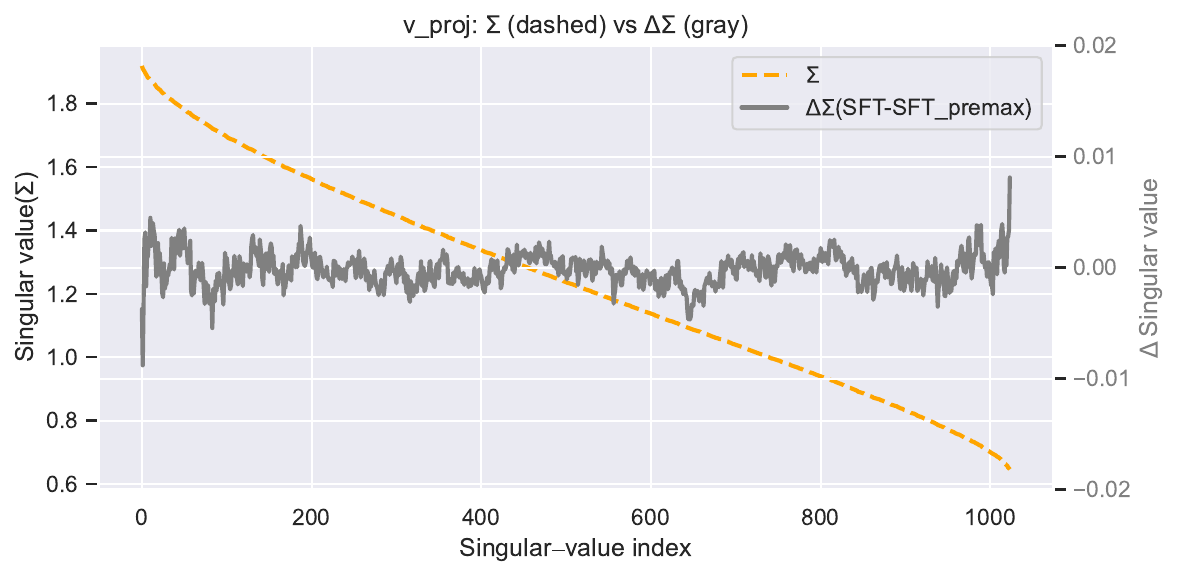}
    \caption{$W_v$ changes during SFT}
    \label{fig:1c}
  \end{subfigure}
  \begin{subfigure}[b]{0.32\textwidth}
    \includegraphics[width=\linewidth]{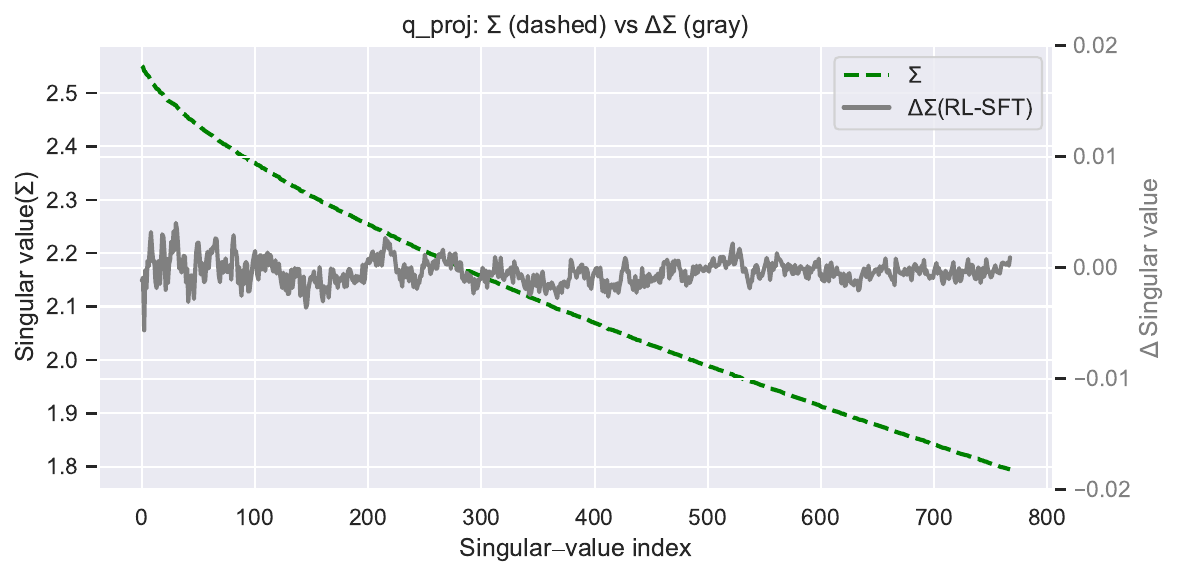}
    \caption{$W_q$ changes during RL}
    \label{fig:2a}
  \end{subfigure}\hfill
  \begin{subfigure}[b]{0.32\textwidth}
    \includegraphics[width=\linewidth]{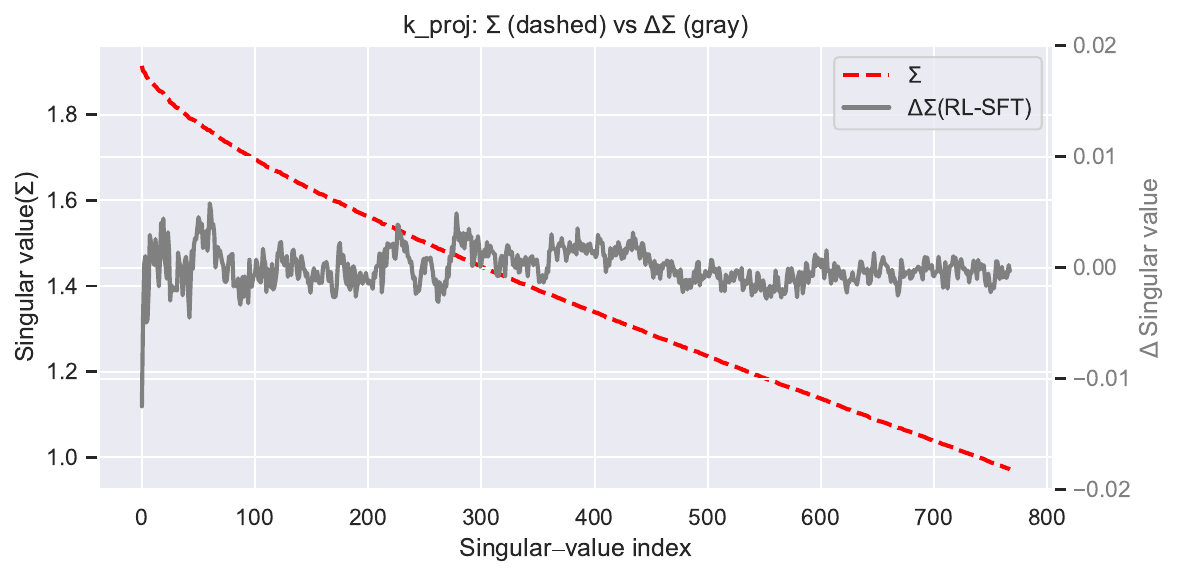}
    \caption{$W_k$ changes during RL}
    \label{fig:2b}
  \end{subfigure}\hfill
  \begin{subfigure}[b]{0.32\textwidth}
    \includegraphics[width=\linewidth]{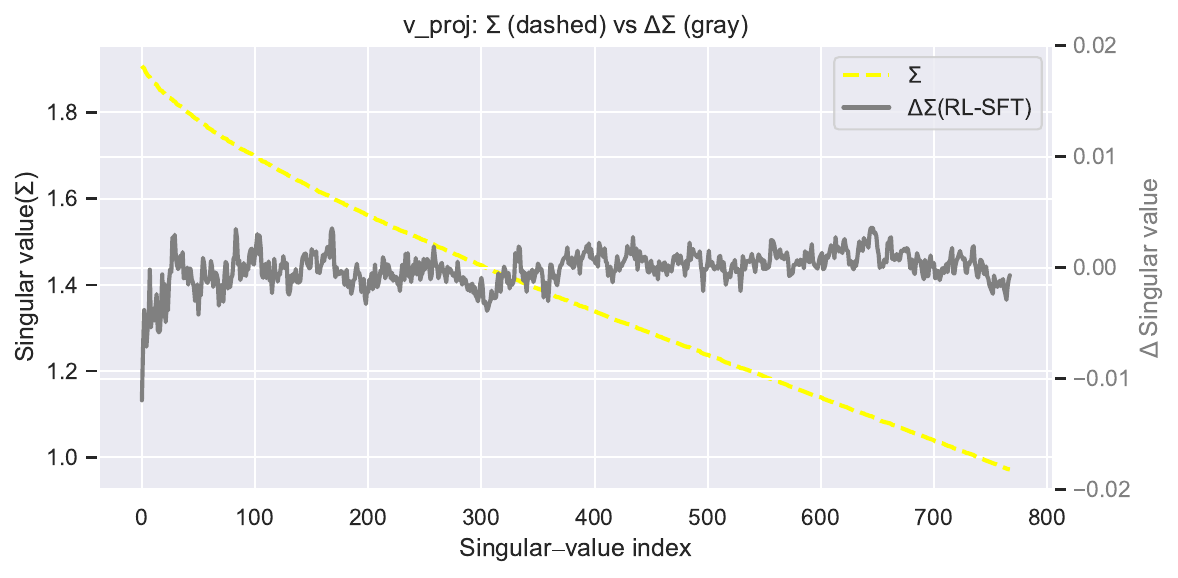}
    \caption{$W_v$ changes during RL}
    \label{fig:2c}
  \end{subfigure}
  \caption{
    Singular value changes in the \texttt{q\_{proj}}, \texttt{k\_proj}, and \texttt{v\_proj} matrices of the first self-attention layer (\texttt{layers[5].self\_attn}) in \texttt{LLaMA-3.2-11B-Vision}. Panels (a)--(c) illustrate the impact of supervised fine-tuning (SFT) on $W_q$, $W_k$, and $W_v$, respectively, while panels (d)--(f) depict the corresponding changes following reinforcement learning (RL). Each panel shows the difference in singular values before and after the respective post-training stage. For LLaMA, SFT starts from $\text{SFT}_\text{MaxOOD}$, RL stage begins from $\text{SFT}_\text{End}$.
    }
  \label{fig:singular_value_sft_rl}
\end{figure}

To investigate how does SFT and RL reshape the spectral structure of the parameter matrices, we analyze the singular values of $\bm{W_q, W_k, W_v}$ and their differences ($\Delta\sigma_i = \sigma_i^{\text{SFT}_{\text{End}}} - \sigma_i^{\text{SFT}_{\text{MAXOOD}}}$) before/after different training stage. The results are shown in the Figure~\ref{fig:singular_value_sft_rl}. We found that: \textbf{the changes of singular values of the $\bm{Q, K, V}$ matrices are negligible after both SFT and RL stages across all experiments}. Compared to the original singular values, $\Delta\sigma$ fluctuates from $0$ to $0.005$, which acts similar as a low-magnitude, zero-centered noisy signals. This indicates that the fine-tuning process does not significantly amplify or diminish specific singular values. 

\subsection{Exploring the Rotation of Singular Vector with Principal Angles}
\label{app:principle angle_singular_vectors}

There exists two ways to measure the changes of singular vectors during fine-tuning: vector-level metrics and subspace-level metrics.


Principal angles (or canonical angles) quantify how far two subspaces are within the same Euclidean space. To quantify the differences between the subspaces spanned by the singular vectors of base model $\bm{W}_{\text{Base}}$ and fine-tuned model $\bm{W}_{\text{FT}}$, we measure the amount of rotations between two subspaces by how much their dominant singular vector directions have \emph{rotated}, which is a commonly used method in machine learning~\citep{huang2015role, vahidian2023efficient} and numerical computation \citep{bjorck1973numerical}. We provide a brief introduction and we take the left singular vectors for example and the computation includes,

\paragraph{(i) SVD.}
For each matrix, we keep all singular vectors in our experiments,
\[
  W
  = U\,\Sigma\,V^{\top},
  \qquad
  U\in\mathbb{R}^{m\times r},\;
  V\in\mathbb{R}^{n\times r},
  \tag{2.4.1}
\]
where the columns of \(U\) and \(V\) are orthonormal and
\(\Sigma=\operatorname{diag}(\sigma_1,\dots,\sigma_r)\) with
\(\sigma_1\!\ge\!\dots\!\ge\!\sigma_r\!\ge\!0\), $r$ is the rank.

\paragraph{(ii) Computation of Principal Angles Between Subspaces (PABS).}
Let \(\bm{U}_{\text{Base}}, \bm{U}_{\text{FT}}\in\mathbb{R}^{m\times k}\) be the left singular blocks from the previous step.  Define
\(\! \bm{M} := \bm{U}_{\text{Base}}^{\!\top} \bm{U}_{\text{FT}} \in \mathbb{R}^{r \times r} \).
Since both of them are orthonormal, the singular values of
\(M\) lie in \([-1,1]\)~\citep{bjorck1973numerical}.  Suppose the SVD of $\bm{M}$ is 
\[
  \bm{M} = \bm{U}_M\,\text{diag}(s_1,\dots,s_r)\, \bm{V}_M^{\!\top},
\]
the \emph{principal angles} \(\theta_i\in[0,\pi/2]\) between $\bm{U}_{\text{Base}}^{\!\top}$ and $\bm{U}_{\text{FT}}$ are
\begin{equation}
  \theta_i \;=\; \arccos(s_i), 
  \quad i=1,\dots,r.
  \tag{2.4.2}
\end{equation}
The computational complexity is \(O(\min\{m,n\}^3)\). An identical procedure on \(\bm{V}_{\text{Base}}, \bm{V}_{\text{FT}}\) yields angles for the right subspaces. In practice we clamp the numerical values of \(s_i\) to \([-1,1]\) before calling \(\arccos\) to avoid floating-point overflow. The Principal angles measure the 'tilt' between corresponding singular vectors of two matrices, \ie{} the degree to which two parameter matrices are different from each other in terms of singular vectors under the rank $r$. The angle set \(\{\theta_i\}\) serves as a fine-grained measure of subspace rotation: \(\theta_i=0\) means the \(i\)-th principal direction is preserved, whereas values approaching \(\pi/2\) indicate maximal misalignment.  

\paragraph{Advantages of PABS}
\begin{itemize}
    \item \textbf{Numerical Stability:} Consider when two singular values are very close and their corresponding singular vectors are orthogonal. After one step of SFT, the singular values and vectors might only make subtle shifts but the singular values might swap orders. Therefore, the pairwise cosine similarity might demonstrate a very large angle, while the parameter matrices only make subtle changes. Therefore, vector-level metrics are not as robust as subspace-level metrics like PABS.
    \item Cosine similarity between singular vectors only compares one dimension at a time, without accounting for interdependence between directions. PABS derives angles that reflect the relative orientation of the entire subspace, providing a more informative measure than isolated vector-to-vector comparisons.
    \item PABS is a true metric for comparing subspaces, ideal for measuring alignment or divergence holistically.
\end{itemize}

\begin{figure*}[htbp]
    \centering
    \includegraphics[width=0.75\textwidth]{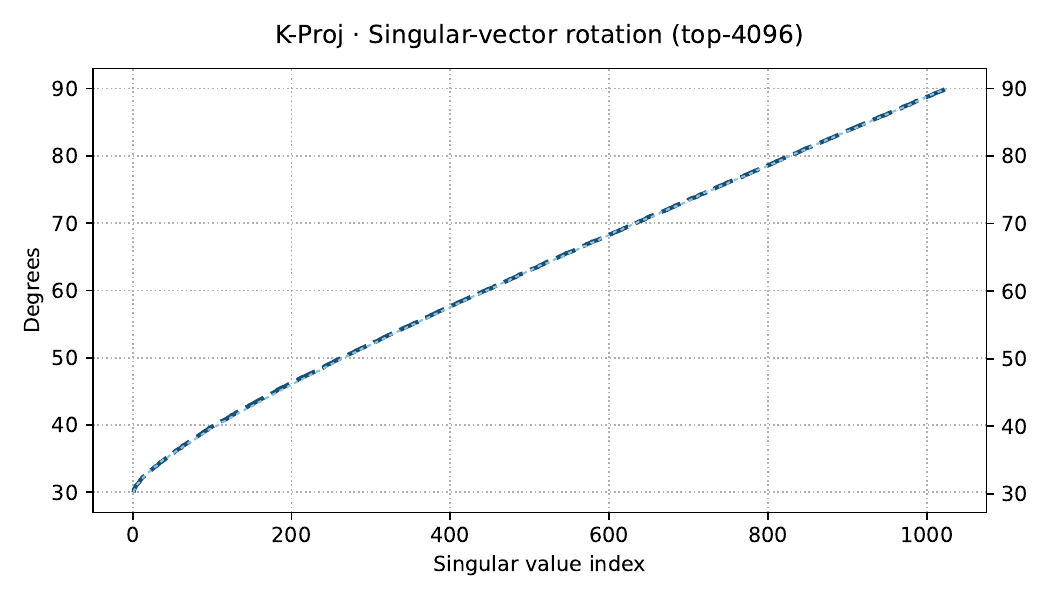}
    \captionsetup{justification=centering}
    \caption{An example of rotation between SFT and RL.}
    \label{fig:principle_angle}
\end{figure*}

We use principle angle to analyze the pattern of subspace rotation during SFT and RL. To this end, we calculate the principal angle spectrum of the layer-0 $k_\text{proj}$ matrix between checkpoint $0$ vs. $\text{SFT}_\text{End}$, and checkpoint $0$ and $\text{RL}_\text{End}$, and plot them in Figure~\ref{fig:principle_angle}. For both SFT and RL, the two monotonically increasing curves overlap each other: the smallest angle is around $25-30$ degrees and the angles increase smoothly and linearly toward $90$ degree in the tail. 

These curves imply that both of the two fine-tuning stages adjust the model primarily by rotating its singular vectors, which is already verified in Section~\ref{sec:param_matrices_analysis}. However, we cannot find out the differences in their rotation patterns. The exact mechanism of the rotation patterns remains unresolved and understanding the two fine-tuning behaviors in parameter space, especially in high-dimensional space, is an open question that we will investigate in future work.

 
 
\newpage

\subsection{PCA Visualization of Embedding Shifts}
\label{Appendix:visualization_pca}
\begin{figure}[htbp]
  \centering
  \begin{subfigure}[b]{0.48\textwidth}
    \includegraphics[width=\linewidth]{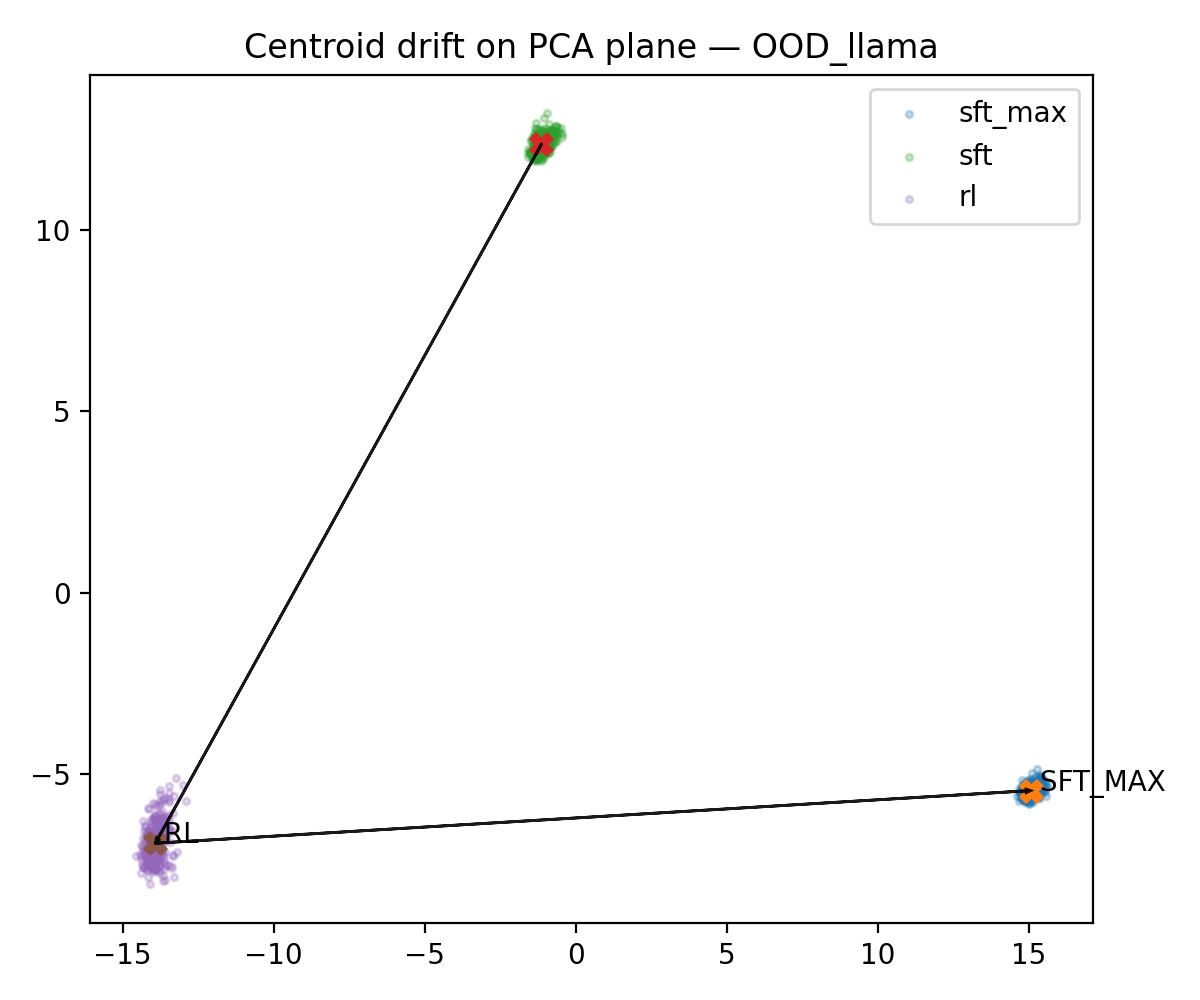}
    \caption{OOD hidden states for LLaMA}
    \label{fig:pca_ood_llama}
  \end{subfigure}\hfill
  \begin{subfigure}[b]{0.48\textwidth}
    \includegraphics[width=\linewidth]{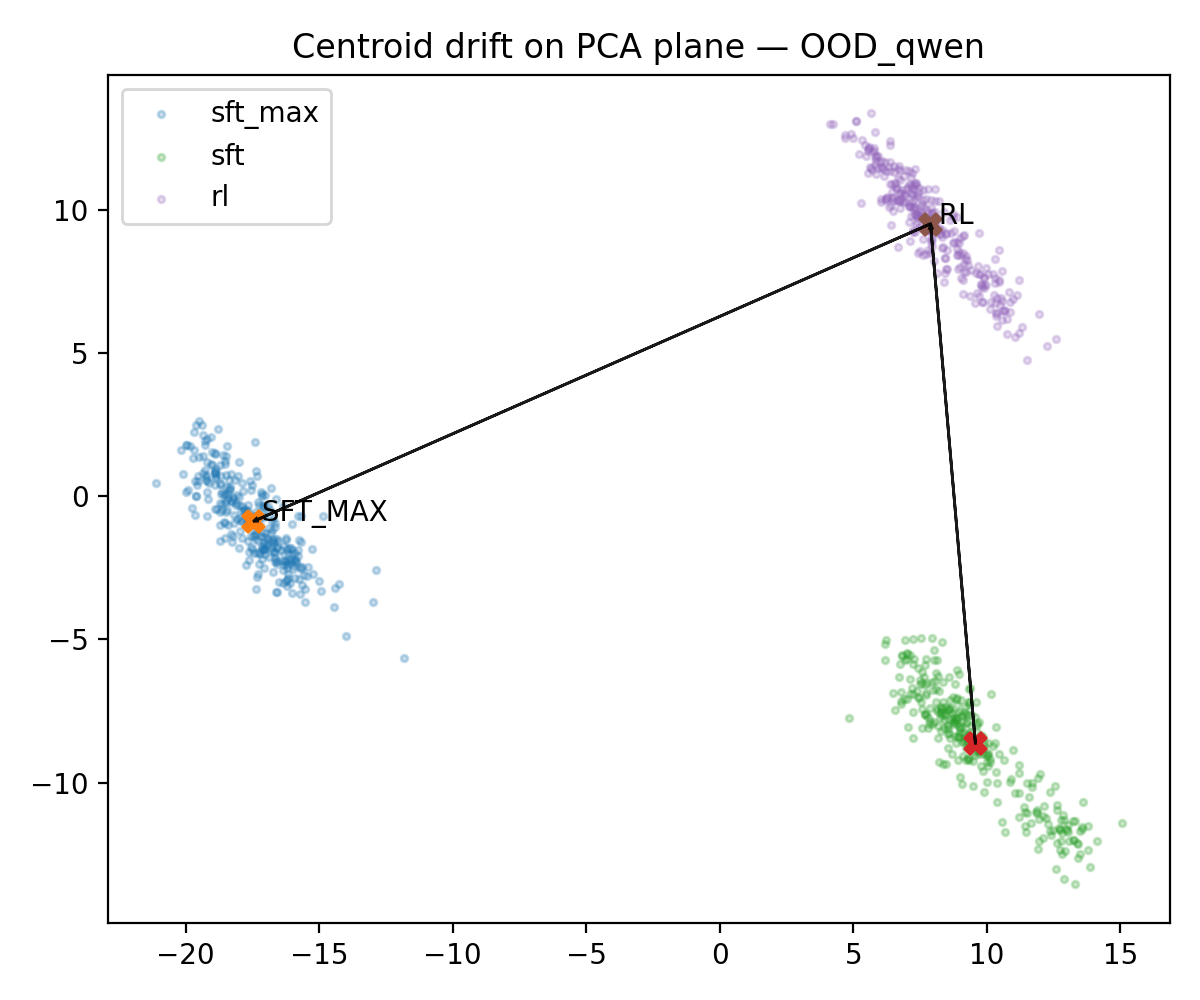}
    \caption{OOD hidden states for Qwen}
    \label{fig:pca_ood_qwen}
  \end{subfigure}\hfill
  \begin{subfigure}[b]{0.48\textwidth}
    \includegraphics[width=\linewidth]{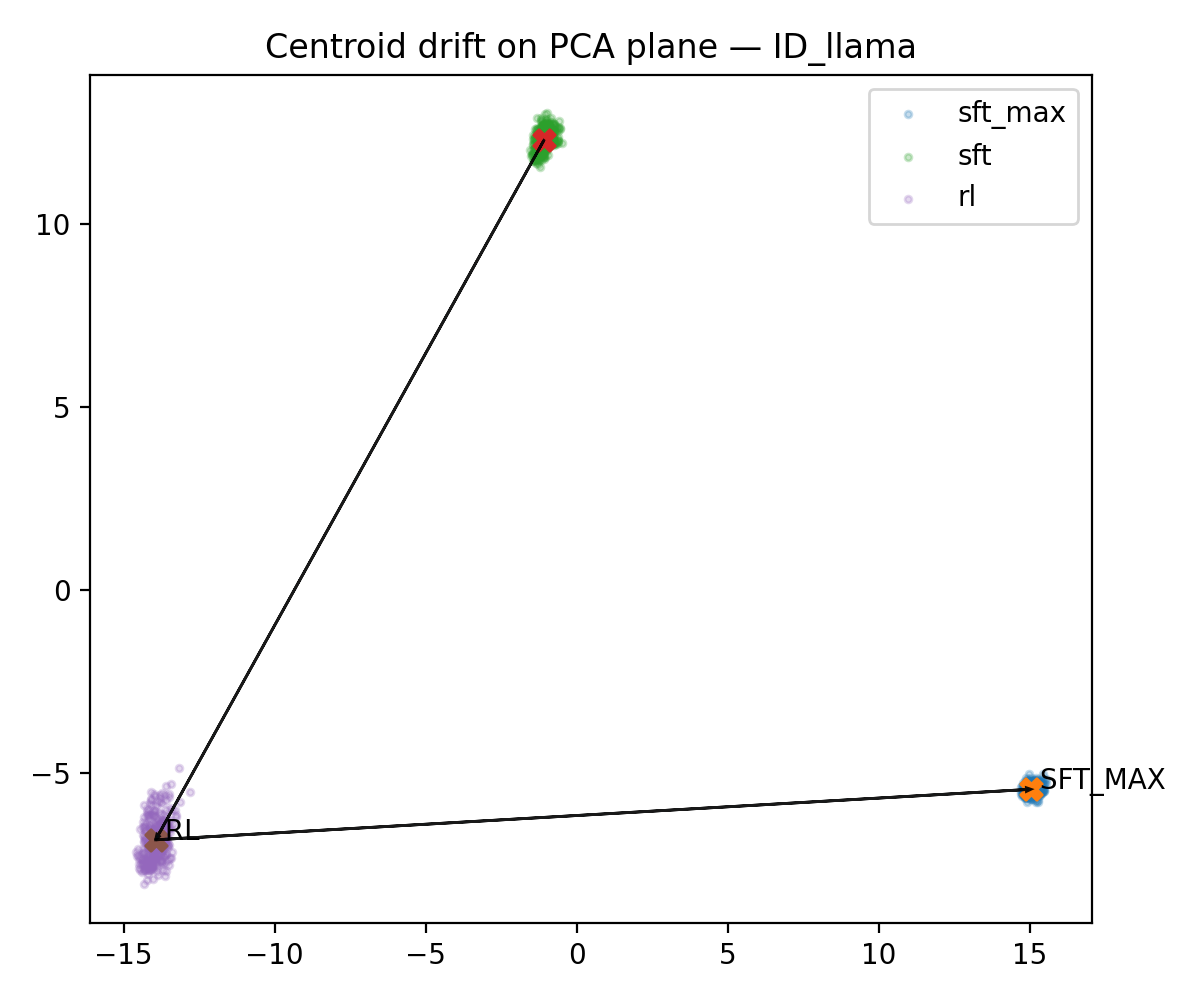}
    \caption{ID hidden states for LLaMA}
    \label{fig:pca_id_llama}
  \end{subfigure}\hfill
  \begin{subfigure}[b]{0.48\textwidth}
    \includegraphics[width=\linewidth]{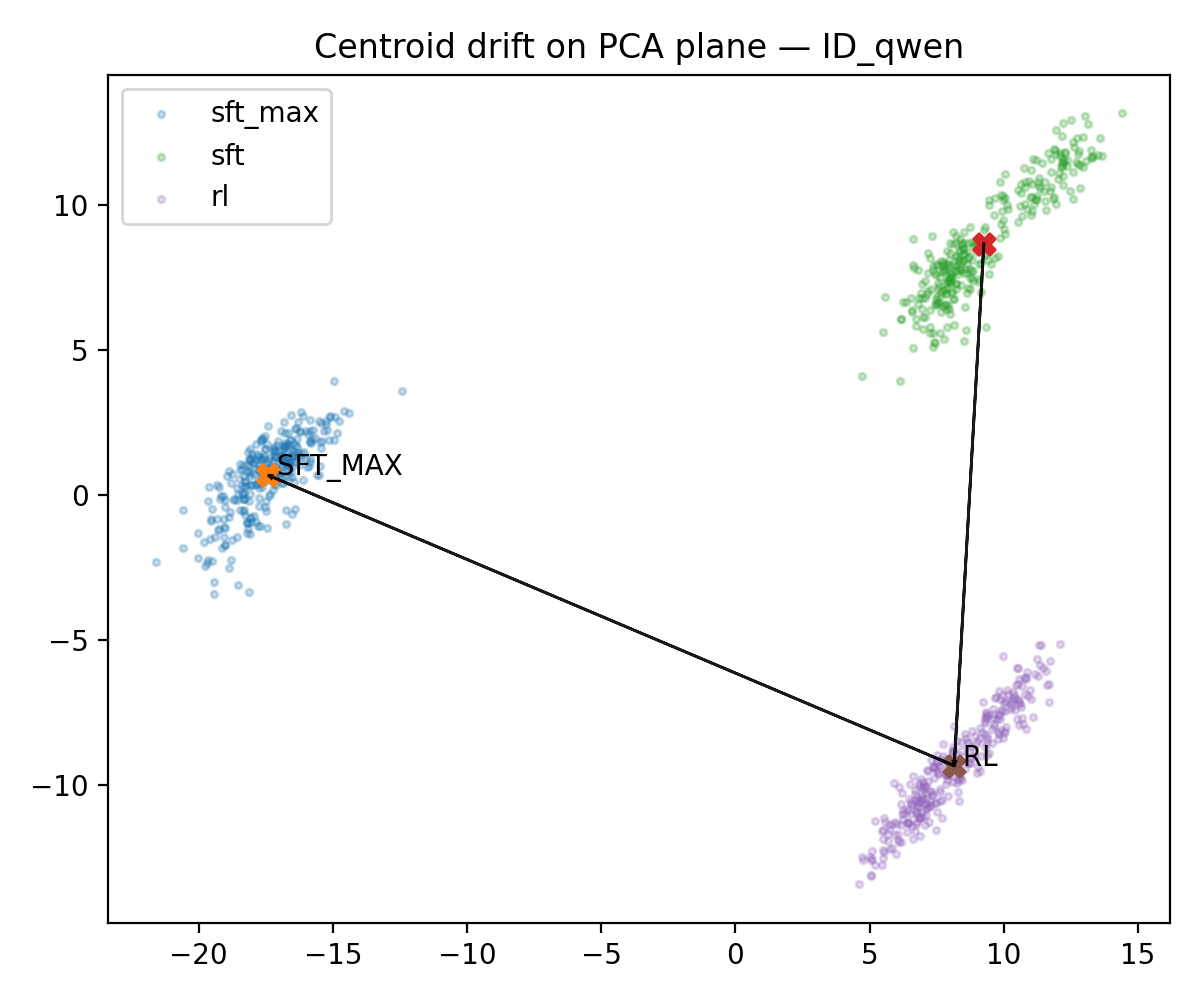}
    \caption{ID hidden states for Qwen}
    \label{fig:pca_id_qwen}
  \end{subfigure}\hfill
  \caption{PCA visualization of the hidden representations at checkpoints $\text{SFT}_\text{MaxOOD}$, $\text{SFT}_\text{END}$ and $\text{RL}_\text{END}$.}
  \label{fig:pca}
\end{figure}
We use 300 in-distribution prompts and 300 out-of-distribution prompts to activate hidden states respectively at certain fine-tuning checkpoint, compute PCA for the representation matrix and use the first two principle components to visualize the embedding shifts for both models. We find that RL fine-tuning slightly drags the hidden representation away from the $\text{SFT}_\text{MaxOOD}$, \ie the embedding distance between $\text{RL}_\text{END}$ and $\text{SFT}_\text{MaxOOD}$ is farther than the $\text{SFT}_\text{END}$ and $\text{SFT}_\text{MaxOOD}$. The representation shift for Qwen is smaller than LLaMA. This also indicates Qwen is a more robust model than LLaMA during SFT and RL fine-tuning.

\subsection{Potential OOD forgetting explanation}
\label{Appendix:ood_forgetting_explanatino}

\paragraph{Setting}
We do SFT with cross-entropy (CE) on the train set.
We evaluate OOD on 24 problems. For each problem we sample up to 6 attempts.
Loss is CE on train tokens. OOD accuracy is pass@6 over the 24 OOD problems and reported as a percentage.

\paragraph{Verifier and check order (per step)}

1. Format parse: if parse fails $\rightarrow$ ILLEGAL\_FORMAT. Stop other checks for that step.\\
2. Number check: if numbers in formula are invalid (not from set, wrong count, etc.) $\rightarrow$ INCORRECT\_NUMBER.\\
3. Solution check: if no valid ``final answer'' after format checking $\rightarrow$ NO\_SOLUTION.\\
4. Aggregation: if $\ge 2$ of the above are true for a step $\rightarrow$ also count AGGREGATED\_ERR.

\paragraph{How we compute metrics}

- Loss: mean token-level CE on train data (OOD).\\
- CORRECT\_SOLUTION(CS): a problem is correct if any of its 6 attempts ends with a correct final answer;
accuracy is the percentage over 24.\\
- Step-level rates (NO\_SOLUTION(NS), ILLEGAL\_FORMAT(IF), INCORRECT\_NUMBER(IN), AGGREGATED\_ERR(AE)): count steps with
the label divided by all steps. Rates can co-occur and do not sum to 100\%.

\begin{table}[htbp]
\centering
\small
\begin{tabular}{rcccccc}
\hline
ck & CS & NS & IF & IN & AR & Loss \\
\hline
10  & 0.00  & 18.95 & 80.06 & 0.78 & 0.21 & 1.5411 \\
20  & 1.28  & 33.57 & 54.62 & 7.59 & 4.01 & 0.6293 \\
30  & 10.26 & 63.68 & 0.99  & 20.59 & 12.92 & 0.4262 \\
40  & 9.40  & 84.07 & 0.23  & 7.13 & 6.90 & 0.5994 \\
50  & 12.39 & 91.09 & 0.08  & 4.30 & 2.30 & 0.6944 \\
60  & 16.24 & 90.95 & 0.08  & 3.70 & 2.28 & 0.7076 \\
70  & 16.24 & 86.44 & 0.00  & 5.34 & 5.27 & 0.7211 \\
80  & 13.25 & 90.80 & 0.00  & 4.49 & 2.32 & 0.7273 \\
90  & 14.10 & 82.64 & 0.00  & 7.97 & 6.80 & 0.7212 \\
100 & 17.09 & 87.76 & 0.00  & 4.74 & 4.34 & 0.7166 \\
110 & 15.38 & 88.54 & 0.00  & 5.69 & 2.92 & 0.7267 \\
120 & 15.38 & 91.82 & 0.00  & 4.17 & 1.18 & 0.7350 \\
130 & 15.81 & 90.66 & 0.00  & 3.14 & 3.30 & 0.7625 \\
140 & 17.52 & 90.13 & 0.00  & 3.82 & 2.87 & 0.7660 \\
150 & 14.96 & 92.55 & 0.00  & 2.90 & 1.80 & 0.7581 \\
\hline
\end{tabular}
\caption{OOD accuracy (trajectory-level pass@6 on 24 problems), step-level error rates, and train CE
loss.}
\end{table}

\paragraph{Observation}

- Loss is lowest at ck=30 (0.426) and later rises.\\
- CORRECT (OOD) grows from 0.00\% (ck=10) to $\approx 17.00\%$ (ck=100--150).\\
- ILLEGAL\_FORMAT drops to 0.00\% by ck$\ge$70 (better syntax).\\
- INCORRECT\_NUMBER falls after peaking near ck=30 (better number use).\\
- AGGREGATED\_ERR stays low and trends down.

\paragraph{Why loss and OOD move differently} 

- Different targets: CE fits train tokens; CORRECT measures end-to-end success on OOD with a verifier
and pass@6 with verifier feedback.\\
- Structure over tokens: cleaner format and number use can boost pass@6 even if CE rises.\\
- Search effect: as more attempts are valid, at-least-one-success increases.\\
- Apparent ``forgetting'': later checkpoints may drift from train token distribution (higher CE) while
generalizing structure better on OOD (higher CORRECT).

\subsection{Advantage Distribution Metrics}
\label{appendix:advantage_metric_definitions}

For each SFT checkpoint $k$, we collect the advantage samples produced during RL rollouts and compute basic statistics: center $\mu_k$, standard deviation $\sigma_k$, and skewness~\citep{zwillinger1999crc}. To capture distribution shape beyond the first moments, we also use entropy and KL divergence to a matched normal distribution.

\paragraph{Entropy}
We use entropy to measure flatness and uniformity. Flat advantage distributions without pronounced modes have higher entropy than sharply concentrated distributions~\citep{petty2018some}. For a discrete approximation $\hat p_k$ over bins $b$, we compute
\[
H(\hat p_k) = -\sum_b \hat p_k(b)\log \hat p_k(b).
\]
For finite discrete support, the uniform distribution has maximum entropy~\citep{hyvarinen1997new}.

\paragraph{KL divergence to a matched normal distribution}
We use KL divergence to measure non-Gaussian structure~\citep{hyvarinen1997new,hyvarinen2000independent}, or negentropy~\citep{hyvarinen2013independent}. For checkpoint $k$, the matched normal distribution is $q_k = N(\mu_k,\sigma_k^2)$, where $\mu_k$ and $\sigma_k$ are estimated from the advantage samples. For the same discrete approximation, we compute
\[
D_{\mathrm{KL}}(\hat p_k\|q_k) = \sum_b \hat p_k(b)\log \frac{\hat p_k(b)}{q_k(b)}.
\]
Given fixed second moment or variance, a Gaussian maximizes differential entropy~\citep{david2017information}, so larger KL indicates more non-Gaussian structure, such as sharper peaks, multiple modes, heavier tails, or asymmetry~\citep{freeman2013assessing,silverman1981using}.

\subsection{Full Results of Advantage Distribution}
\label{Appendix:advantage_comparison}
\begin{figure}[ht]
  \begin{subfigure}[t]{0.32\linewidth}
    \centering
    \includegraphics[width=\linewidth]{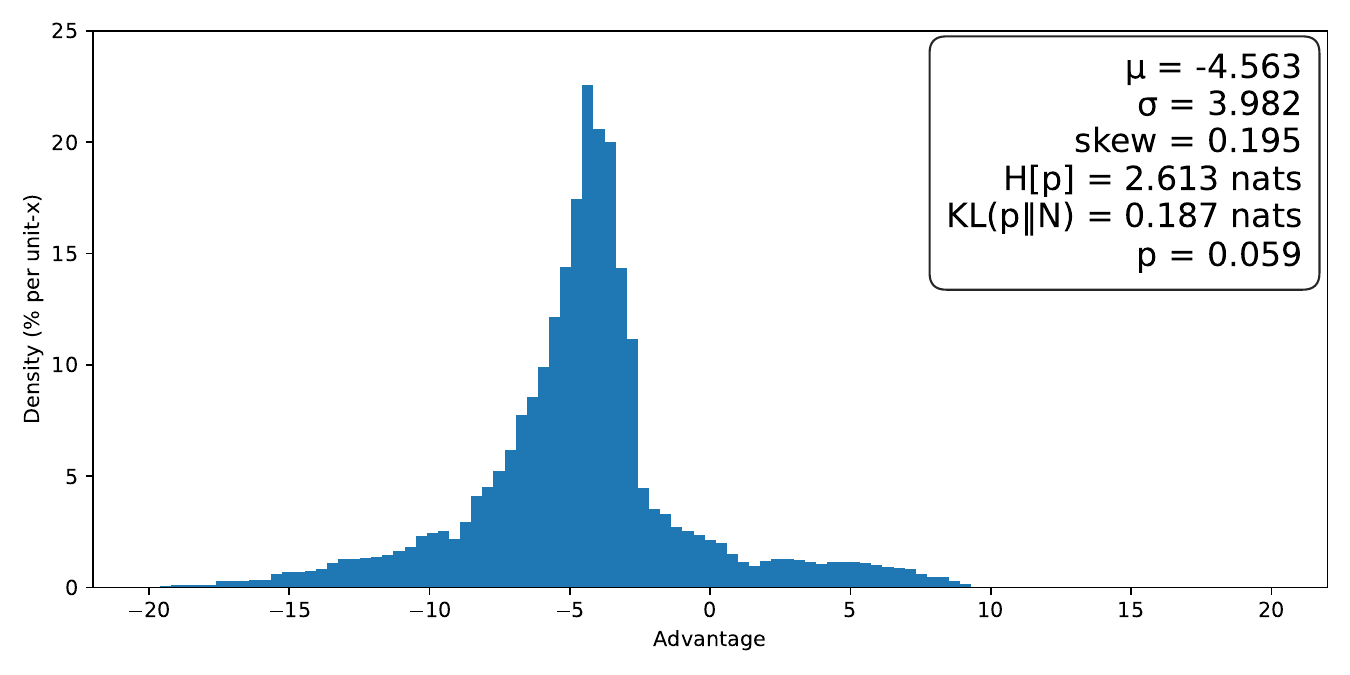}
    \caption{Checkpoint 90}
    \label{fig:rl_adv_hist_90}
  \end{subfigure}                        
  \begin{subfigure}[t]{0.32\linewidth}
    \centering
    \includegraphics[width=\linewidth]{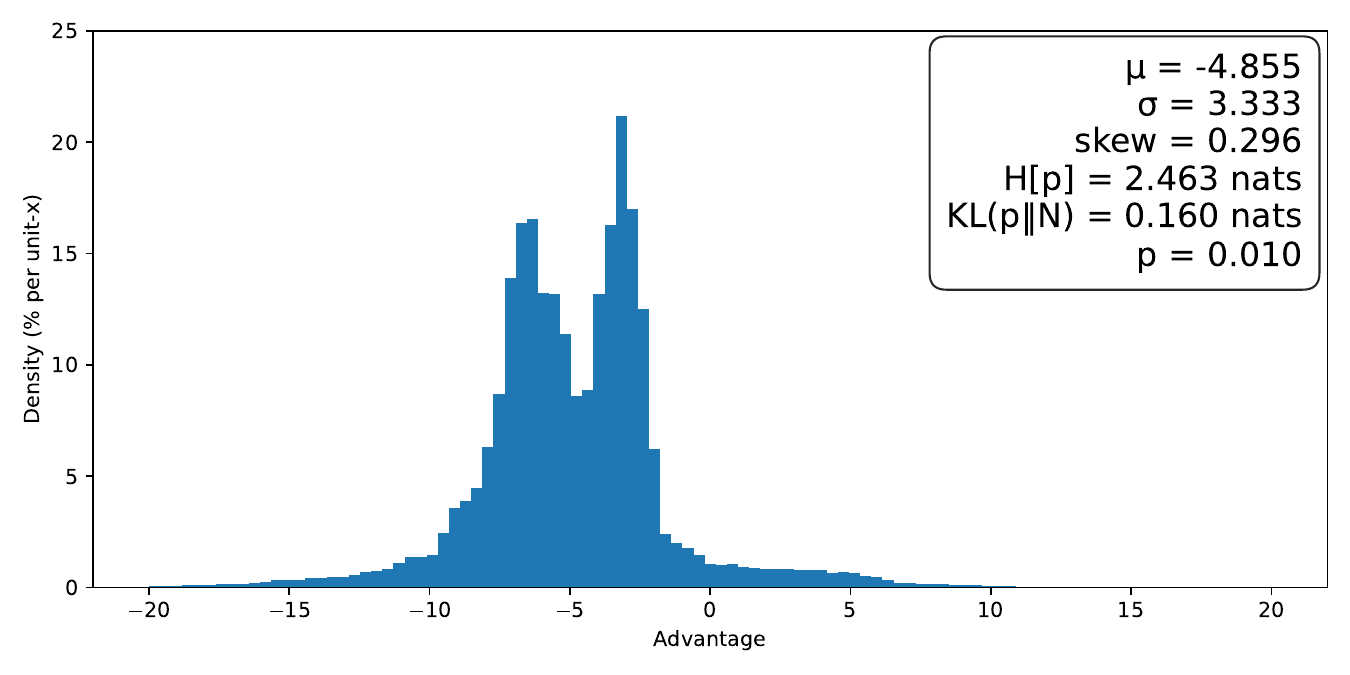}
    \caption{Checkpoint 140}
    \label{fig:rl_adv_hist_140}
  \end{subfigure}
   \begin{subfigure}[t]{0.32\linewidth}
    \centering
    \includegraphics[width=\linewidth]{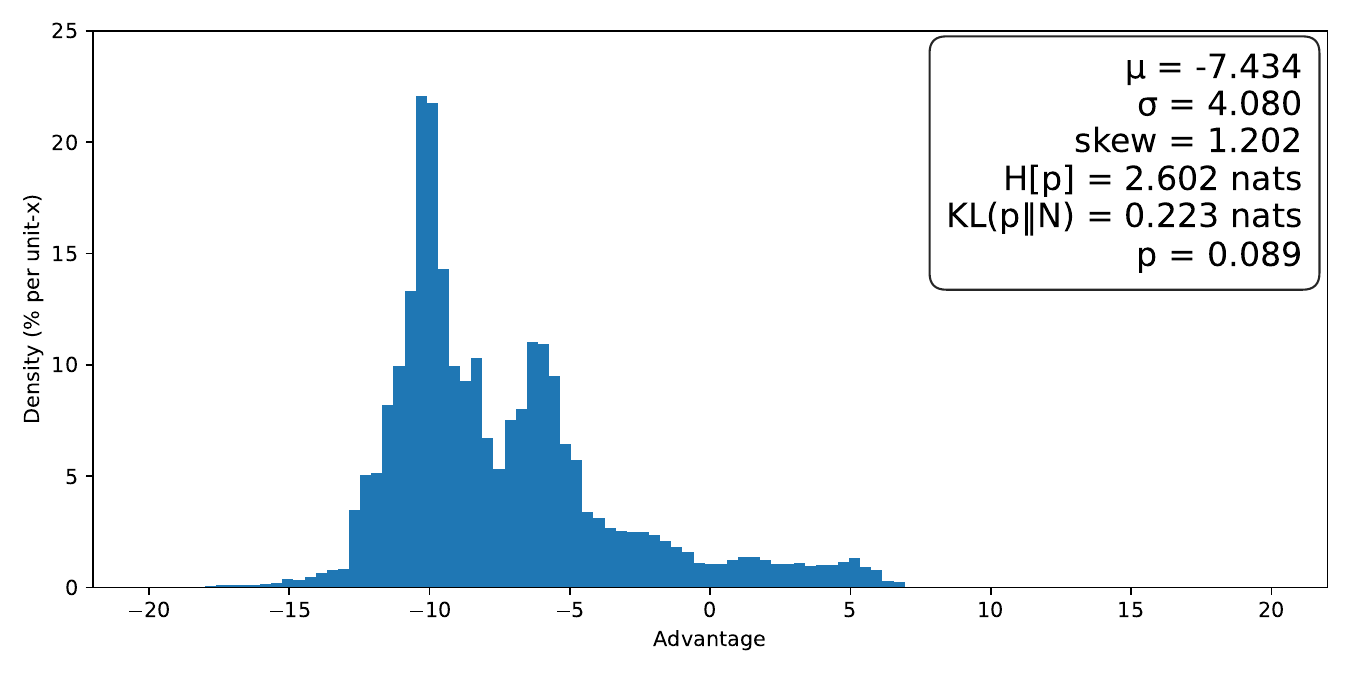}
    \caption{Checkpoint 200}
    \label{fig:rl_adv_hist_200}
  \end{subfigure}                        
  \\
  \begin{subfigure}[t]{0.32\linewidth}
    \centering
    \includegraphics[width=\linewidth]{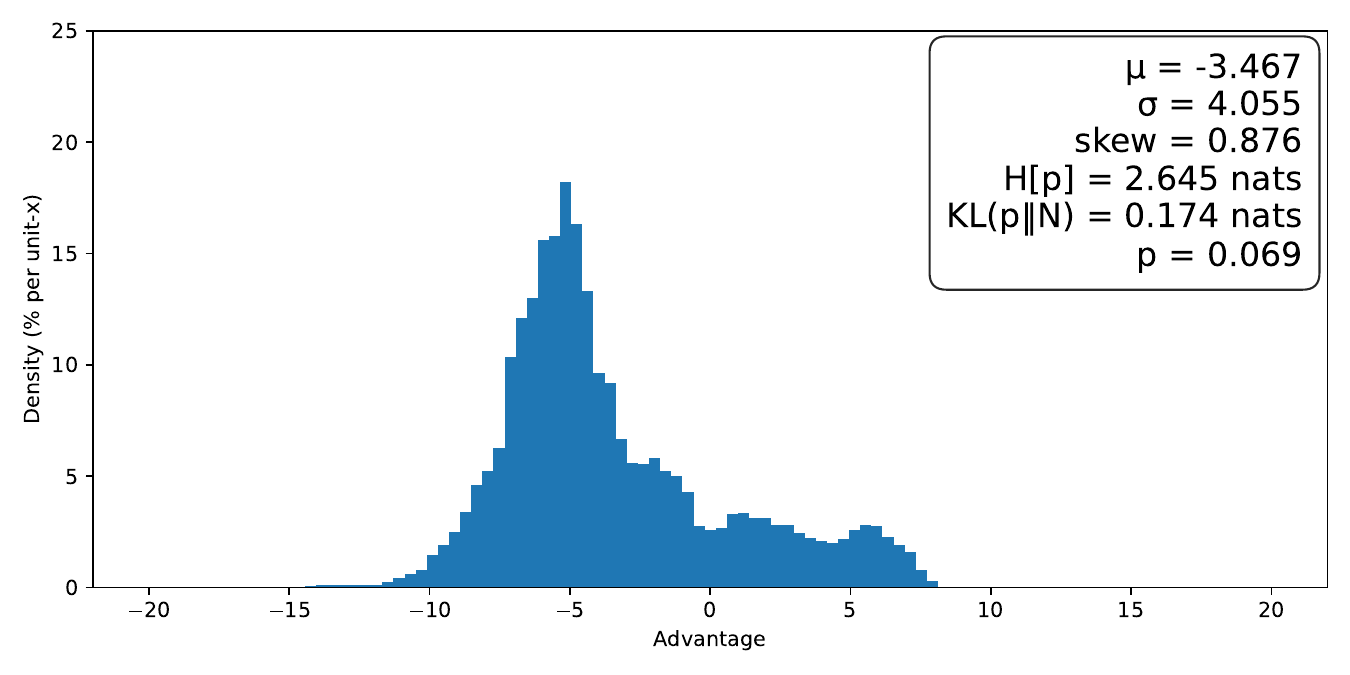}
    \caption{Checkpoint 300}
    \label{fig:rl_adv_hist_300}
  \end{subfigure}                        
  \begin{subfigure}[t]{0.32\linewidth}
    \centering
    \includegraphics[width=\linewidth]{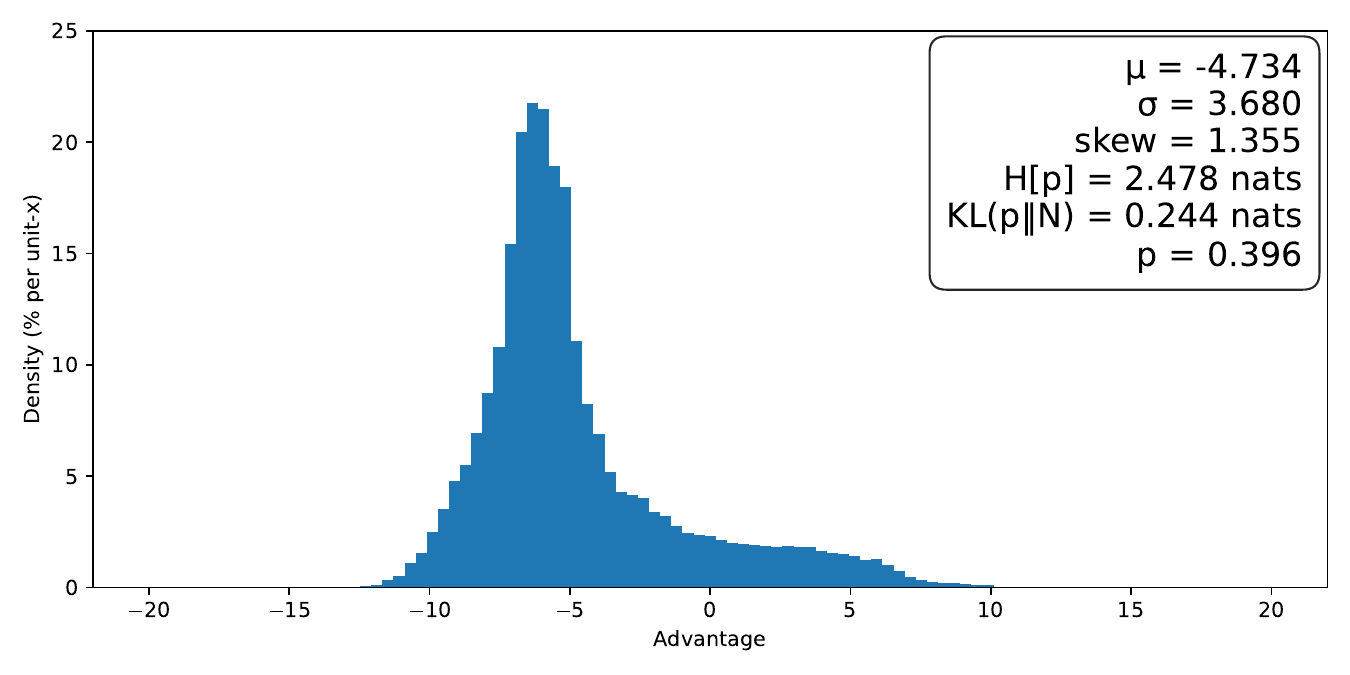}
    \caption{Checkpoint 400}
    \label{fig:rl_adv_hist_400}
  \end{subfigure}
   \begin{subfigure}[t]{0.32\linewidth}
    \centering
    \includegraphics[width=\linewidth]{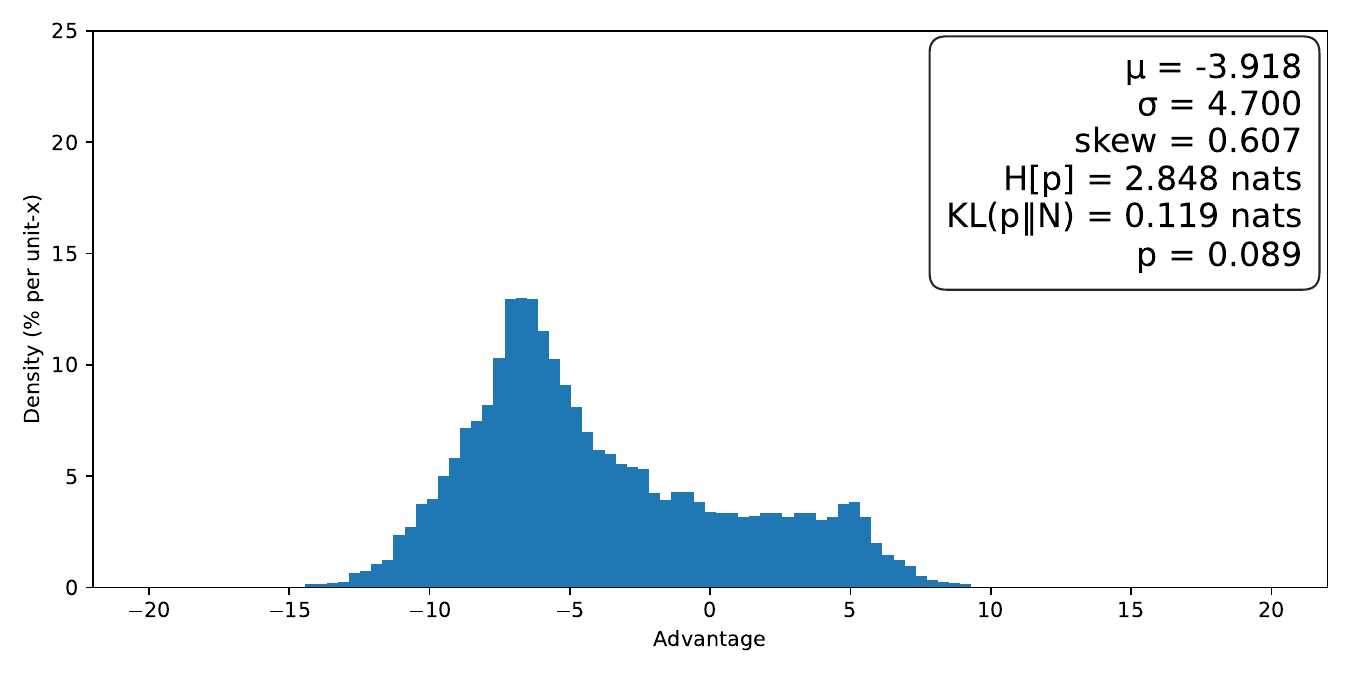}
    \caption{Checkpoint 500}
    \label{fig:rl_adv_hist_500}
  \end{subfigure}                        
  \\
  \begin{subfigure}[t]{0.32\linewidth}
    \centering
    \includegraphics[width=\linewidth]{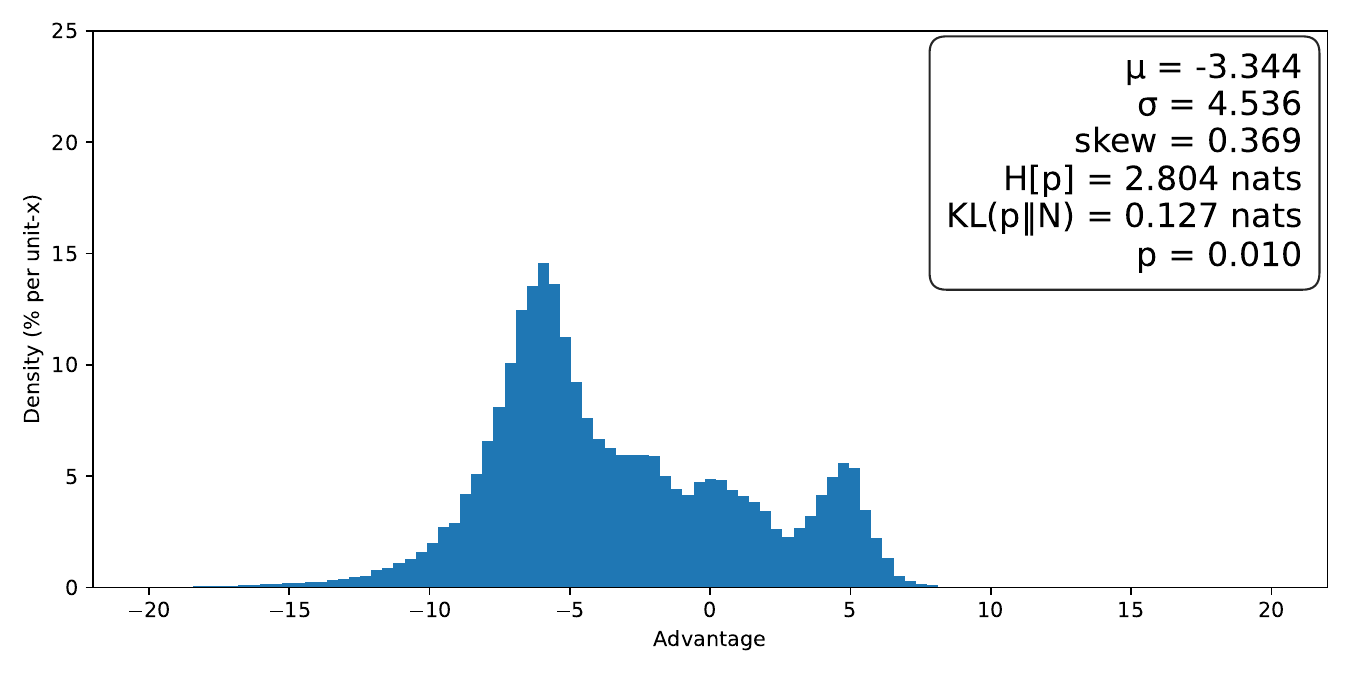}
    \caption{Checkpoint 600}
    \label{fig:rl_adv_hist_600}
  \end{subfigure}                        
  \begin{subfigure}[t]{0.32\linewidth}
    \centering
    \includegraphics[width=\linewidth]{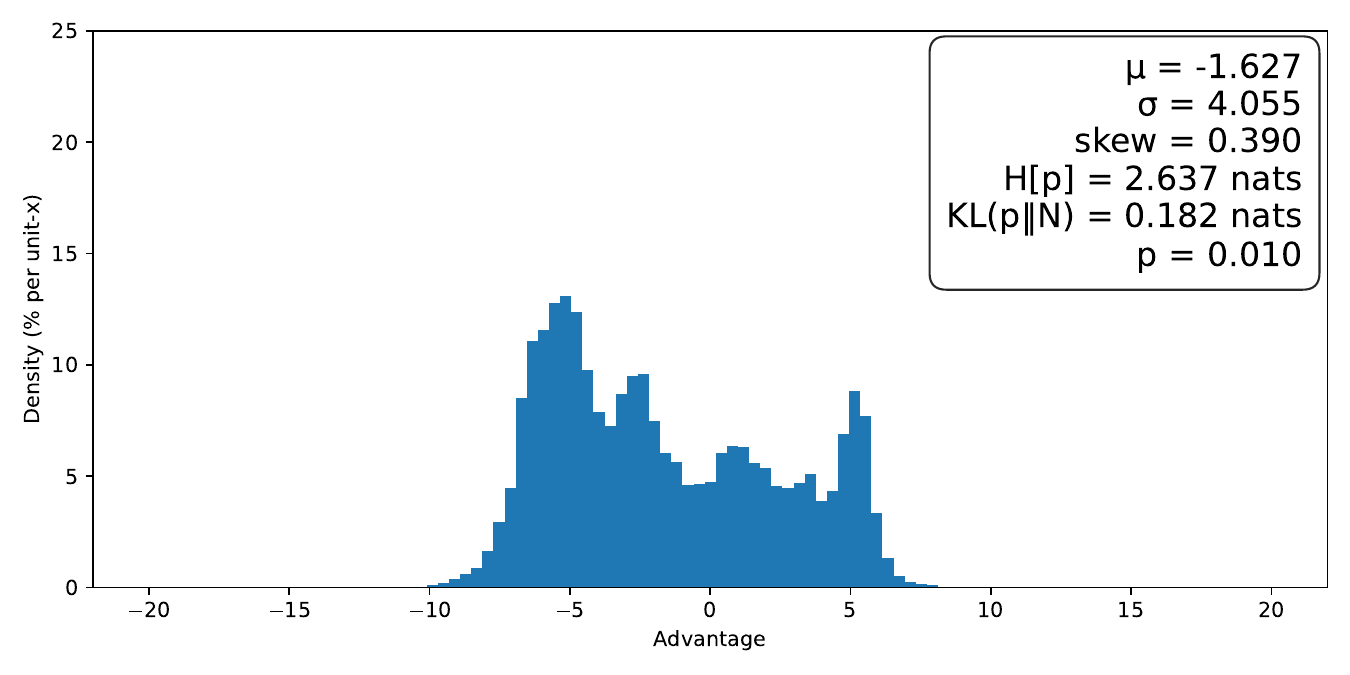}
    \caption{Checkpoint 700}
    \label{fig:rl_adv_hist_700}
  \end{subfigure}
   \begin{subfigure}[t]{0.32\linewidth}
    \centering
    \includegraphics[width=\linewidth]{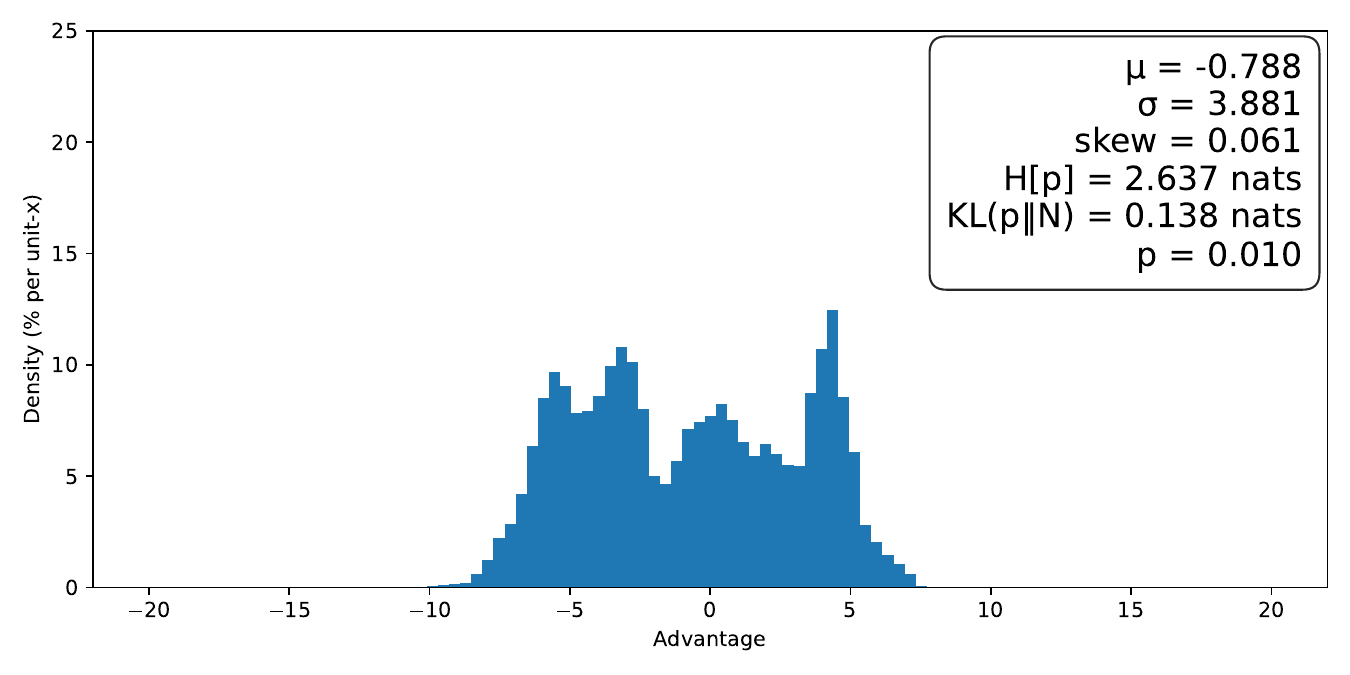}
    \caption{Checkpoint 800}
    \label{fig:rl_adv_hist_800}
  \end{subfigure}                        
  \\
  \begin{subfigure}[t]{0.32\linewidth}
    \centering
    \includegraphics[width=\linewidth]{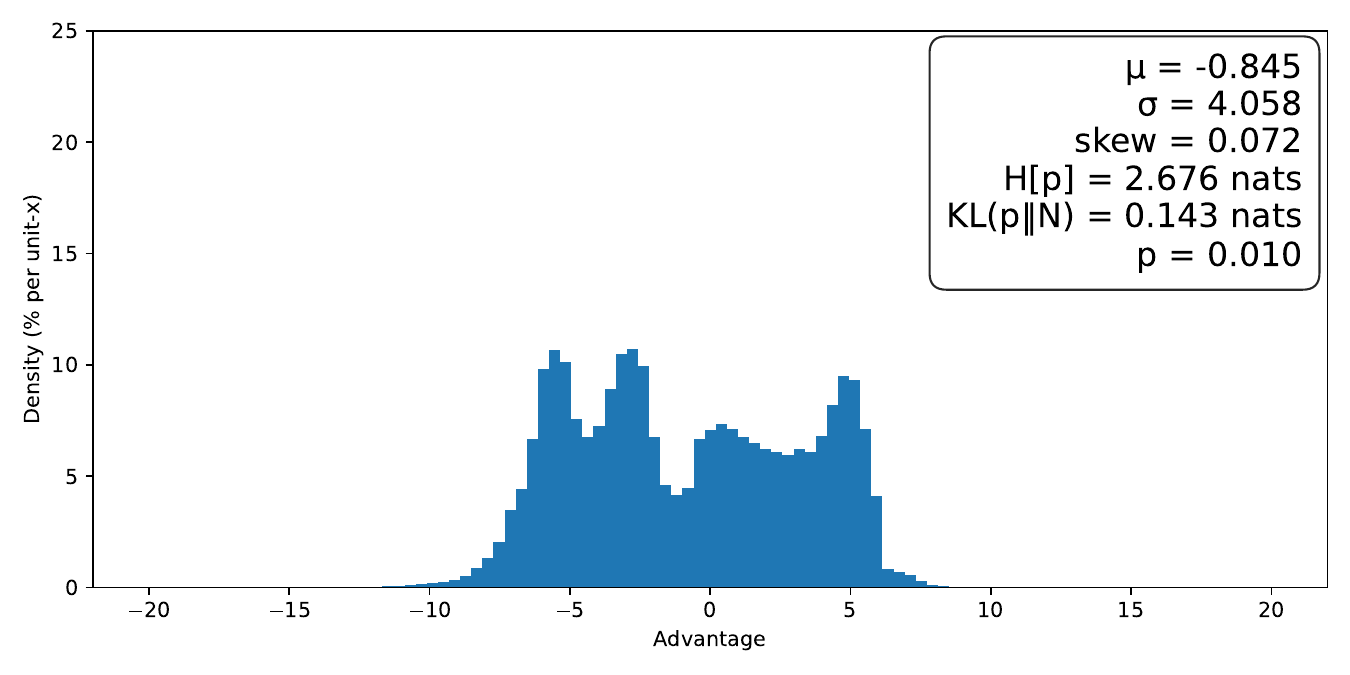}
    \caption{Checkpoint 900}
    \label{fig:rl_adv_hist_900}
  \end{subfigure}                        
  \begin{subfigure}[t]{0.32\linewidth}
    \centering
    \includegraphics[width=\linewidth]{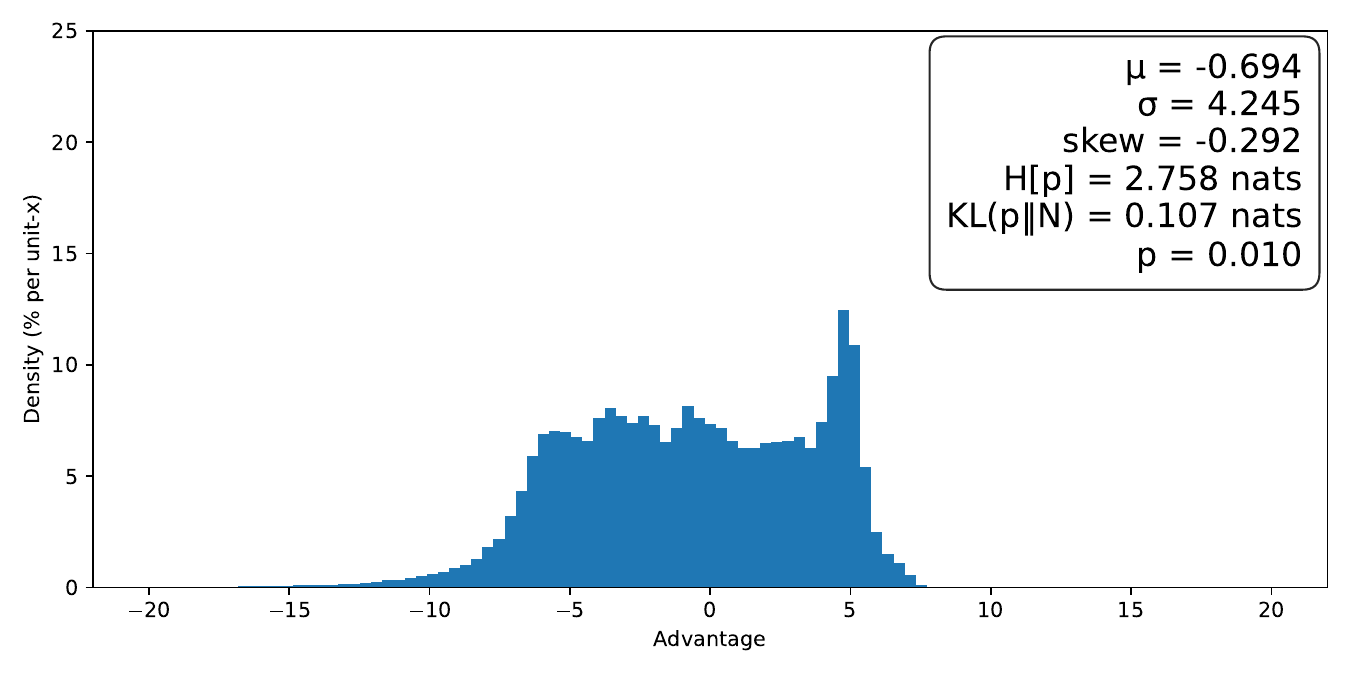}
    \caption{Checkpoint 1000}
    \label{fig:rl_adv_hist_1000}
  \end{subfigure}
   \begin{subfigure}[t]{0.32\linewidth}
    \centering
    \includegraphics[width=\linewidth]{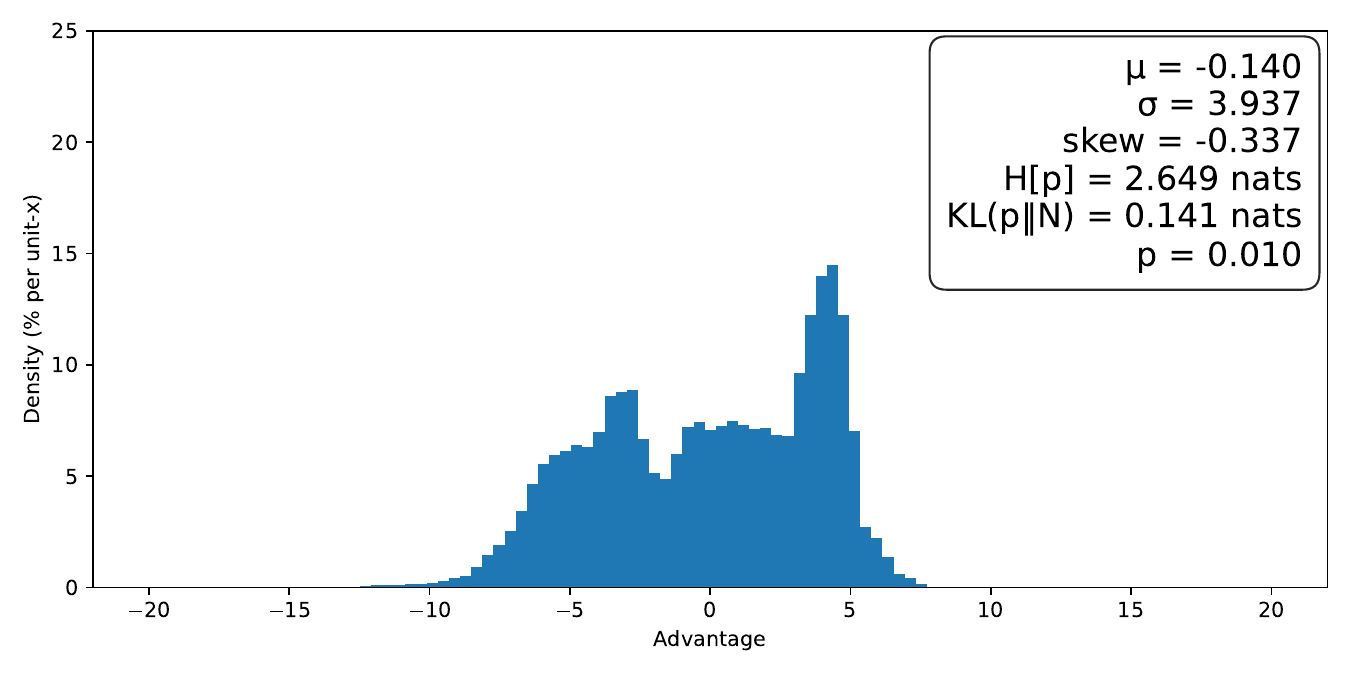}
    \caption{Checkpoint 1100}
    \label{fig:rl_adv_hist_1100}
  \end{subfigure}                        
  \\
  \begin{subfigure}[t]{0.32\linewidth}
    \centering
    \includegraphics[width=\linewidth]{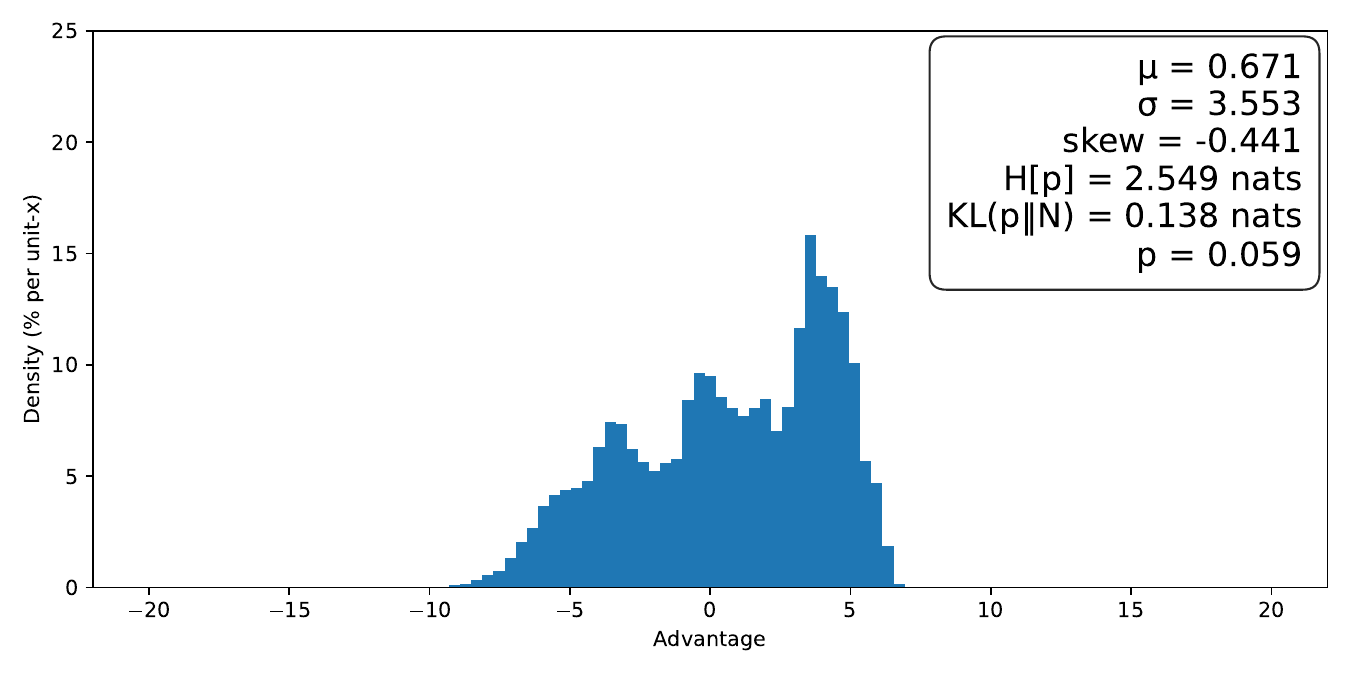}
    \caption{Checkpoint 1200}
    \label{fig:rl_adv_hist_1200}
  \end{subfigure}                        
  \begin{subfigure}[t]{0.32\linewidth}
    \centering
    \includegraphics[width=\linewidth]{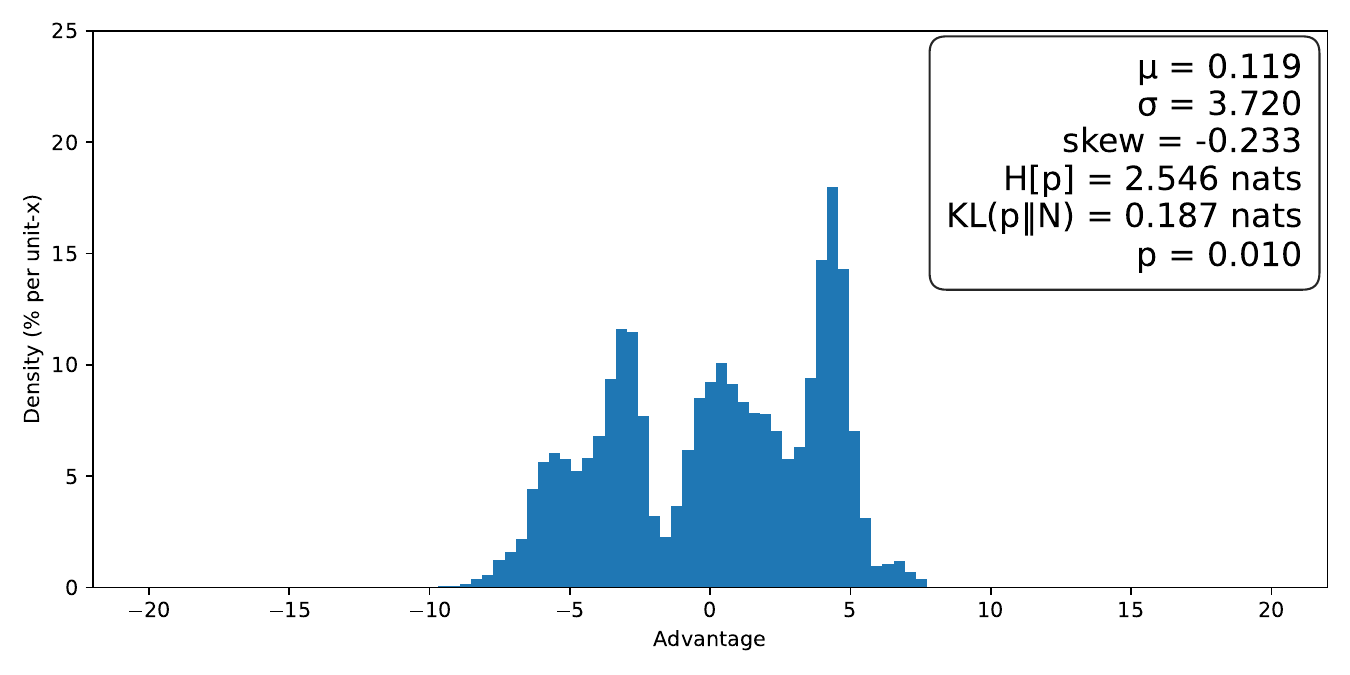}
    \caption{Checkpoint 1300}
    \label{fig:rl_adv_hist_1300}
  \end{subfigure}
   \begin{subfigure}[t]{0.32\linewidth}
    \centering
    \includegraphics[width=\linewidth]{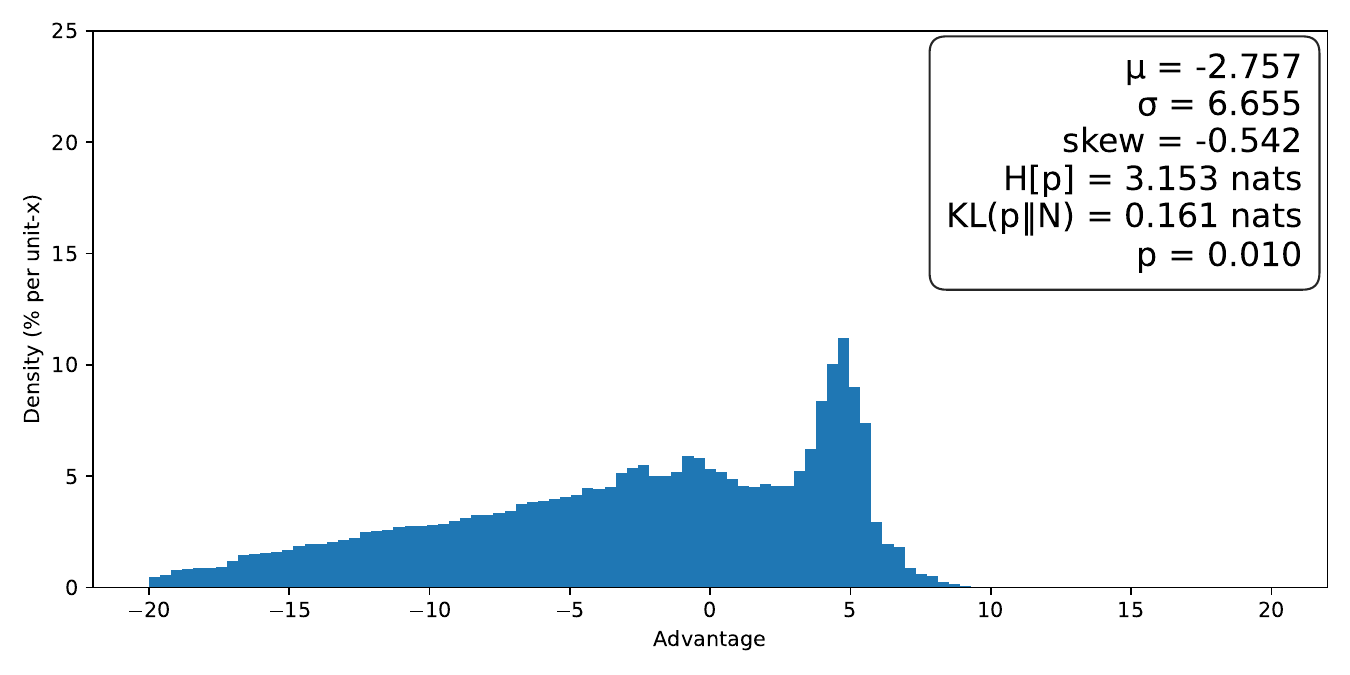}
    \caption{Checkpoint 1400}
    \label{fig:rl_adv_hist_1400}
  \end{subfigure}                        
  \\
   \begin{subfigure}[t]{0.32\linewidth}
    \centering
    \includegraphics[width=\linewidth]{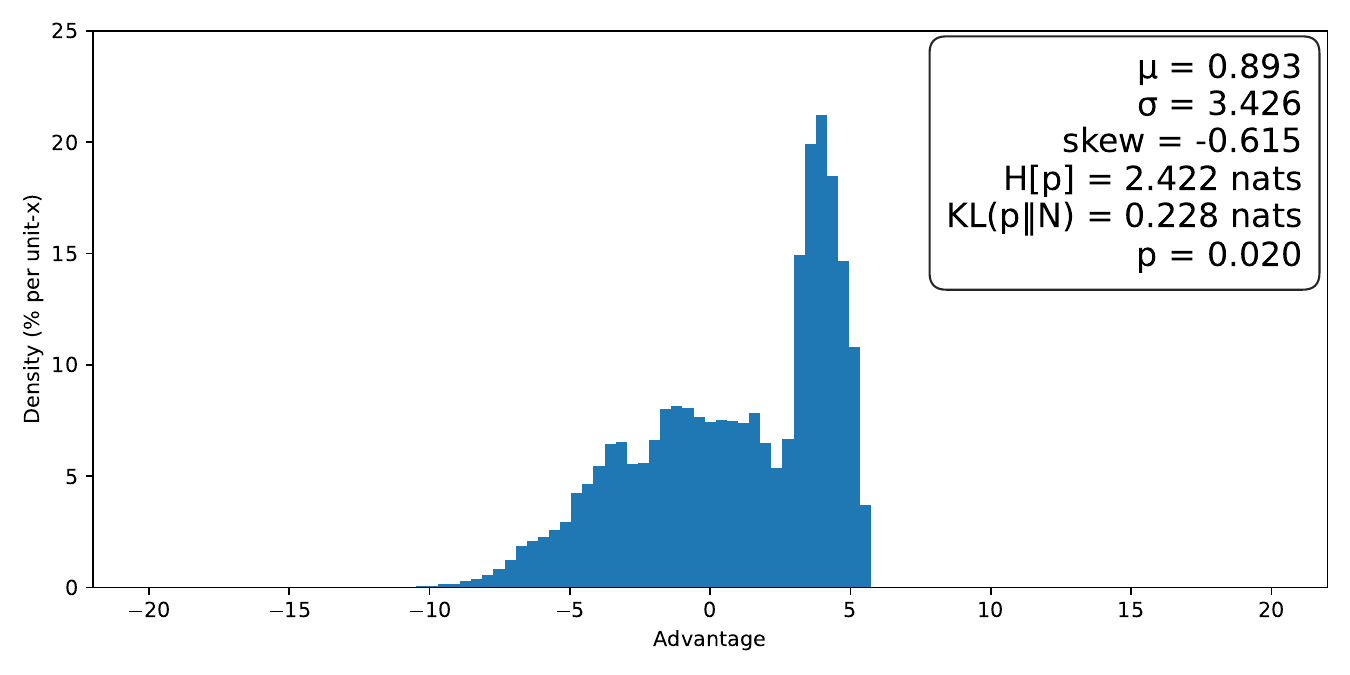}
    \caption{Checkpoint 1500}
    \label{fig:rl_adv_hist_1500}
  \end{subfigure}                        
   \begin{subfigure}[t]{0.32\linewidth}
    \centering
    \includegraphics[width=\linewidth]{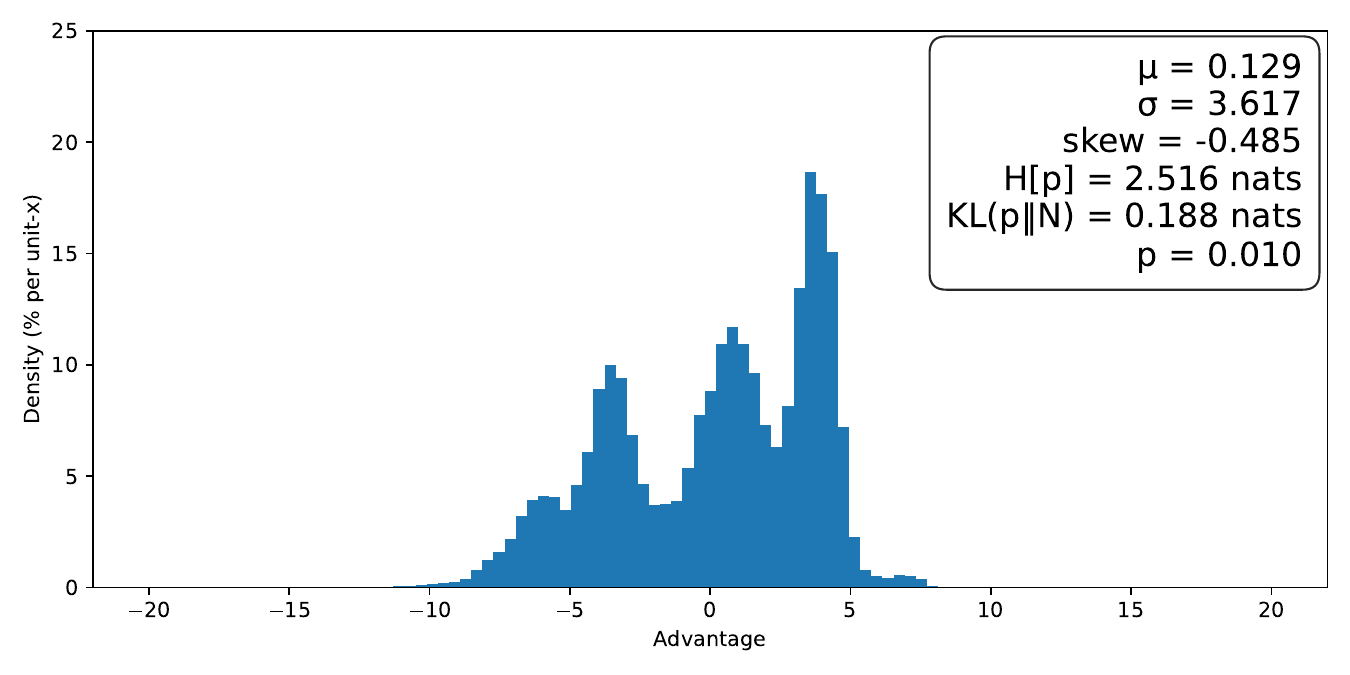}
    \caption{Checkpoint 1600}
    \label{fig:rl_adv_hist_1600}
  \end{subfigure}                        
\end{figure}

\begin{figure}[ht]
  \begin{subfigure}[t]{0.32\linewidth}
    \centering
    \includegraphics[width=\linewidth]{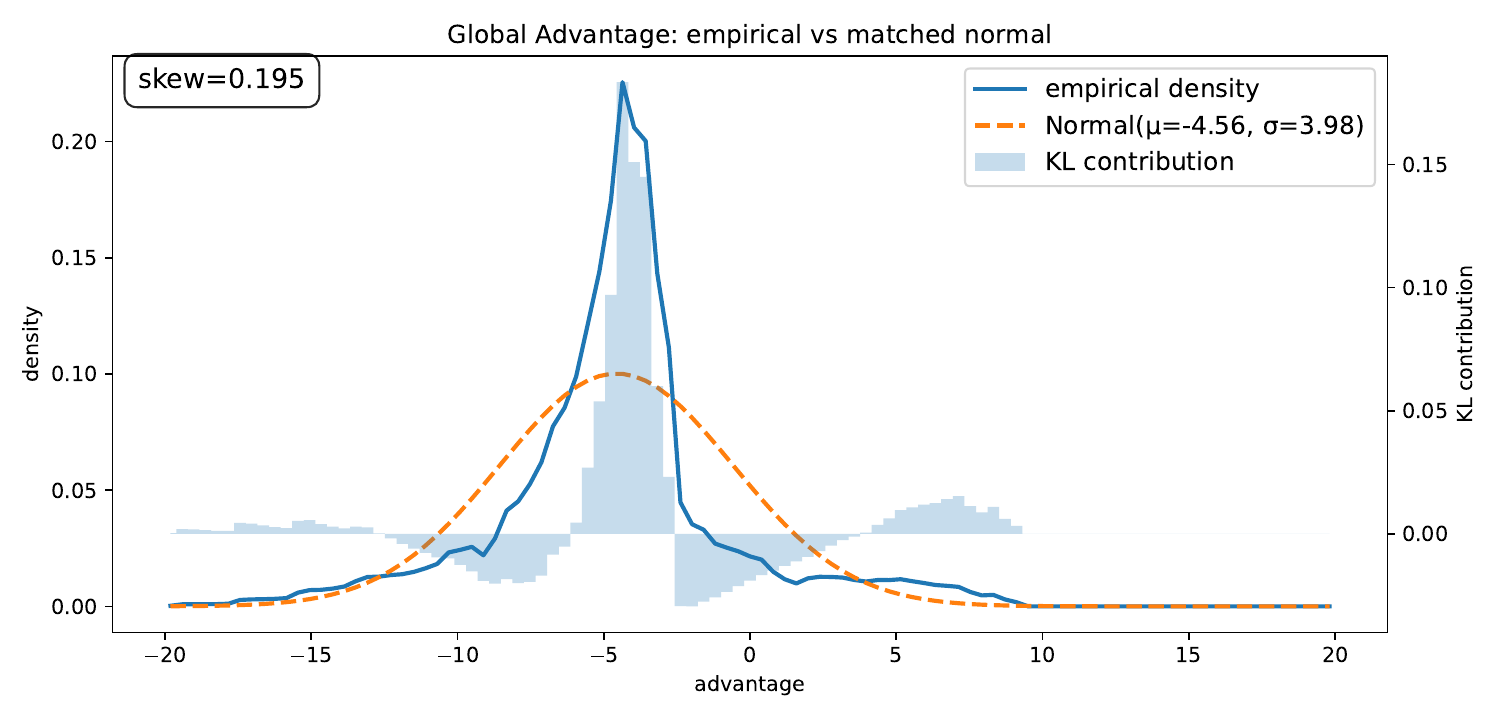}
    \caption{Checkpoint 90}
    \label{fig:rl_adv_density_comparison_90}
  \end{subfigure}                        
  \begin{subfigure}[t]{0.32\linewidth}
    \centering
    \includegraphics[width=\linewidth]{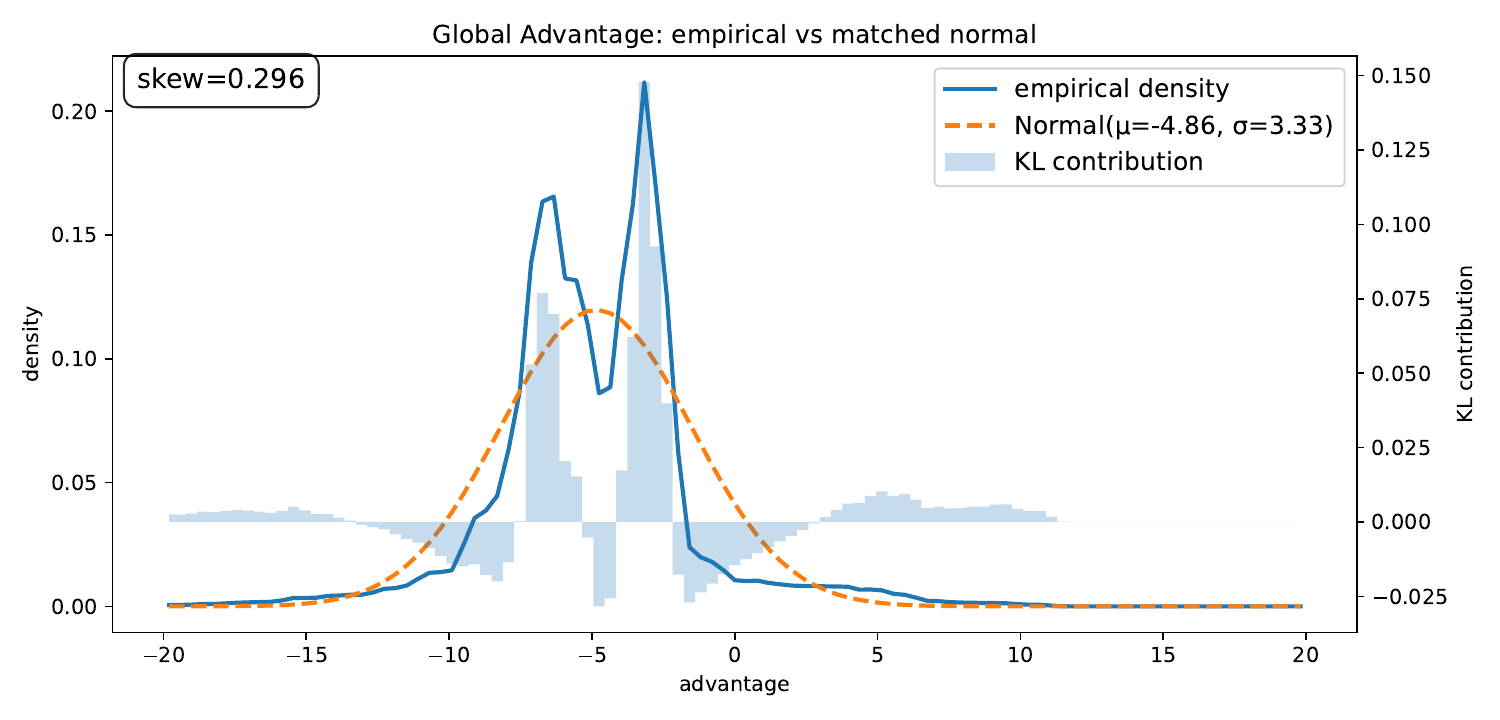}
    \caption{Checkpoint 140}
    \label{fig:rl_adv_density_comparison_140}
  \end{subfigure}
   \begin{subfigure}[t]{0.32\linewidth}
    \centering
    \includegraphics[width=\linewidth]{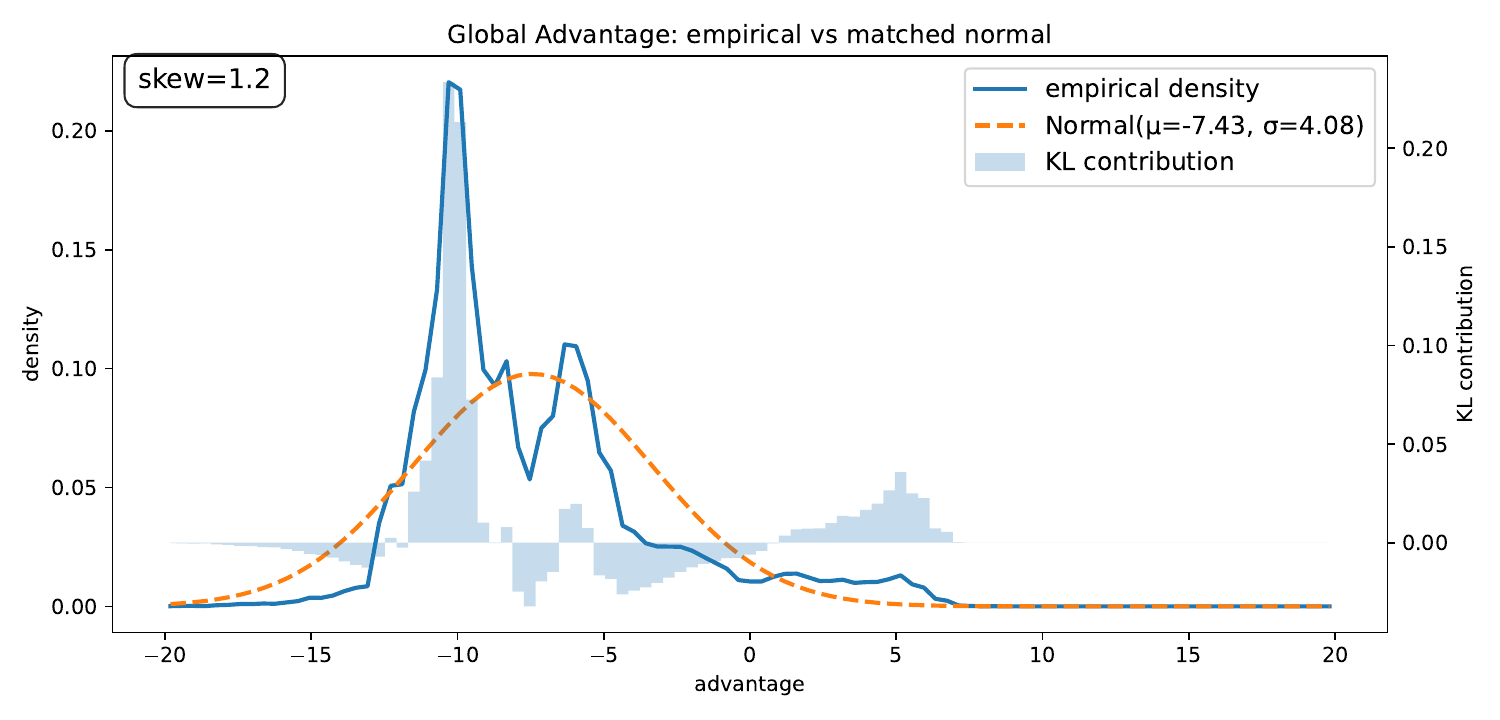}
    \caption{Checkpoint 200}
    \label{fig:rl_adv_density_comparison_200}
  \end{subfigure}                        
  \\
  \begin{subfigure}[t]{0.32\linewidth}
    \centering
    \includegraphics[width=\linewidth]{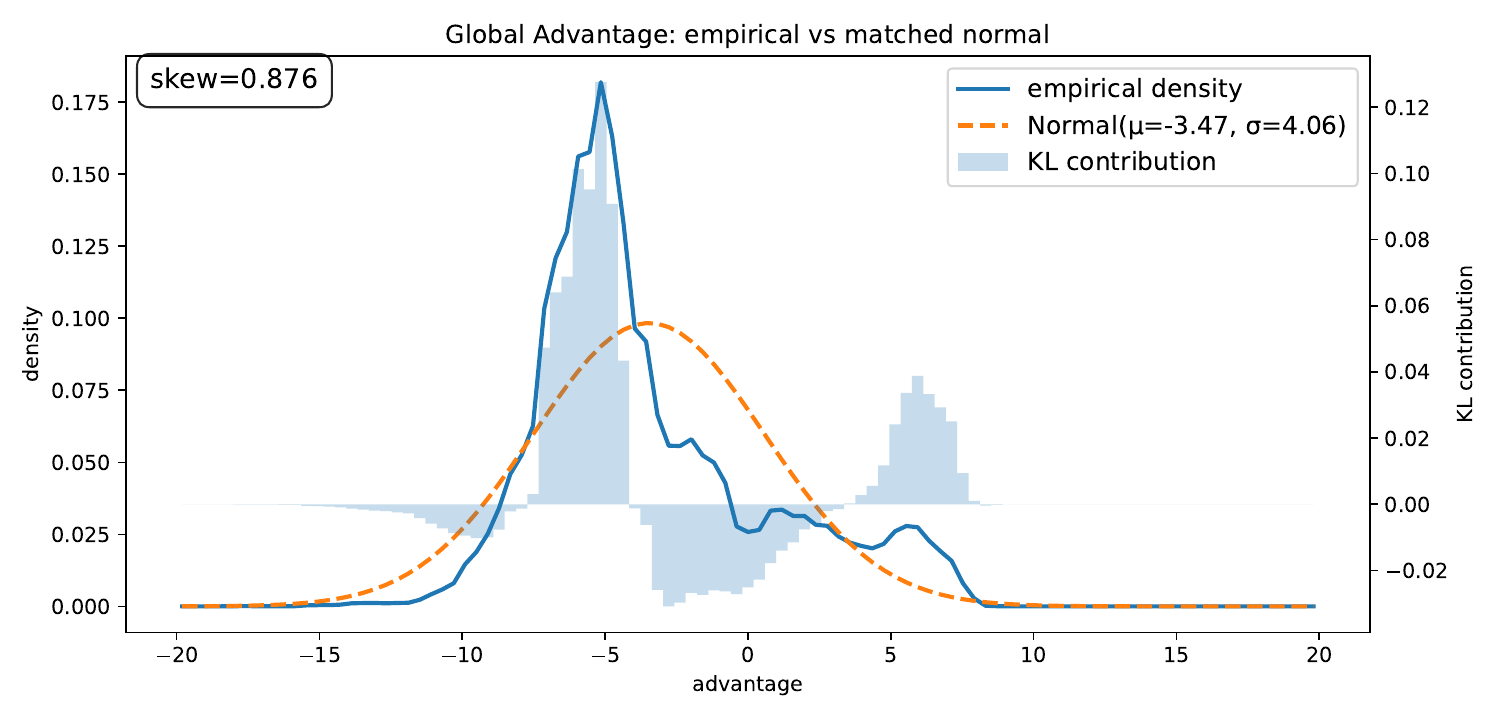}
    \caption{Checkpoint 300}
    \label{fig:rl_adv_density_comparison_300}
  \end{subfigure}                        
  \begin{subfigure}[t]{0.32\linewidth}
    \centering
    \includegraphics[width=\linewidth]{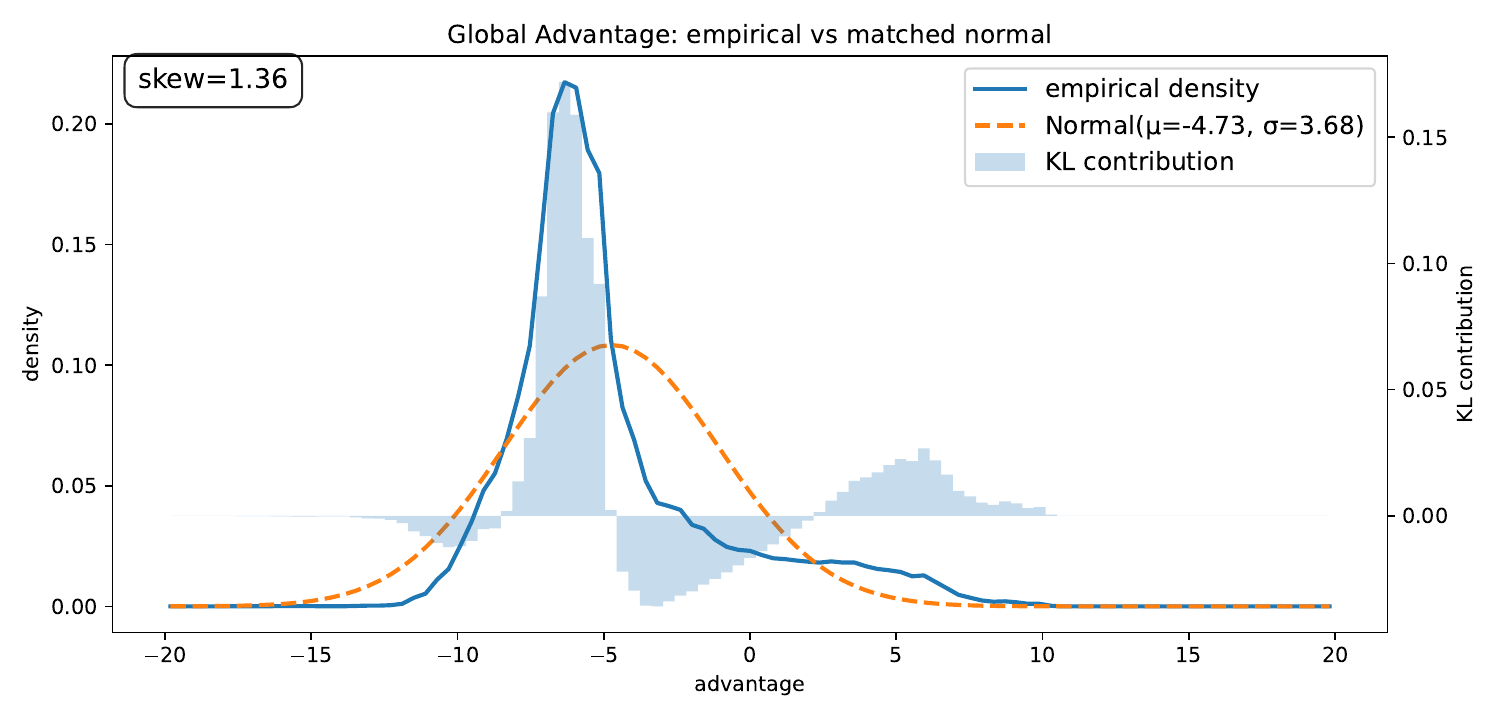}
    \caption{Checkpoint 400}
    \label{fig:rl_adv_density_comparison_400}
  \end{subfigure}
   \begin{subfigure}[t]{0.32\linewidth}
    \centering
    \includegraphics[width=\linewidth]{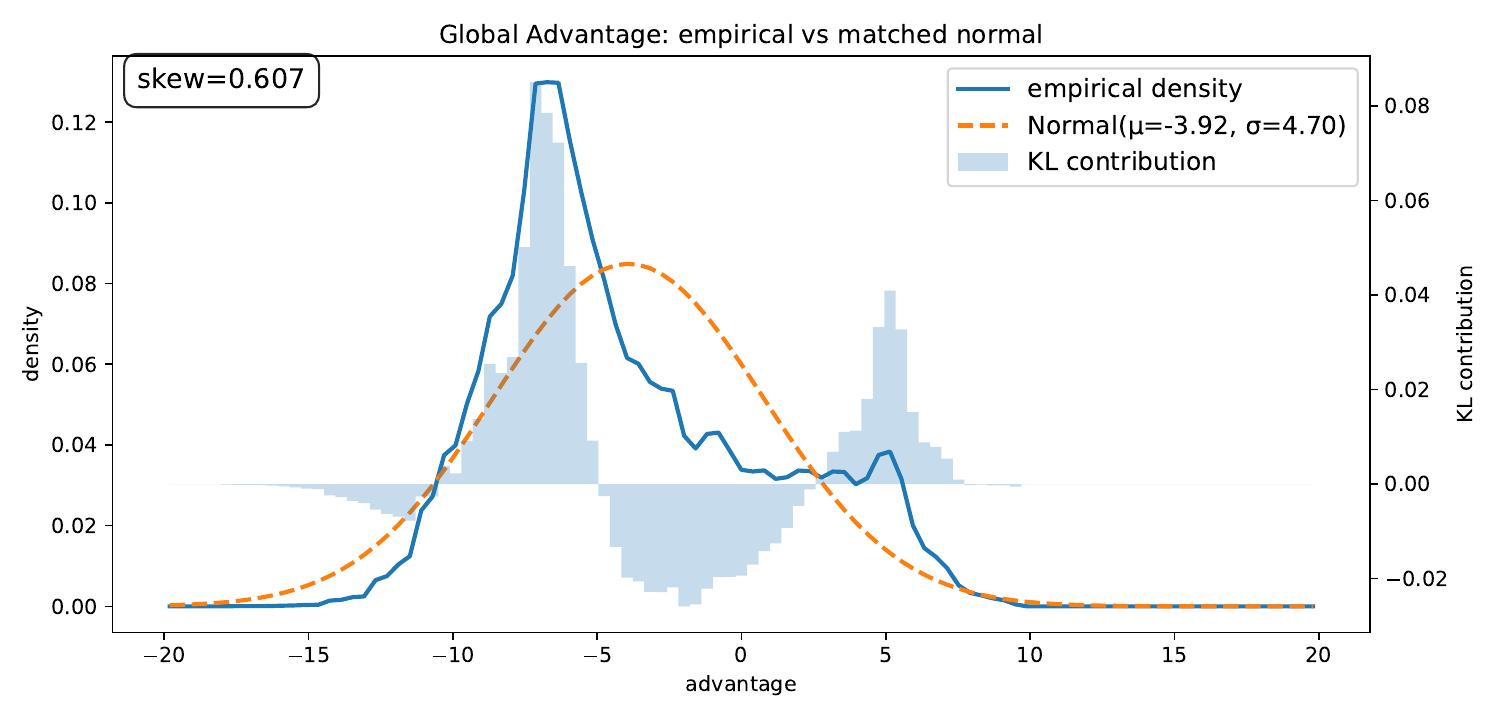}
    \caption{Checkpoint 500}
    \label{fig:rl_adv_density_comparison_500}
  \end{subfigure}                        
  \\
  \begin{subfigure}[t]{0.32\linewidth}
    \centering
    \includegraphics[width=\linewidth]{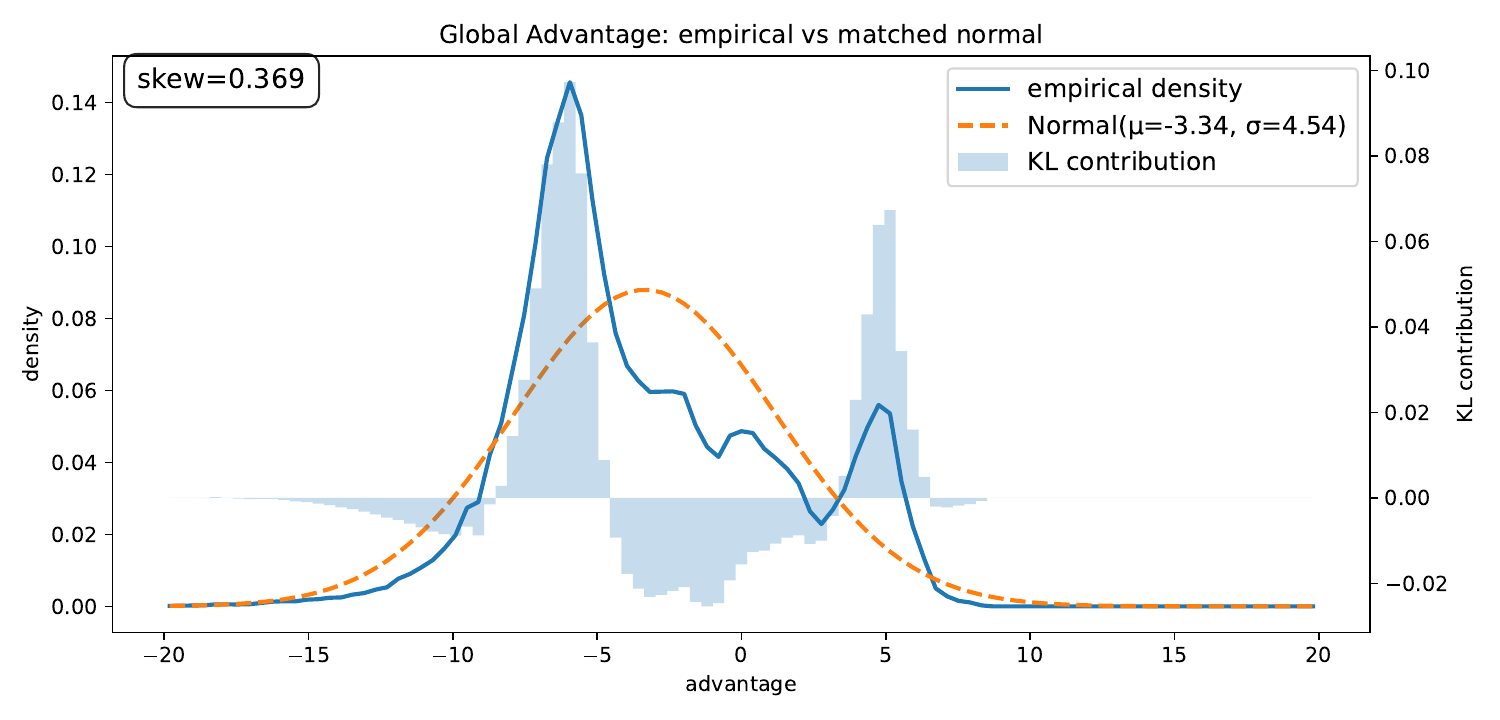}
    \caption{Checkpoint 600}
    \label{fig:rl_adv_density_comparison_600}
  \end{subfigure}                        
  \begin{subfigure}[t]{0.32\linewidth}
    \centering
    \includegraphics[width=\linewidth]{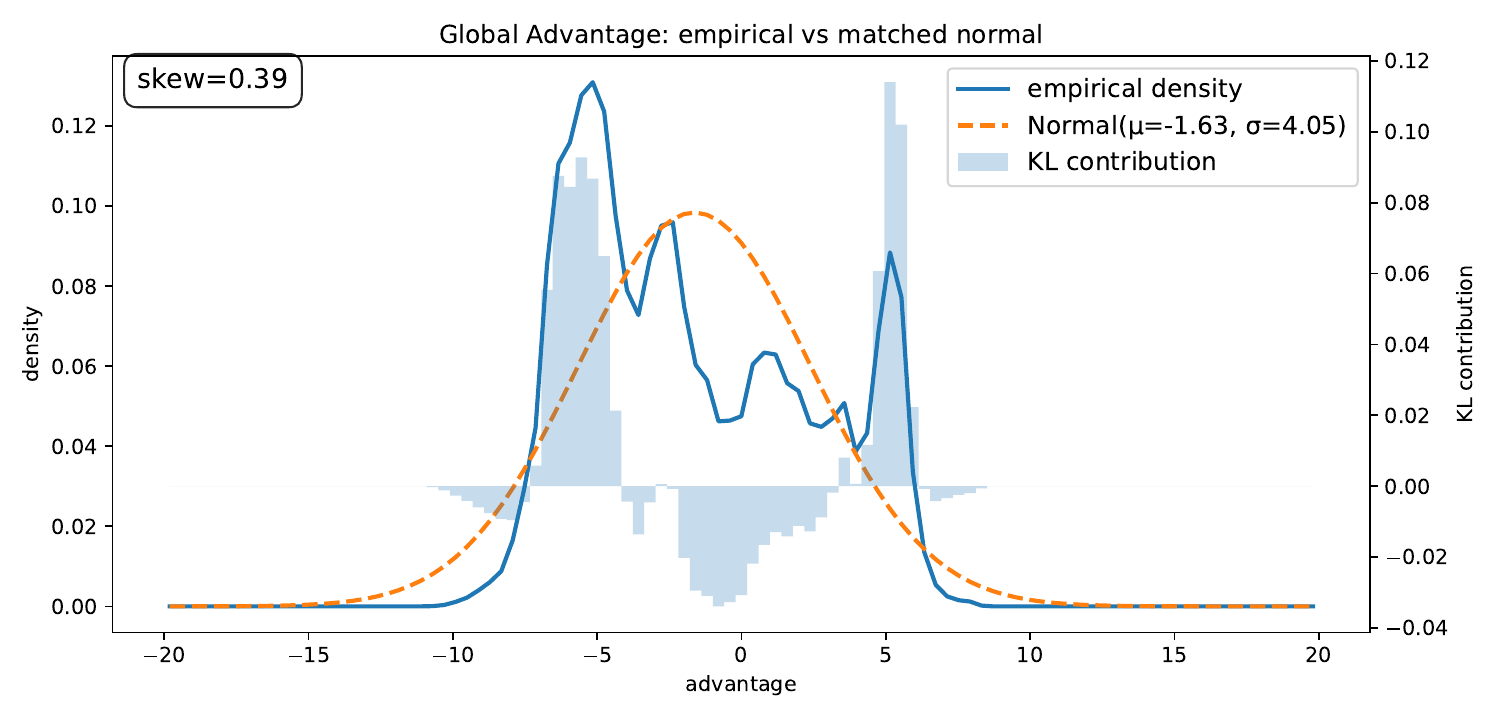}
    \caption{Checkpoint 700}
    \label{fig:rl_adv_density_comparison_700}
  \end{subfigure}
   \begin{subfigure}[t]{0.32\linewidth}
    \centering
    \includegraphics[width=\linewidth]{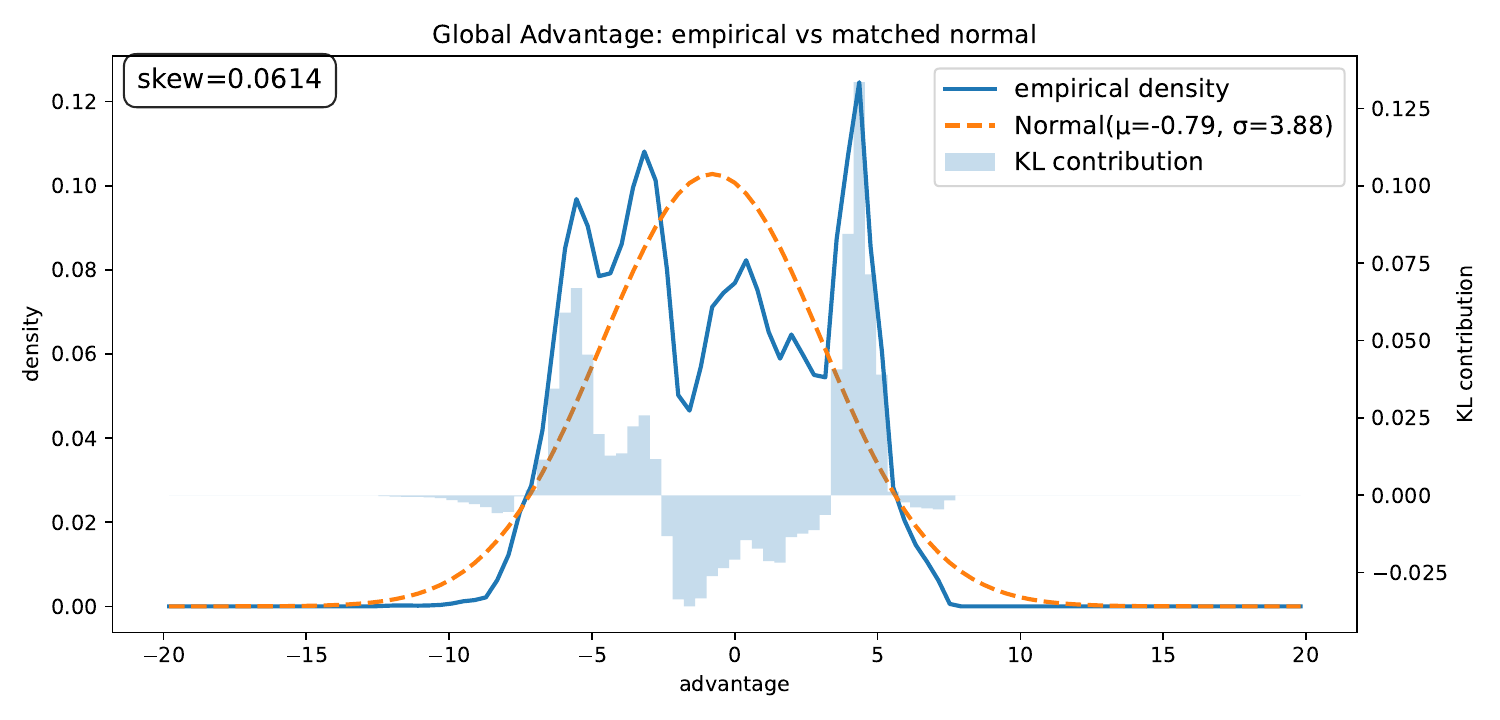}
    \caption{Checkpoint 800}
    \label{fig:rl_adv_density_comparison_800}
  \end{subfigure}                        
  \\
  \begin{subfigure}[t]{0.32\linewidth}
    \centering
    \includegraphics[width=\linewidth]{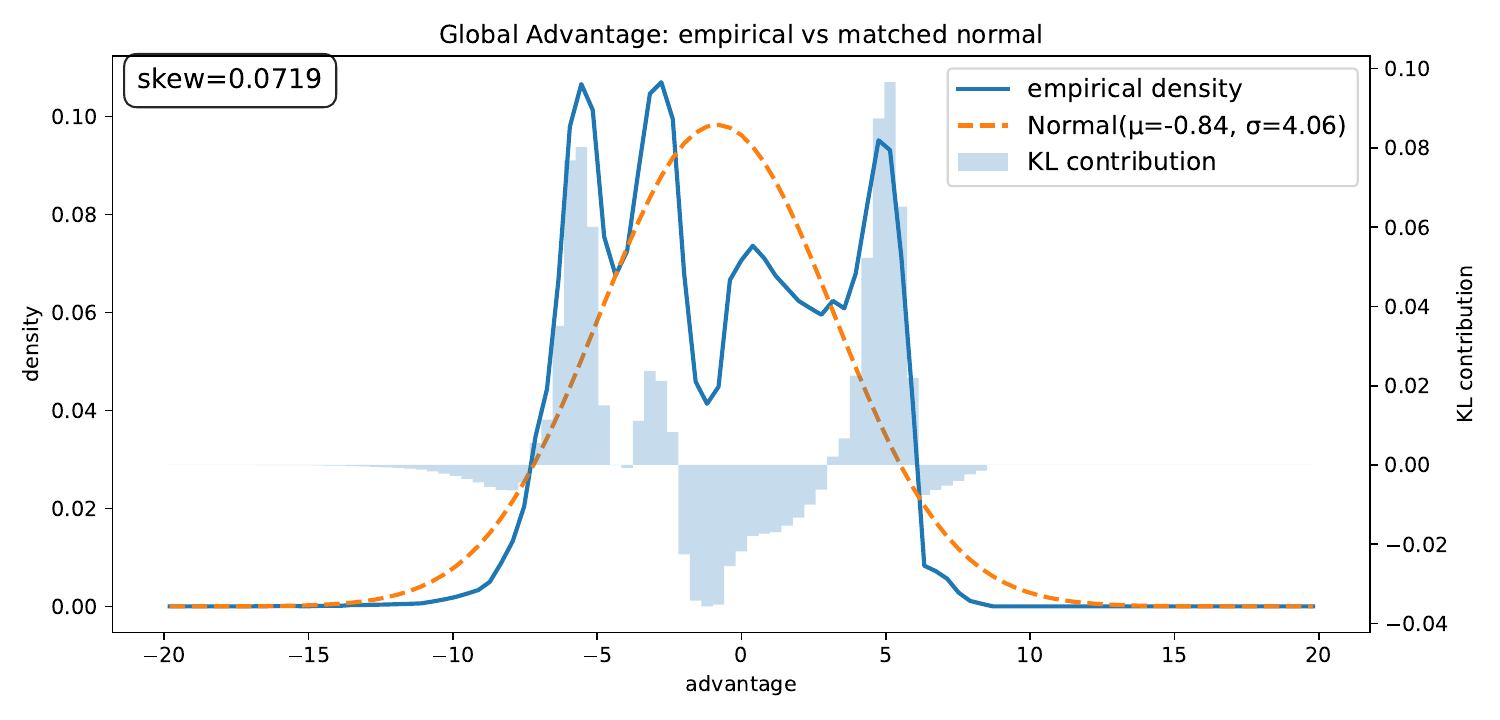}
    \caption{Checkpoint 900}
    \label{fig:rl_adv_density_comparison_900}
  \end{subfigure}                        
  \begin{subfigure}[t]{0.32\linewidth}
    \centering
    \includegraphics[width=\linewidth]{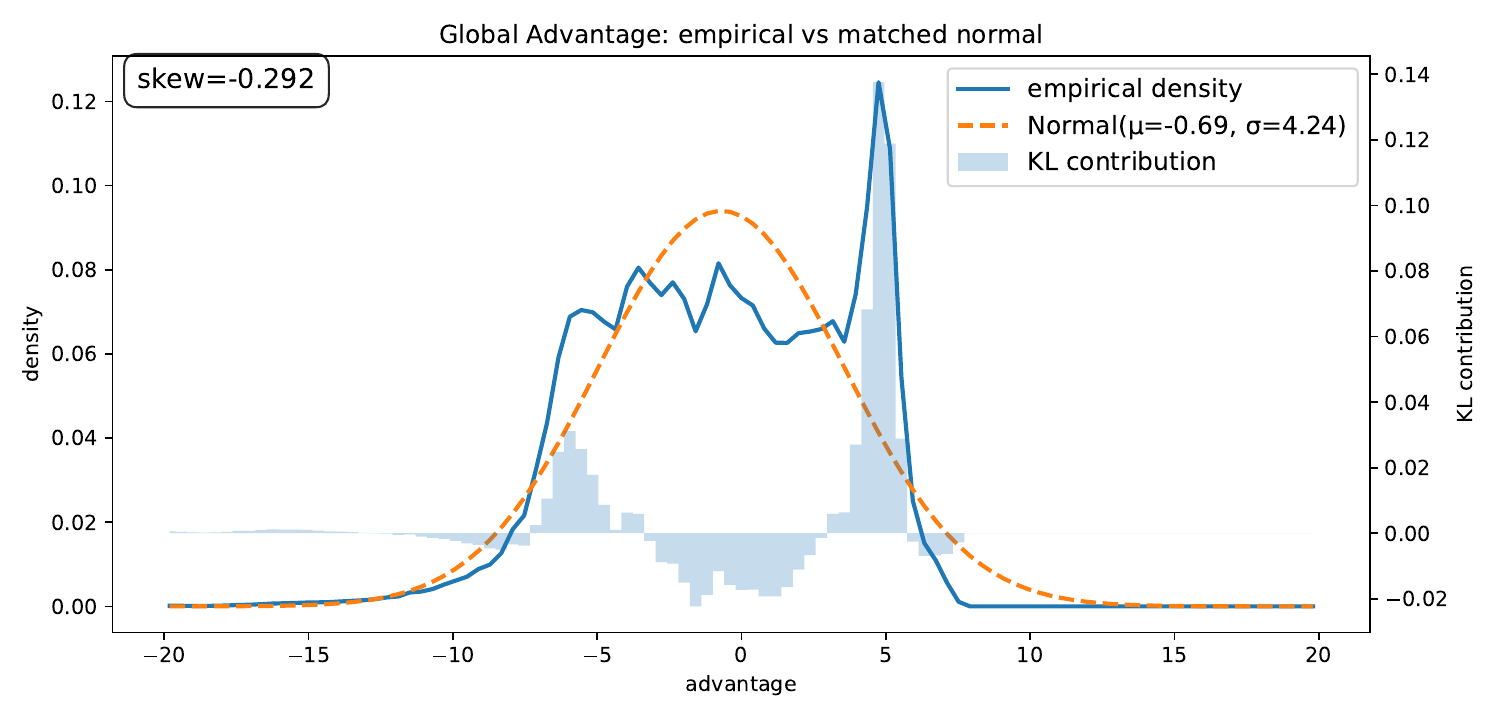}
    \caption{Checkpoint 1000}
    \label{fig:rl_adv_density_comparison_1000}
  \end{subfigure}
   \begin{subfigure}[t]{0.32\linewidth}
    \centering
    \includegraphics[width=\linewidth]{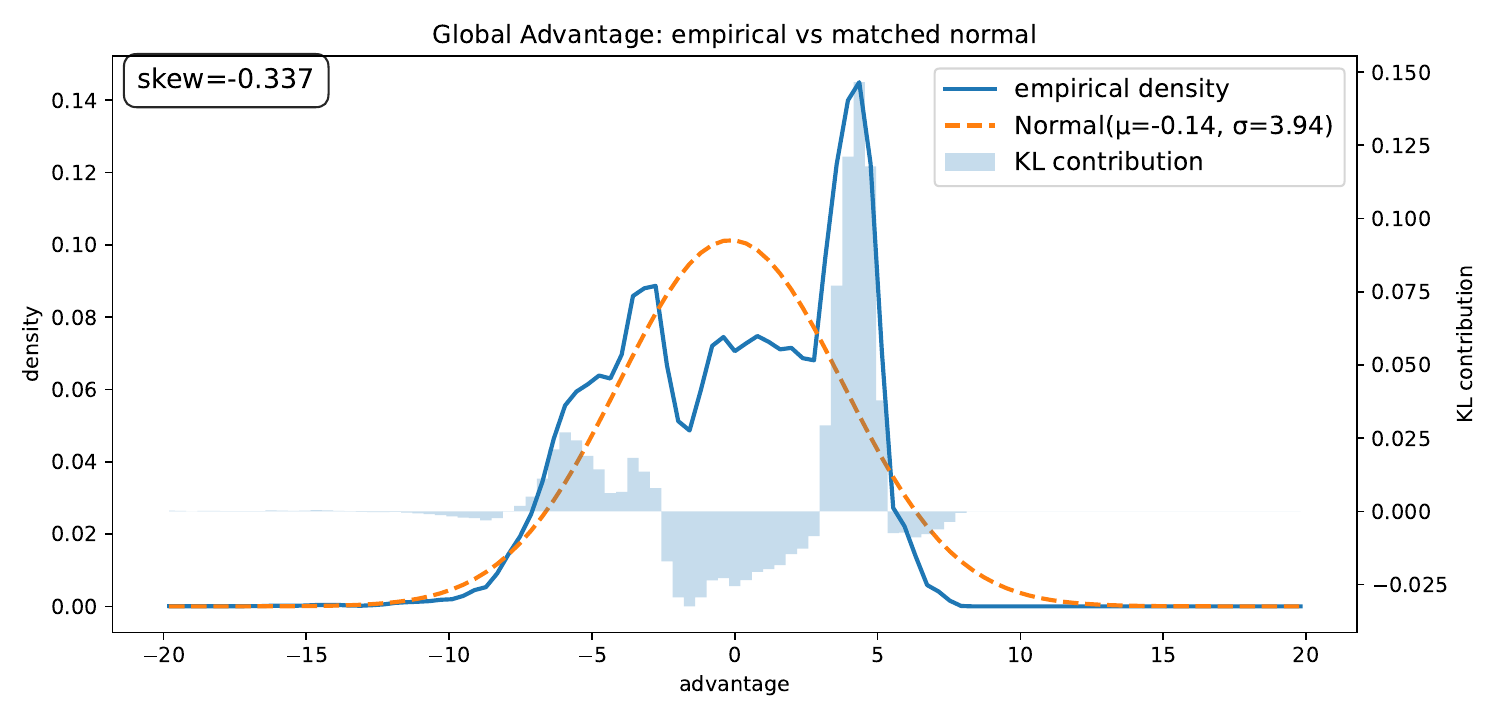}
    \caption{Checkpoint 1100}
    \label{fig:rl_adv_density_comparison_1100}
  \end{subfigure}                        
  \\
  \begin{subfigure}[t]{0.32\linewidth}
    \centering
    \includegraphics[width=\linewidth]{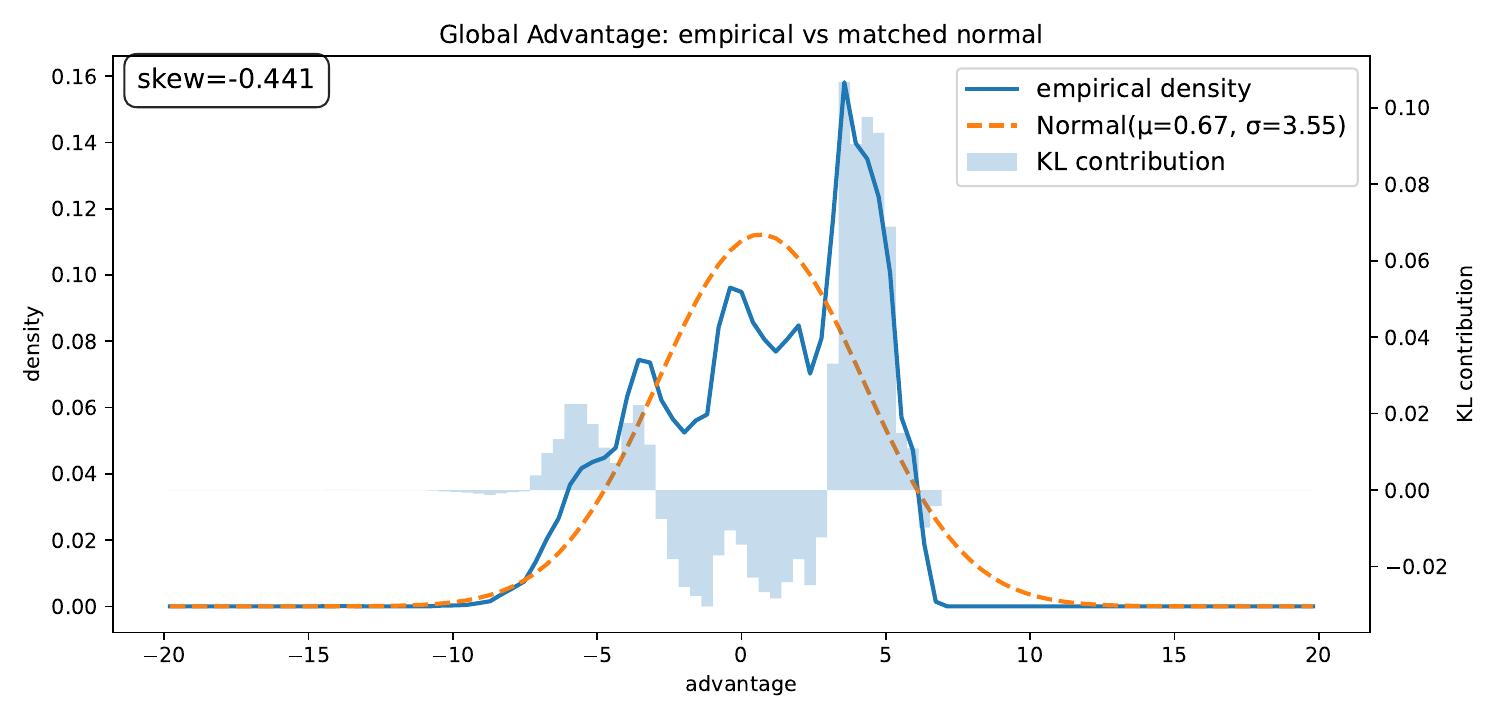}
    \caption{Checkpoint 1200}
    \label{fig:rl_adv_density_comparison_1200}
  \end{subfigure}                        
  \begin{subfigure}[t]{0.32\linewidth}
    \centering
    \includegraphics[width=\linewidth]{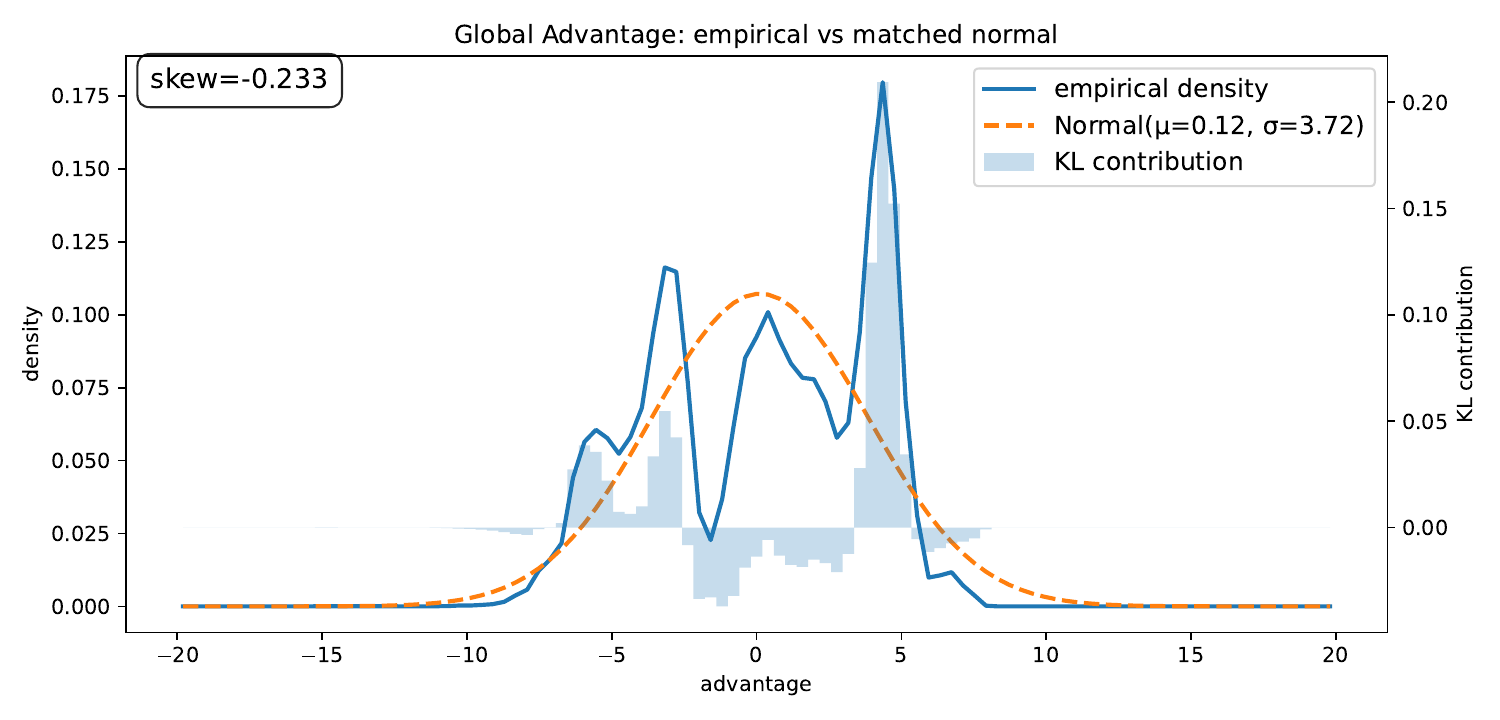}
    \caption{Checkpoint 1300}
    \label{fig:rl_adv_density_comparison_1300}
  \end{subfigure}
   \begin{subfigure}[t]{0.32\linewidth}
    \centering
    \includegraphics[width=\linewidth]{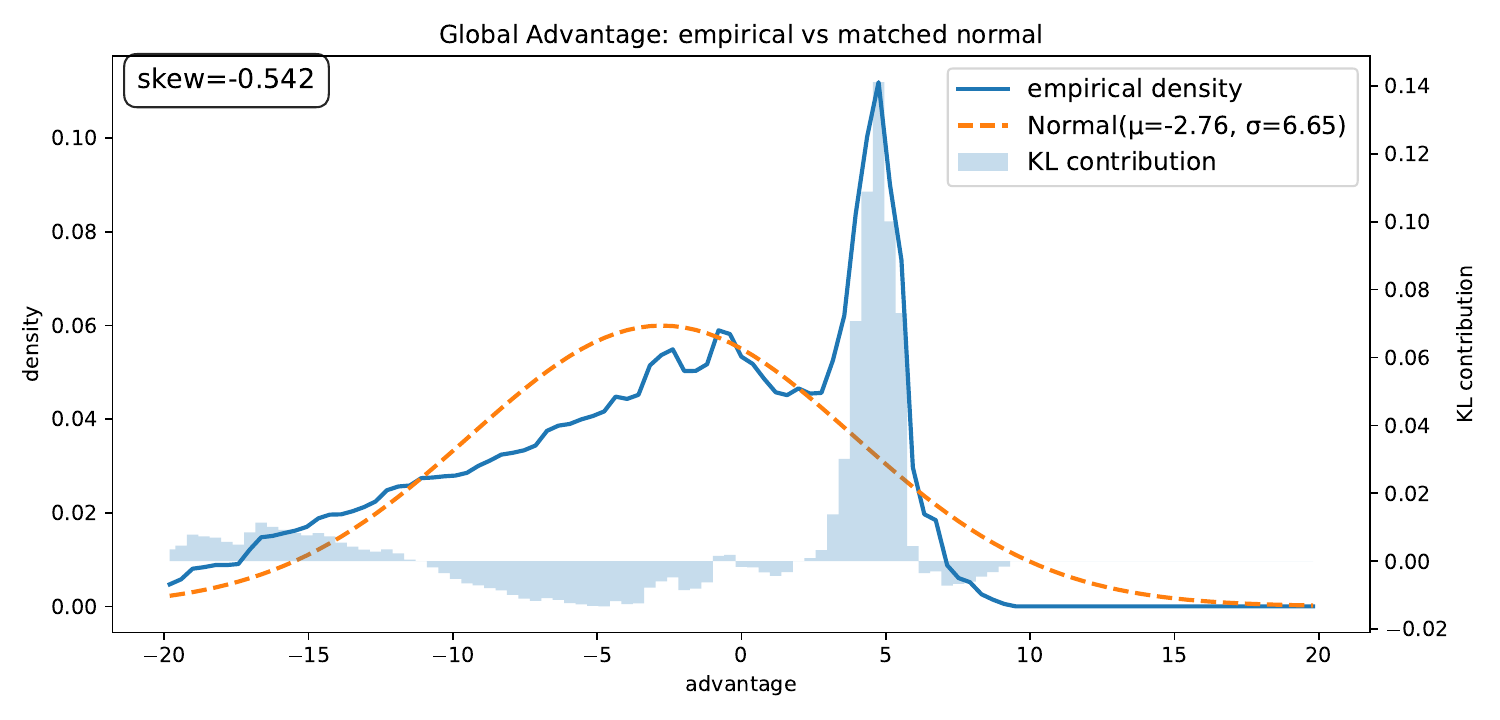}
    \caption{Checkpoint 1400}
    \label{fig:rl_adv_density_comparison_1400}
  \end{subfigure}                        
  \\
   \begin{subfigure}[t]{0.32\linewidth}
    \centering
    \includegraphics[width=\linewidth]{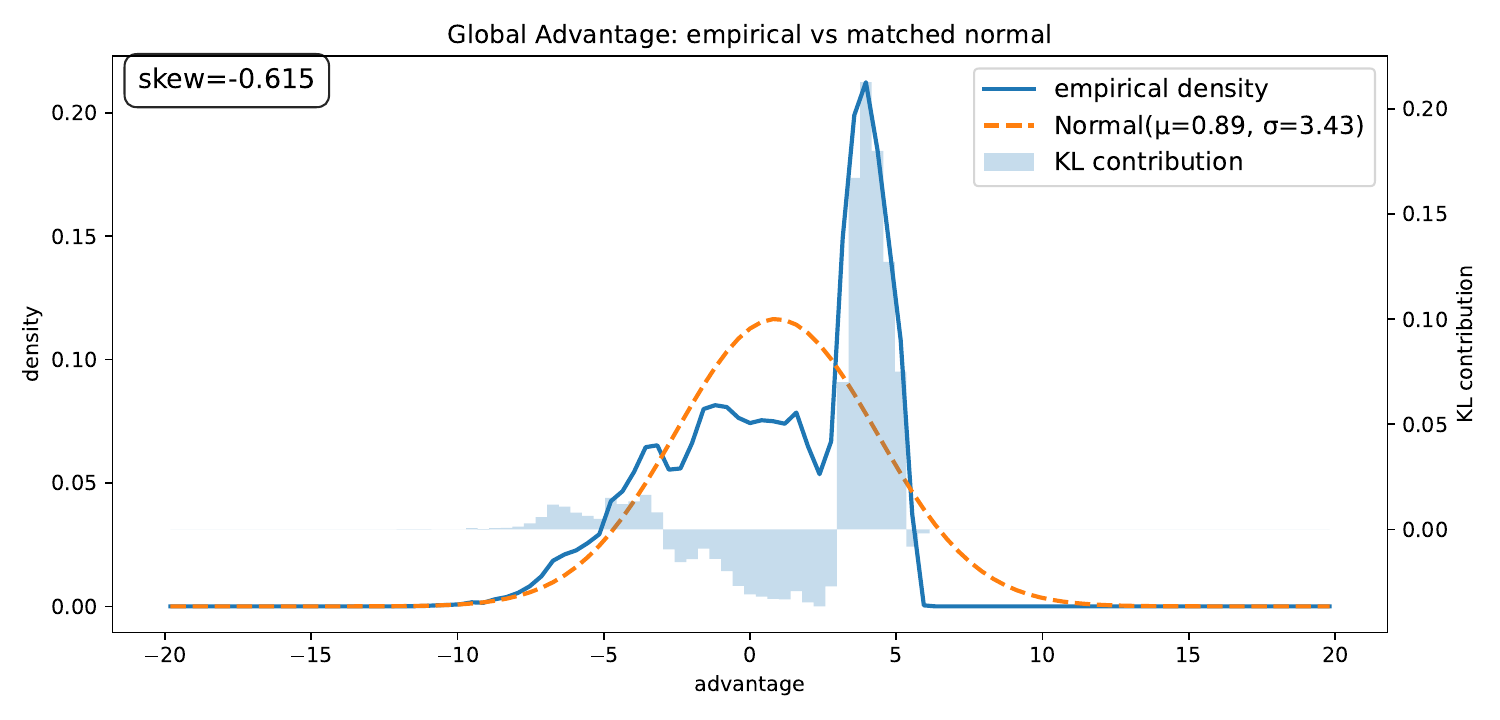}
    \caption{Checkpoint 1500}
    \label{fig:rl_adv_density_comparison_1500}
  \end{subfigure}                        
   \begin{subfigure}[t]{0.32\linewidth}
    \centering
    \includegraphics[width=\linewidth]{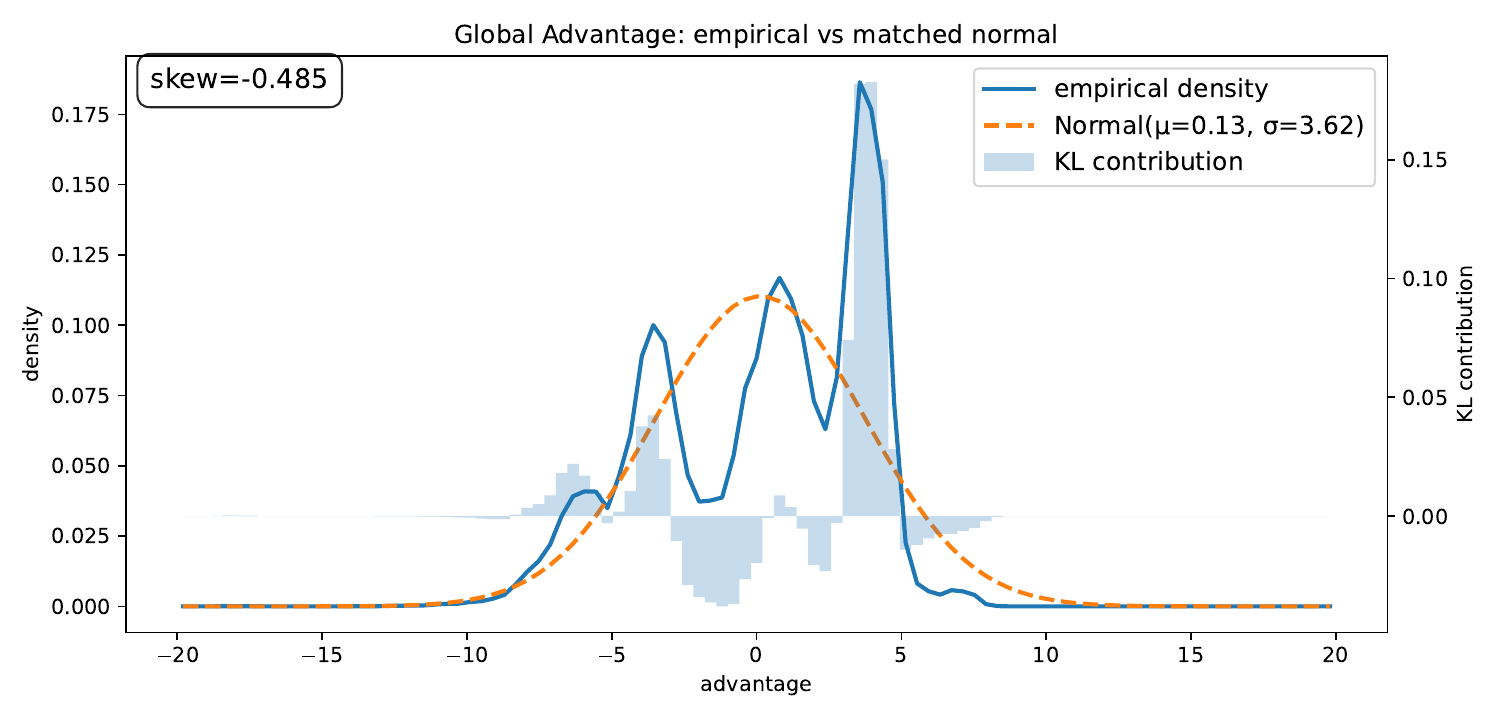}
    \caption{Checkpoint 1600}
    \label{fig:rl_adv_density_comparison_1600}
  \end{subfigure}                        
\end{figure}

\clearpage

\end{document}